\documentclass[sigconf,nonacm]{acmart}

\usepackage{booktabs}
\usepackage{multirow}
\usepackage{graphicx}
\usepackage{algorithm}
\usepackage{algpseudocode}
\usepackage{amsmath}  %
\usepackage{xspace}
\usepackage{enumitem}
\usepackage{float}
\usepackage{array}     %
\usepackage{tabularx}
\usepackage[table]{xcolor}

\usepackage{adjustbox}
\newenvironment{fitwide}%
  {\begin{adjustbox}{max width=\linewidth}}%
  {\end{adjustbox}}

\newcommand{\demowidth}{\ifdim\linewidth>300pt 0.42\linewidth\else\linewidth\fi}

\usepackage{tikz}
\usetikzlibrary{shapes.geometric, arrows.meta, positioning, calc}
\definecolor{stepfill}{RGB}{234,245,255}
\definecolor{revertfill}{RGB}{255,241,219}
\definecolor{reportfill}{RGB}{232,250,236}
\definecolor{gatefill}{RGB}{255,255,255}
\definecolor{gateAfill}{RGB}{243,232,250}

\usepackage{forest}

\newlength{\leafwidth}
\newcommand{\leafnote}[1]{\\{\scriptsize\textcolor{black!55}{\parbox{\leafwidth}{#1}}}}
\forestset{
  cmptree/.style={
    for tree={grow'=east, parent anchor=east, child anchor=west, anchor=west,
      align=left, font=\footnotesize, inner xsep=2.5pt, inner ysep=2pt,
      l sep=2.2mm, s sep=1.1mm, edge={gray!70, semithick},
      edge path={\noexpand\path[\forestoption{edge}]
        (!u.parent anchor) -- +(2mm,0) |- (.child anchor)\forestoption{edge label};},
    },
    where n children=0{draw=black!45, rounded corners=2pt, fill=blue!6,
      text width=\leafwidth, tier=rec}{font=\footnotesize\itshape},
  }
}

\usepackage{listings}
\definecolor{codegreen}{rgb}{0,0.6,0}
\definecolor{codegray}{rgb}{0.5,0.5,0.5}
\definecolor{codepurple}{rgb}{0.58,0,0.82}
\definecolor{backcolour}{rgb}{0.95,0.95,0.92}
\definecolor{recblue}{HTML}{1A5FB4}
\definecolor{rivalorange}{HTML}{A03000}
\definecolor{ppiorange}{RGB}{235, 130, 40}
\definecolor{humangreen}{RGB}{45, 160, 75}
\definecolor{uncorrectedpink}{RGB}{225, 100, 160}
\definecolor{swappedpurple}{RGB}{145, 80, 180}
\definecolor{customgray}{RGB}{128, 128, 128}
\definecolor{evalpink}{HTML}{E7298A}
\definecolor{evalyellow}{HTML}{E6AB02}

\lstdefinestyle{mystyle}{
    backgroundcolor=\color{backcolour},   
    commentstyle=\color{codegreen},
    keywordstyle=\color{magenta},
    numberstyle=\tiny\color{codegray},
    stringstyle=\color{codepurple},
    basicstyle=\ttfamily\footnotesize,
    breakatwhitespace=false,         
    breaklines=true,                 
    captionpos=b,                    
    keepspaces=true,                 
    numbers=left,                    
    numbersep=5pt,                  
    showspaces=false,                
    showstringspaces=false,
    showtabs=false,                  
    tabsize=2,
    escapeinside={(*}{*)}
}

\lstdefinestyle{terminal}{
    backgroundcolor=\color{black!88},
    basicstyle=\ttfamily\footnotesize\color{gray!15},
    lineskip=-0.5pt, %
    frame=single,
    rulecolor=\color{black!88},
    keepspaces=true,
    columns=fixed,
    showstringspaces=false,
    xleftmargin=4pt,
    xrightmargin=4pt
}

\definecolor{evalbg}{RGB}{235,235,255}
\newcommand{\pkgname}{evalstats}

\newcommand{\pkgbig}{%
  \textsf{\colorbox{evalbg}{\strut\pkgname}}%
}
\newcommand{\pkg}{%
  \texttt{\small \colorbox{evalbg}{\pkgname}}%
}
\newcommand{\pkgcode}{\texttt{\pkgname}}

\AtBeginDocument{%
  }

\makeatletter\def\@titlefont{\fontsize{16.8}{20.5}\selectfont\sffamily\bfseries}\makeatother

\begin{document}

\title[\pkgbig{}: Calibrated Statistical Inference for LLM Judges and Small-Sample AI Evaluations]{How to Run Statistics over LLM Judges and Trust the Results:\texorpdfstring{\\}{ }Calibrated Inference for Small-Sample AI Evaluation with \pkgbig{}}

\author{Ian Arawjo}
\affiliation{%
  \department{Department of Computer Science and Operations Research (DIRO)}
  \institution{Université de Montréal}
  \city{Montréal}
  \state{Quebec}
  \country{Canada}
}
\email{ian.arawjo@umontreal.ca}

\renewcommand{\shortauthors}{Ian Arawjo}

\begin{abstract}
Researchers across academia increasingly base significance claims on LLM judge scores and small-sample AI evaluations. %
Yet without well-calibrated confidence intervals (CIs), hypothesis tests, and judge-bias corrections, such claims are unreliable. We address these issues in several contributions. First, we find that running statistics over raw LLM judge scores leads to inflated false positives: counterintuitively, for many inter-rater agreement metrics, false positive risk peaks at ``almost perfect'' human-LLM agreement. To help researchers understand how to run statistics over LLM judges responsibly, we present guidance and tooling for the statistical analysis of mixed human-AI judge designs, and implement nine hypothesis tests via prediction-powered inference (PPI), including the first known PPI corrections for four rank-based tests (Wilcoxon signed-rank, Mann-Whitney U, and omnibus variants). To keep PPI++ stable with small human-labeled calibration sets, we introduce bootstrap-adaptive power tuning, which shrinks the estimated weight toward a target estimated from the labeled data, and accounts for that weight's own sampling variance. Second, through Monte Carlo simulations, we derive recommendations for what CI, $p$-value, and FWER correction methods to use for small-sample AI evaluations ($N{<}100$), and warn researchers against bootstrap CIs. We package these recommendations into \pkg{}, an open-source Python package that selects calibrated methods automatically, and demonstrate it in three scenarios, including one where a real LLM judge validated at ``substantial agreement'' would have led a researcher to publish a spurious finding. \pkg{} is publicly available at \url{https://github.com/ianarawjo/evalstats}.

\end{abstract}


\begin{teaserfigure}
    \centering
    \includegraphics[width=\linewidth]{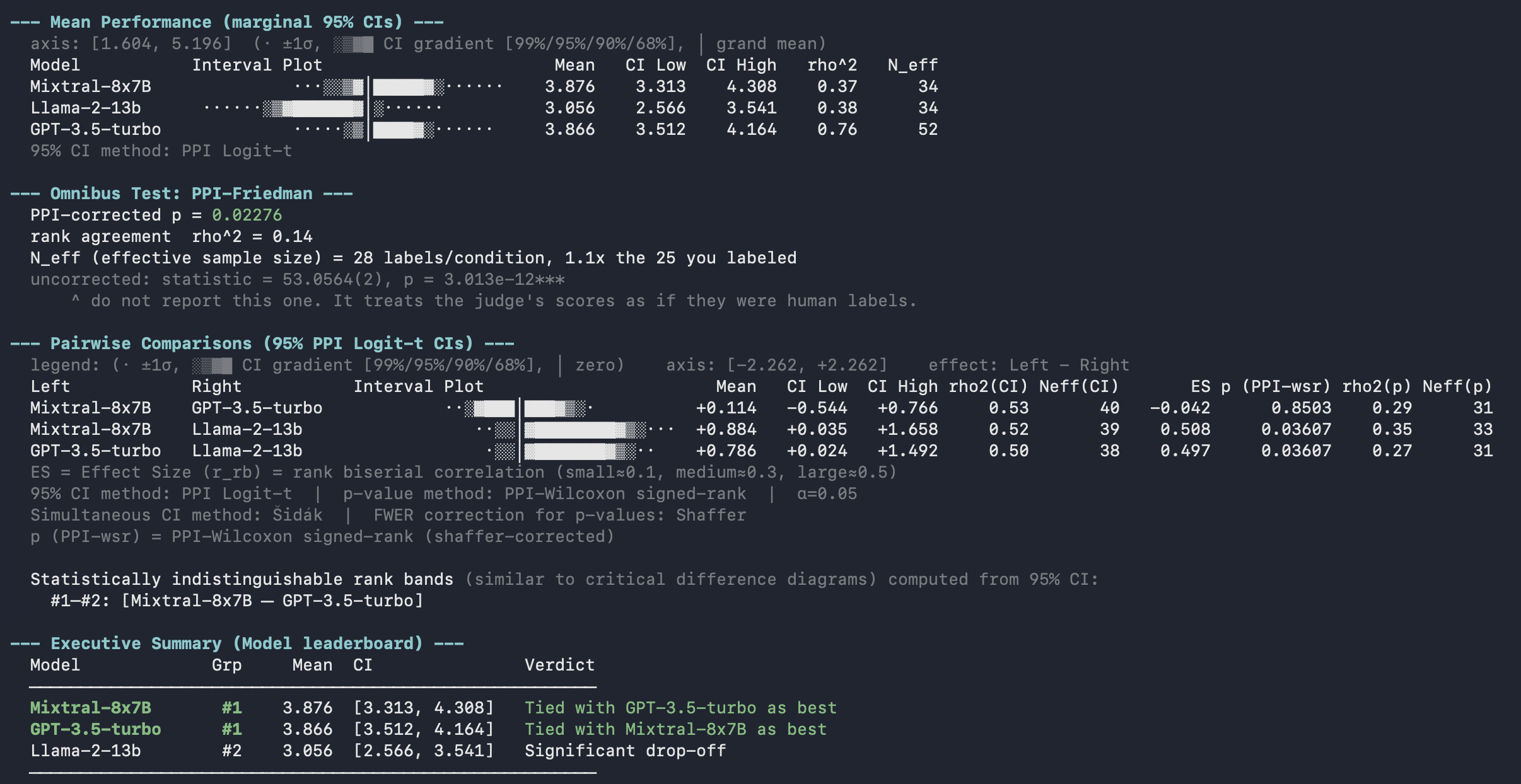}
    \caption{Running \pkg{} from the command line on a spreadsheet of AI evaluation results prints 95\% confidence intervals, $p$-values, gradient plots~\cite{correll2014error}, pairwise tests, and an executive summary, with conditions grouped into rank bands~\cite{demvsar2006statistical}. Each method was chosen automatically for the sample size and data type, backed by our simulations (\S\ref{sec:techvalid:main}-\ref{sec:ppi:main}). Shown are results from BiGGen Bench~\cite{kim2025biggen}, a benchmark that uses LLM judges as evaluators, sampling three models at $N{=}80$ items each with $N_\mathrm{lab}{=}25$ human labels as calibration. Every estimate on screen---means, CIs, tests, $p$-values, and effect size---is corrected for LLM judge bias using prediction-powered inference~\cite{angelopoulos2023prediction}. The PPI-Friedman and PPI-Wilcoxon tests shown are contributions of this paper.}
    \label{fig:terminal-model-comparison}
\end{teaserfigure}

\maketitle

\section{Introduction} \label{sec:intro}

In AI evaluation, statistics is an afterthought, if it is a thought at all. Only 16\% of over 400 published benchmark papers report any statistics, let alone the correct statistics~\cite{bean2025measuring}. Researchers across AI, NLP, and human-computer interaction (HCI) often choose improper statistical methods for experiment analysis~\cite{dror2018hitchhikers, bean2025measuring, jun2019tea}. Among tools that help users conduct small-scale AI evaluations, almost none quantify uncertainty, and many rely upon LLM judges to work~(e.g., \cite{kim2024evallm, kim2026evalet, gebreegziabher2025metricmate, chiang2026multeval, desmond2024evalullm}), raising concerns of accuracy and over-reliance. %

In response to this state of affairs, %
researchers are now calling for improved statistical rigour around AI evaluations. %
Proposed guidelines from the Canadian AI Safety Institute and NIST ask practitioners to quantify uncertainty via confidence intervals (CIs)~\cite{keller2026practices} and conduct ``error mitigation'' for LLM judges~\cite{caisi2026evaluators}. Guidelines in software engineering and HCI similarly ask researchers to evaluate LLM components on small test sets~\cite{navarro2026reporting}, run ``appropriate inferential statistics,'' 
and report ``inter-rater reliability'' (IRR) between LLM judges and human raters~\cite{baltes2025guidelines}. %

These calls are necessary, but often stop short of methodological recommendations and tooling, leaving practitioners to choose methods themselves. Where concrete recommendations do exist, they tend to assume a large benchmark regime ~\cite{evans2024addingerrorbarsevals, keller2026expanding, agarwal2021deep}, remaining silent on two common scenarios: small-sample evals, and LLM judge scores. %
As a result, well-meaning practitioners may make choices such as bootstrapping CIs, running overly conservative family-wise error rate (FWER) corrections, or running statistical tests over raw judge scores---choices which can produce miscalibrated estimates in these settings, sometimes wildly so (S\ref{sec:techvalid:main}, \S\ref{sec:ppi:main}, and \S\ref{sec:demos}). On the other hand, the methods they \textit{would} need to better trust their results %
are most likely unknown to them (Figure~\ref{fig:ci-decision-tree} and \S\ref{sec:ppi:main}). %

Of these choices, running statistical tests over raw LLM judge scores is the most dangerous. 
Published papers at AI, NLP, and HCI venues now use LLM judges as metrics in controlled experiments, run statistical tests directly over their outputs (e.g., \cite{bai2026iruler, gao2025homeworkwars, tang2026programming, nahar2026, liu2025illusion, ICLR2026_d3362f84, wang2024sotopia, zhou-etal-2024-real, sadallah2025good, panda2025accesseval, fanous2025syceval}), and produce $p$-values that scientific claims rest upon. High IRR with human raters, using phrases like ``substantial agreement,'' is often used to justify doing so. As we shall show, this practice is not only wrong, but has it backwards: for numeric metrics, false positive risk peaks \textit{at high IRR} (Figure~\ref{fig:alignment-panels}). And, even for  Cohen's $\kappa$ over binary scores, false positive risk does not go away entirely.

In AI evaluation settings---at small sample sizes, and with LLM judges---what statistical methods best quantify uncertainty? In this paper, we systematically investigate this question, %
and package the resulting recommendations and implementations into \pkg{}, a Python toolkit that automatically selects well-calibrated, uncertainty-aware methods for both settings. We make four contributions:

\begin{enumerate}
    \item \textbf{LLM judge-bias-corrected hypothesis tests and guidance for mixed human-AI judge designs} (\S\ref{sec:ppi:main}). We present guidelines for analysis of mixed human-AI judge designs (\S\ref{sec:guidelines}), and validated tests that researchers can use to perform inference over LLM judge scores. Applying prediction-powered inference (PPI)~\cite{angelopoulos2023prediction, broska2025mixed, hullman2026human}, we implement and validate judge-bias-corrected versions of nine hypothesis tests commonly used in empirical research, including, to our knowledge, the first PPI corrections for rank-based tests Wilcoxon signed-rank, Mann-Whitney $U$, and their omnibus variants. To keep PPI++ stable with small  calibration sets~\cite{mani2026nofreelunch}, we also introduce bootstrap-adaptive power tuning (\S\ref{app:ppi:adaptive}). %
    We find that, for many numeric IRR metrics, false-positive risk peaks around ``almost perfect'' agreement (\S\ref{sec:ppi-q1}), and derive rules of thumb that human-LLM judge alignment $\rho^2$ should be $\ge0.4$ to be worthwhile, and below $0.2$ the judge is too poor to use. %
    
    \item \textbf{Recommendations for statistical tests to use in small-sample AI evaluation
    settings.} There is virtually no existing guidance on which CI, $p$-value, or FWER correction methods to use in small-sample eval settings ($N{<}100$). We run extensive Monte Carlo simulations to derive recommendations across eval data types, sample sizes, and setups (\S\ref{sec:techvalid:main}; full details in Appendix~\ref{appendix:techval}). 

    \item \textbf{The \pkg{} Python toolkit} (\S\ref{sec:toolkit}). We
    package these results into an open-source Python library\footnote{\url{https://github.com/ianarawjo/evalstats}} built around a
    single \texttt{compare()} method that, given a spreadsheet of AI evaluation results, \textit{automatically} selects a
    well-calibrated CI, $p$-value, FWER correction, and (where relevant)
    PPI-corrected method from the data type, sample size, and setup, with a
    CLI, uncertainty visualizations, and PPI-corrected tests independently
    available in \texttt{scipy}-like format. Because methods validated in
    isolation can still fail in composition, we additionally validate
    \texttt{compare()} end-to-end (\S\ref{sec:e2e}).

    \item \textbf{Demonstrations of toolkit utility} (\S\ref{sec:demos}). We showcase \pkg{} through three demonstrations~\cite{ledo2018toolkitevaluating} of common scenarios developers and researchers face analyzing AI evaluation results, including one where \pkg{} prevents a spurious LLM judge finding from entering peer review.
\end{enumerate}

\textbf{Our work warns researchers never to run statistical tests directly over raw LLM judge scores.} In \S\ref{sec:demo:mixed-rater} we demonstrate how a real LLM judge validated at ``substantial agreement'' with human raters can create a spurious finding that looks very significant, but is entirely false. This false finding has nothing to do with lack of rigour: researchers ensemble judges, use a held-out validation set, have two human raters score a random subset of items, etc.---and still produce a vastly miscalibrated result. %
Authors reporting mixed human-AI judge studies therefore \textit{must} employ a correction when analyzing data. Like previous methodological contributions brought lesser-known statistical methods to computing communities~\cite{wobbrock2011art, elkin2021art, kay2016researcher, kaptein2012rethinking, kaptein2010, dragicevic2016fair, agarwal2021deep}, our work seeks to make PPI ~\cite{angelopoulos2023prediction} more accessible to researchers and practitioners, providing concrete tooling they can employ, without requiring them to abandon familiar statistical practice. The simulations undergirding our technical sections are extensive; we thus push finer details and large tables to the Appendix, and use figures to summarize the major results in the main text.

\section{Background and Related Work} \label{sec:related}

\noindent \textbf{\textit{LLM evaluation tools and statistical support for evals.}} A body of work studies tools and interfaces for LLM evaluation. EvalLM~\cite{kim2024evallm}, LLMComparator~\cite{kahng2024llmcomparator}, and ChainForge~\cite{arawjo2024chainforge} help practitioners compare and explain differences between LLM outputs, while MultEval~\cite{chiang2026multeval}, Evalet~\cite{kim2026evalet}, 
EvalGen~\cite{shankar2024evalgen}, EvaluLLM~\cite{desmond2024evalullm}, and MetricMate~\cite{gebreegziabher2025metricmate} help align evaluation criteria to user preferences. All address the ``upstream'' problem of running evaluations, not the downstream one of analyzing results statistically: none perform validated inference beyond a point estimate (LLM Comparator displays 95\% intervals but does not describe how they are computed; our own prior tool, ChainForge, plots bar charts without error bars, inviting overconfident decisions), and, at time of writing, industry tools like Weights \& Biases Weave and promptfoo offer no uncertainty quantification either.

Many of these systems also rely on an LLM to judge output quality~(e.g., \cite{chiang2026multeval, kim2024evallm, gebreegziabher2025metricmate, desmond2024evalullm}). Such LLM-as-a-judge systems rarely separate the data used to align the judge from the data used to validate it (unlike ML practice, which splits ``validation'' from ``test'' sets~\cite{hardt2026emerging}), and none help users run calibrated statistics over judge scores that incorporate the much smaller set of human annotations.

This is not the fault of authors or developers: neither they nor many AI researchers seem to know what statistics are appropriate here. Existing guidance targets large-sample benchmarks and leans on normality-assuming approximations like the Wald interval~\cite{evans2024addingerrorbarsevals, angelopoulos2023prediction}, unrealistic for today's custom, resource-limited test sets~\cite{shankar2024evalgen}. Even the rare small-sample Bayesian paired method for binary evals~\cite{dontusetheclt} deteriorates beyond $N{\approx}125$ (\S\ref{sec:ci_paired:single}, Supplementary~Fig.~S5), and the closest related work, \citet{keller2026expanding}, restricts its analysis to binary data and notes that very small benchmarks limit the usefulness of any frequentist method. In ML and NLP,  researchers have long flagged poor statistical norms, such as under-reported variance across seeds~\cite{bouthillier2021accounting, henderson2018deep} or point estimates without uncertainty~\cite{dodge2019showyourwork, dror2018hitchhikers, agarwal2021deep, evans2024addingerrorbarsevals, keller2026expanding}, prompting remedies like the \texttt{rliable} library for bootstrapped benchmark-scale CIs~\cite{agarwal2021deep}. But these target an already-fluent audience at benchmark scale, not practitioners writing small, custom eval sets~\cite{shankar2024evalgen}.

Despite this gap, researchers and developers are now being called to compute 95\% CIs on AI evaluation results~\cite{keller2026practices, caisi2026evaluators} with no recommendations for the small-sample and LLM judge regimes common in practice. \citet{navarro2026reporting} tell HCI researchers to evaluate LLM components on modestly-sized eval sets but offer no statistical guidance, and \citet{baltes2025guidelines} ask software engineers for ``appropriate inferential statistics,'' suggesting options such as bootstrap-based comparisons that our simulations find often miscalibrated at small $N$, and for ``inter-rater agreement,'' without saying how to fold judge alignment into inference. These gaps in expectation vs. reality mirror \citet{jun2022hypothesis}'s finding that statistical packages ``expect analysts to have more statistical expertise than may be realistic.''

\paragraph{\textbf{Mixed human-AI judge designs and analysis}}

Researchers are eager to use LLMs as ``judges'' of quality to scale up grading or annotation~\cite{he2024ifinacrowdsource}. \citet{kumar2026large}, for instance, found LLM judges outperformed crowdworkers at judging empathy in text chats. Researchers typically validate a judge against a smaller sample of human scores, report an inter-rater reliability (IRR) metric like Cohen's $\kappa$, then run statistics over the judge scores as if ``high IRR'' licensed treating them as human surrogates---a pattern now visible in published work at major HCI, AI, and NLP venues~(e.g., \cite{bai2026iruler, nahar2026, gao2025homeworkwars, li2025gestura, li2026nsfwgenai, liu2025illusion, ICLR2026_d3362f84, wang2024sotopia, sadallah2025good}). %
This practice may only grow, and is worrying. LLM judges carry systematic biases, toward longer or confident responses, response position, or themselves~\cite{zheng2023judging, xu2024prideprejudice, wang2024fair}, casting doubt on them as ``surrogates'' for humans~\cite{hullman2026human}. And, as we shall show (\S\ref{sec:ppi-q1}), reporting IRR alone means almost nothing for trusting claims.

Where researchers have often treated LLM judges as \textit{replacements} for human subjects (a fraught framing~\cite{agnew2024illusion}), emerging ``mixed-subjects'' framings~\cite{broska2025mixed} position them as noisy, cheap predictors that \textit{complement} a smaller set of human labels, which \citet{hullman2026human} term \textit{statistical calibration}. %
We adopt the term \textbf{\textit{mixed human-AI judge} designs} for our setting, reserving \citet{broska2025mixed}'s ``mixed subjects'' for designs where the LLM stands in for a human \textit{participant} (although our tests \textit{could} be used for the latter, see Discussion~\S\ref{sec:discussion:syntheticusers}).
Prediction-powered inference (PPI)~\cite{angelopoulos2023prediction} is one such calibration method. Rather than aligning the judge upstream, it assumes the judge may \textit{always} be biased, estimates that bias from a small human-labeled sample, and uses it to correct downstream inference---finding signal in the judge's noise to improve power to detect real effects (\S\ref{background-ppi}). Despite PPI's establishment in the ML statistics literature, researchers that we spoke to in software engineering, HCI, and even NLP seemed unaware of it.\footnote{At time of writing, a full-text search of the ACM library for ``prediction-powered inference'' yields only 10 responses.} The PPI authors provide a reference implementation, \texttt{ppi\_py},\footnote{\url{https://github.com/aangelopoulos/ppi_py}} covering means, quantiles, and regression coefficients~\cite{angelopoulos2023prediction}, but it provides no two-sample, paired, rank-based, or omnibus tests, no multiple-comparison correction, no automatic selection of methods, and no small-sample calibration of the power-tuning weight (\S\ref{app:ppi:adaptive}).

\paragraph{\textbf{Statistical methodology and toolkits}}

Statistical toolkit research is rare but has standout examples, notably by Jun~\cite{jun2022tisane, jun2019tea, jun2024rtisane, jun2022hypothesis}, with many toolkits introducing domain-specific languages (DSLs) for encoding analyst insight (such as rTisane~\cite{jun2024rtisane}, which builds mixed-effects models from a DSL description of hypotheses and causal relationships). These toolkits generally target scientists analyzing complex, multi-factor experiments at scale; AI evals bring their own needs---LLM judge score calibration, small-sample data hard to fit GLMMs to---that make dedicated support worthwhile. General-purpose libraries such as  \texttt{statsmodels} implement individual tests and corrections, leaving choice of method to the analyst. %

Second, HCI in particular has had a long history of meta-scholarship on statistical methodology. \citet{wobbrock2011art}'s Aligned Rank Transform and ARTool gave a nonparametric alternative to factorial ANOVA; \citet{kaptein2010} brought nonparametric tests to experiment analysis; \citet{kaptein2012rethinking} catalogued errors in how CHI papers interpret $p$-values, power, and effect sizes; \citet{dragicevic2016fair} argued for estimation statistics~\cite{cumming2014new}; and \citet{kay2016researcher} advocated for Bayesian methods, with \citet{jun2026priorweaver} building tooling that helps users define priors by externalizing domain knowledge. Here, the choice to PPI-correct rank-based tests derives from their common usage in experiment analysis.

Finally, our work draws upon uncertainty visualization research by Hullman, Kay, Padilla, Gleicher, and others~\cite{correll2014error, padilla2022uncertainty, kay2023ggdist}. \pkg{} adopts gradient plots, shown to produce better-calibrated inferential judgments than bar charts with error bars~\cite{correll2014error}; violin plots offer similar benefits, but gradient plots render in the terminal, take less space, and stay agnostic to the underlying CI method.

\section{Gaps, Scope, and Commitments} \label{sec:motivation}

Four gaps follow from the work reviewed above. We state each one with an example situation that a researcher or developer might find themselves in today:

\newcommand{\gapex}[1]{\\[2pt]\hspace*{0.6em}{\color{black!55}$\hookrightarrow$}~\emph{Ex: #1}}
\begin{enumerate}[label=\textbf{G\arabic*.}, leftmargin=2.1em]
    \item \textbf{Limited guidance on statistical methods for small-sample eval
    reporting ($N{<}100$).}
    \gapex{A system developer must choose the best of 8 models and 5 prompts on an
    $N{=}30$ eval set, factor in cost, and justify the choice in a written report.}

    \item \textbf{How to measure human-LLM judge alignment and carry this information into downstream inference.}
    \gapex{A developer scores 150 agent traces per configuration with an LLM judge on a 1-5 rubric, hand-labels some traces, and wants to know which of five coding agents is best.}
    \gapex{A researcher compares three writing interfaces with 1{,}000
    participants on Prolific, grades the essays with an LLM judge, and collects a subset of human ratings to check judge alignment. They report weighted Cohen's $\kappa{=}0.7$ and then run their
    statistics over LLM judge scores directly, arguing that high inter-rater reliability justifies doing so. (This mirrors a practice in recent published work at HCI, AI, and NLP venues; see \S\ref{sec:related}.)}

    \item \textbf{Limited guidance for stability testing across multiple runs or
    minor variations.}
    \gapex{An ML engineer runs $K{=}5$ independent eval runs over 12 configurations
    of a RAG pipeline before deployment. One configuration scores highest, and they
    need to know whether it is stable enough to ship.}

    \item \textbf{Poor norms for visualizing uncertainty.}
    \gapex{Comparing model performance by eyeballing whether error bars of standard errors overlap in a bar chart.}
\end{enumerate}

G1 and G2 are coupled. Judge bias correction is estimated from a small human-labeled subset (\S\ref{sec:ppi:main}), hence every quantity inherits the small-sample behavior of the CI and $p$-value method underlying it. If those methods are miscalibrated at small $n$, the correction inherits that problem. LLM judges are also widely used to evaluate AI systems~\cite{yan2024llmevaluator}, particularly in interactive tools for AI evaluation~\cite{kim2026evalet, kim2024evallm, kahng2024llmcomparator, gebreegziabher2025metricmate}. %

Lastly, a note about scope: we address downstream analysis only and assume the user has already collected evaluation results into spreadsheet format, deferring the separate upstream problems of eval set creation, construct validity, prompt optimization, and judge creation to other work. The LLM judges we consider score artifacts produced by people or models; we do not treat LLMs as stand-ins for human participants (\S\ref{sec:discussion} discusses the synthetic users case).

\subsection{Philosophy and Approach} \label{sec:philosophy}

Statistics is full of competing philosophies, rather than one right way to analyze data. Thus, before we proceed, we declare our commitments and perspective. 

\pkg{} foregrounds an \textit{estimation statistics} framework~\cite{cumming2014new, dragicevic2016fair}, emphasizing CIs and point-estimate effect sizes over null hypothesis significance testing (NHST), whose interpretive problems are well documented~\cite{kaptein2012rethinking}. However, like Statslator~\cite{masson2023statslator} and \citet{kaptein2016modernstatshci}, we still support $p$-values and tests like Wilcoxon, since many venues and researchers may expect them. Our project is also \textit{empiricist}: defaults are chosen by simulation-based evidence. We adopt a frequentist framing throughout; where our CI methods draw on Bayesian ones, we retain the term ``confidence interval'' for simplicity, while acknowledging they are technically credible intervals.

``Best method'' is also, to some degree, a matter of fit to circumstance. AI evaluation brings nuances---high covariance, many pairwise comparisons---that we engineered our investigation and package around. Our governing principles were:

\begin{enumerate}
    \item \textbf{Err slightly conservative.} CI methods target slightly-above-nominal coverage (e.g., $\ge$0.95 for $\alpha{=}0.05$), and we apply FWER correction for simultaneous pairwise comparisons.
    \item \textbf{Prefer CIs over standard errors (SEs).} SEs assume normality and invite overconfident conclusions~\cite{belia2005researchers}. We default to 95\% CIs with $\alpha{=}{.05}$, which users can adjust globally.
    \item \textbf{Prefer pairwise comparisons} and report CIs on differences, since within-item comparisons cancel shared noise and are commonplace in AI evals (e.g., benchmarking).
    \item \textbf{Prefer non-parametric methods}, as LLM outputs (pass/fail, Likert, bounded floats) are rarely normal.
    \item \textbf{Enforce a minimum sample floor}: To focus our analysis, we needed a cutoff sample size beyond which no guarantees are made. \pkg{} reports no statistics below $N{=}15$, a hard cutoff that also applies to the number of human labels under PPI correction.\footnote{This cutoff is loosely informed by our simulations, where $N{=}15$ is where Type I error control gets hardest to hold without sacrificing power, and by \citet{eyre2025ppi}, who find that PPI performs poorly when very few labeled examples are available.}
    \item \textbf{Provide automatic defaults, grounded in simulation.}
    \item \textbf{Support data types common to AI evaluation}: Binary data (0/1); continuous bounded data (BLEU scores, etc.); and discrete / Likert scores (e.g., rubrics). Unbounded numeric data is supported, but is not a focus here.
    \item \textbf{Preserve user control}: defaults can be overridden, including opting out of $p$-values and NHST output.
\end{enumerate}

\begin{table*}[t]
\centering
\footnotesize
\renewcommand{\arraystretch}{1.18}
\begin{tabularx}{\textwidth}{@{}>{\raggedright\arraybackslash}p{0.20\textwidth} >{\raggedright\arraybackslash}X >{\raggedright\arraybackslash}p{0.30\textwidth}@{}}
\toprule
\textbf{What users need to know} & \textbf{Common practice without \pkg{}} & \textbf{With \pkg{}} \\
\midrule
Is the best-scoring condition really best? &
Read overlapping error bars by eye, which does not answer the question; or run pairwise
tests and correct them via a familiar correction like Bonferroni when comparisons share items and are
therefore strongly dependent. &
\texttt{compare(factors=...)}: pairwise CIs already widened for simultaneous coverage,
plus rank bands marking statistical ties. \\
Does the result hold if I run it again? &
Averaging or pass@k~\cite{hariri2026don} across runs, which hides per-item disagreement that caused the instability. Even if trying to accommodate runs, struggle to pick an estimator that respects the nesting. &
Multi-run data is detected; per-input variance is reported alongside the means, or in
isolation via \texttt{stability()}. \\
Which condition wins once cost or latency counts? &
Rank on the primary metric and inspect the secondary one informally, eyeballing means. &
Simply passing \texttt{secondary\_metric=} on \textit{compare()} to get an uncertainty-aware Pareto front labeling each condition \emph{frontier}, \emph{dominated}, or \emph{ambiguous}. \\
Which combination of two factors is best? &
Test every factor level against every other---e.g., $8{\times}5$ grid of model-prompt pairs, for 780 comparisons---then choose a basic FWER correction that is overly conservative under strong correlation between conditions. &
The same call with two factors: \v{S}id\'ak simultaneous CIs and step-down FWER correction (Shaffer or Romano--Wolf) over the full grid. \\
\midrule
My metric is an LLM judge. Can I trust the comparison? &
Compute pooled inter-rater reliability once against a hand-labeled sample collected by convenience. &
\texttt{label} from the CLI to randomly sample items for labeling, then run \texttt{judge\_alignment()}. \\
Only part of my data is human-labeled, and I need to run statistics over LLM judge scores. &
Analyze the labeled subset alone and discard the judge scores, losing power---or keep the
judge scores, inherit their bias, and argue with reviewers or managers that high IRR with human raters warrants running familiar statistics over LLM judge scores, potentially leading to spurious findings. &
\texttt{alignment=} on \textit{compare()} propagates the labeled sample into
every estimate, CI and $p$-value via PPI correction. PPI tests recover power from the unlabeled items while remaining calibrated,
worth ${\approx}1.5\times$ the human labels at $\rho^2{=}0.4$ (\S\ref{sec:label-efficiency}). \\
\bottomrule
\end{tabularx}
\caption{How \pkg{}' features can help developers and researchers running statistics over AI evaluation results.}
\label{tab:capabilities}
\end{table*}

\section{The \pkgbig{} Toolkit} \label{sec:toolkit}

\pkg{} is an open-source Python package and command-line tool for the statistical analysis of AI evaluation results. It is organized around a single powerful function, \texttt{compare()}, which takes a spreadsheet of results in long format, with one row per item,
condition, and score (optionally with a run column when the eval was executed more than once, as well as other factors and metrics), and automatically returns well-calibrated uncertainty estimates. Specifically, in four lines of Python code (\texttt{results.csv} has item, model, and score columns): 

\begin{lstlisting}[language=Python]
import (*\pkgcode*) as es
data = es.load_from("results.csv")
result = es.compare(data, factors="model", score_range=(1, 5))
result.summary() # prints in similar style to Figure 1
\end{lstlisting}

The same analysis can be run via \texttt{\pkgname{} analyze
results.csv} from the command line, with results returned in color-coded terminal output (Figure~\ref{fig:terminal-model-comparison}). PPI-corrected tests (\S\ref{sec:ppi:main}) are also available independently in \texttt{\pkgname.tests}, with \textit{scipy}-like method signatures.

Table~\ref{tab:capabilities} summarizes what a developer must otherwise do themselves, compared to answering them with simple calls in \pkg{}. The first four rows compress work that is possible but tedious and easy to get wrong. The last two, on correcting for LLM judge bias, are not straightforward at all without our contribution in \S\ref{sec:ppi-calibration}. %

\begin{figure*}[t]
\centering
\begin{minipage}[t]{0.49\textwidth}\centering
{\footnotesize\textbf{(a) 95\% confidence interval}}\\[3pt]
\begin{fitwide}
\begin{forest} cmptree
[Estimand?
  [Mean point\\estimate
    [Binary [\textbf{Wilson} score interval]]
    [Numeric [\textbf{Logit-$t$}\leafnote{Conservative: NIG, smooth bootstrap}]]
  ]
  [Mean of paired\\differences
    [Binary [\textbf{Bonett--Price} adj.\ Wald~\cite{bonett2012adjusted}]]
    [Continuous [\textbf{Logit-$t$}\leafnote{Conservative: smooth bootstrap}]]
    [Likert / Discrete [\textbf{Bayesian NIG}]]
  ]
]
\end{forest}
\end{fitwide}%
\end{minipage}\hfill
\begin{minipage}[t]{0.49\textwidth}\centering
{\footnotesize\textbf{(b) $p$-value / FWER correction}}\\[3pt]
\begin{fitwide}
\begin{forest} cmptree
[Scope?
  [Single pairwise\\comparison
    [Binary [\textbf{McNemar} mid-$p$\leafnote{Agrees with the CI beside it}]]
    [Numeric [\textbf{Wilcoxon} signed-rank\leafnote{Conservative: sign test}]]
  ]
  [Family of\\comparisons\\($k{\ge}3$)
    [Simultaneous\\CIs
      [\textbf{\v{S}id\'ak} procedure\leafnote{Every $n$, every eval type}]
    ]
    [$p$-value\\correction
      [$n{<}30$ [\textbf{Shaffer's} procedure]]
      [$n{\ge}30$ [\textbf{Romano--Wolf}~\cite{romano2005exact}]]
    ]
  ]
]
\end{forest}
\end{fitwide}%
\end{minipage}
\caption{Decision trees for selecting (a)~a 95\% CI method and (b)~a $p$-value
or FWER-correction method, for small-samples AI evaluation settings $N{=}15$-$100$. We focus on bounded scores (binary, continuous $[0,1]$, and Likert), point estimates, and paired data, common in benchmarking. For summary plots backing these recommendations, see Figure~\ref{fig:ci-evidence}; for detailed analysis, see \S\ref{app:ci-methods}--\ref{sec:fwer}. \pkg{} chooses these methods by default in its \texttt{compare()} method (\S\ref{sec:toolkit}). Under LLM judge correction (\texttt{alignment=}), two defaults change because the method has no PPI correction: paired Likert CIs use logit-$t$ instead of NIG, and binary pairwise $p$-values come from a PPI-corrected paired $t$-test instead of McNemar (App.~\ref{app:ppi:ci-methods}).}
\label{fig:ci-decision-tree}\label{fig:fwer-decision-tree}
\end{figure*}
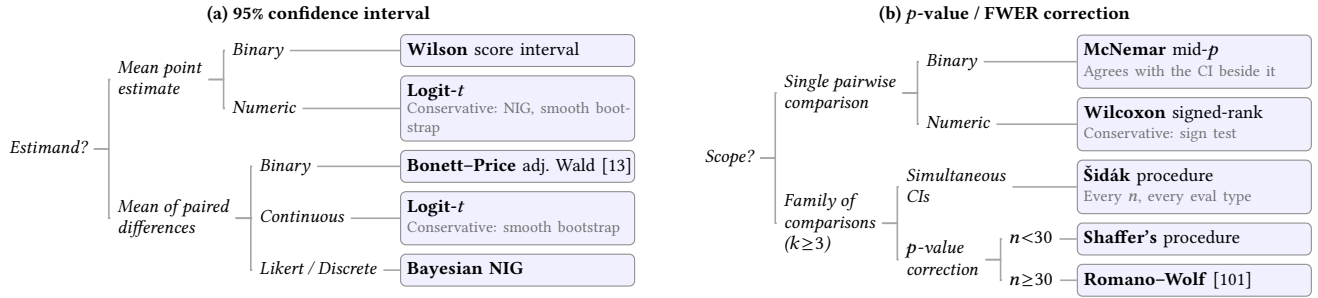

\textbf{Summary of compare().} Figure~\ref{fig:terminal-model-comparison}
shows example command-line output for a three-model comparison. For each condition,
\pkg{} reports the mean with a 95\% CI; for each pair, the mean
difference with a simultaneous CI, an effect size, and an FWER-corrected
$p$-value, plus rank bands~\cite{demvsar2006statistical} grouping conditions that are statistically
indistinguishable. An executive summary states
the verdict in plain language (``tied with Gemma as best'') at $\alpha{=}{.05}$. %
When \texttt{method="auto"}, every method in \texttt{compare()} is automatically selected (G1) from the data type, sample size, design, and
number of conditions compared, via the decision trees of
Figure~\ref{fig:ci-decision-tree}, each branch backed by the simulations in
\S\ref{sec:techvalid:main}-\ref{sec:ppi:main}. Users can override choices %
or suppress $p$-values for an estimation-only
workflow~\cite{dragicevic2016fair, cumming2014new}. %

\textbf{Communicating uncertainty (G4).} Every interval \pkg{} plots
is drawn as a gradient plot~\cite{padilla2022uncertainty} rather than an error bar as four nested CIs at
$68\%$/$90\%$/$95\%$/$99\%$, rendered as block characters of decreasing
visual opacity. Following \citet{correll2014error}, no dot marks
the mean, since a point marker invites binary readings. The encoding is agnostic to CI method.
A similar plot is also available via \texttt{result.plot()}.

\textbf{Two factors (e.g., prompts and models).} At present, users can pass up to two crossed factors. For two factors, \pkg{} compares every combination rather than one variable at a time, and outputs a heatmap in the terminal with statistically indistinguishable top performers (not shown). %

\textbf{Multi-run and two-metric data (G3).} 
\pkg{} will present multi-run CIs by default when repeated-run data is passed. These appear visually as ``noise plots'' in the terminal output, with intra-class correlation (ICC):

\noindent \includegraphics[width=\linewidth]{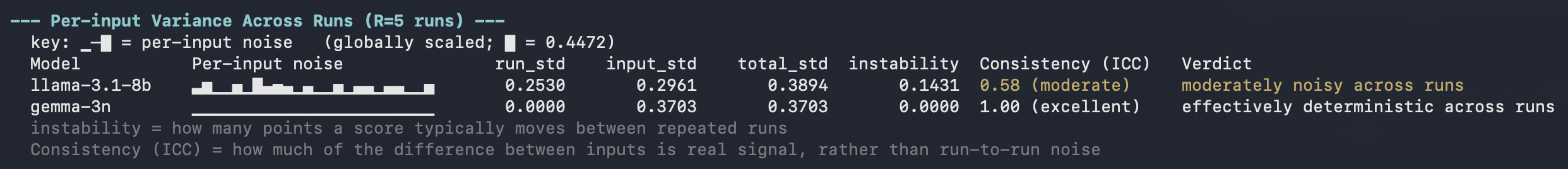}

In addition, when a \texttt{secondary\_metric=} argument is passed, \pkg{} jointly bootstraps the two metrics together (e.g., latency or cost vs. performance), plots an uncertainty-aware Pareto front, and adds a column to its Executive Summary. \pkg{} also ships CI methods for multi-run data, including multi-run adaptations of Wilson and Bonett-Price intervals. Because their construction is more involved, we defer multi-run method validation to future work.

\textbf{Untrusted metrics (G2).} What most distinguishes \pkg{} from a general statistics library is the \texttt{alignment=} argument. When the eval metric
was produced by an LLM judge, the user draws a random sample of items for human labeling with \texttt{\pkgname{} label},
then passes those labels through \texttt{judge\_alignment()}, which reports agreement
metrics, checks that the labeled subset seems random and representative, and returns an alignment object. Passing that object to \texttt{compare()} via \texttt{alignment=} declares the metric untrusted, and each estimate on screen---means, CIs, pairwise differences, and $p$-values---is corrected for LLM judge bias (Figure~\ref{fig:terminal-model-comparison}) using the calibration set. See \S\ref{sec:demo:judge} and \S\ref{sec:demo:mixed-rater}.

\textbf{Availability.} \pkg{} is open source under the MIT license at \url{https://github.com/ianarawjo/evalstats}. The repository also contains the full simulation harness behind our results. There is also a companion website, \url{https://statsforevals.com}.

\pkg{} has some limitations in its current state. Its pairwise CIs for between-subjects designs are not yet validated, since the methods of \S\ref{sec:techvalid:main} target repeated measures. It focuses on means, which are most common in AI evaluation for item-level data, rather than medians or other metrics. Multi-run CI methods' validation is not reported in this paper, and in \texttt{compare()}, PPI correction currently requires a single factor and at least 50 items and cannot be combined with multi-run data. We further discuss limitations in \S\ref{sec:discussion:limitations}.

\textbf{The rest of this paper establishes that \texttt{compare()}'s defaults for single-run data can be trusted.} \S\ref{sec:techvalid:main} finds calibrated CI, $p$-value, and FWER methods for small-sample eval regimes; \S\ref{sec:ppi:main} corrects them for an LLM judge; and \S\ref{sec:e2e} reports a final end-to-end test.

\begin{figure*}[t]
\centering
\includegraphics[width=\textwidth]{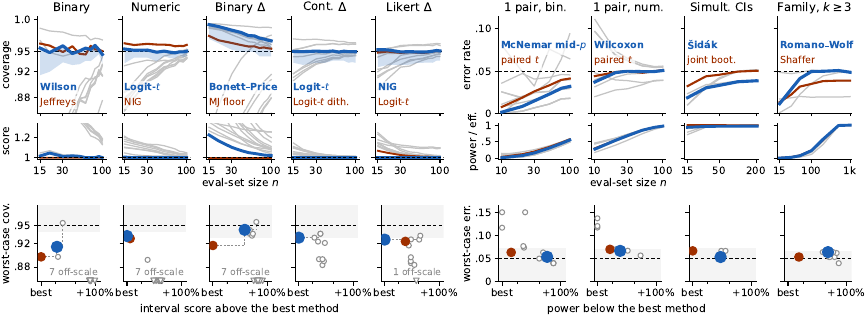}
\caption{Evidence for our recommendations in
Figure~\ref{fig:ci-decision-tree}, showing coverage rates, interval score~\cite{gneiting2007strictly}, and worst-case coverage for CI methods; and error, power, and worst-case error for $p$-values and FWER corrections. \textcolor{recblue}{\textbf{Blue}} marks the recommended method, with
\textcolor{rivalorange}{\textbf{orange}} its strongest runner-up; \textcolor{gray}{\textbf{gray}} is every other
method we tested there (9-14 per panel). Shading marks differences indistinguishable from
Monte Carlo error. Per-method values in Tables~\ref{tab:ci_single:sim}-\ref{tab:pvalues:pairwise:synth} and~\ref{tab:fwer:pvalue:synth}, and full reasoning in \S\ref{app:ci-methods}. Replicates per panel: 2{,}000 (single-sample CIs), 300 (paired CIs, pairwise $p$-values, simultaneous CIs), 500 (multi-arm).}
\label{fig:ci-evidence}
\end{figure*}

\section{What are the best-performing CI, p-value, and FWER-correction methods for small-sample AI evaluation regimes ($N{<}100$)?} \label{sec:techvalid:main}

For \pkg{}' \texttt{compare()} method (\S\ref{sec:toolkit}) to have reasonable defaults, we must establish what those defaults should be. This is also a necessity for PPI correction, especially of CIs, as the correction rests on a smaller sample. We found no prior research that covers this comprehensively for AI evaluation: some work covers specific topics, like binary data evals~\cite{dontusetheclt}, multi-run data~\cite{gonzalez2025repetitions}, or why pass@k should be abandoned~\cite{hariri2026don}, but none surveys CI methods, hypothesis tests, or FWER control across small-sample eval scenarios generally, and LLM outputs are not parametric. Rather than go by rule-of-thumb, the gold standard is to run simulations checking method performance against distributions likely to be seen in practice. 

Thus, we ran Monte Carlo simulations measuring CI, $p$-value, and FWER-correction methods across three data types common to LLM eval settings: binary, continuous [0,1], and Likert-like discrete scores (1-5).\footnote{We only consider ``bounded'' numeric data here, but unbounded data of a finite set maps easily to continuous $[0,1]$; a fourth ``grades'' suite (0-100) we tried performed virtually identically to continuous $[0,1]$.} We sometimes summarize the latter two as ``numeric.'' For CIs we tested two regimes: synthetic data across a diverse array of distributions, including adversarial scenarios like zero-inflated results, and real-data simulations by randomly sampling subsets of large-scale benchmark outputs using the OpenEval dataset~\cite{jiang2026position} and Inspect AI harness. We also weigh each method's computational cost against its performance. Alongside coverage, width, Type-I error, power, and run time, we report the
\emph{interval score}~\cite{gneiting2007strictly}, a proper scoring rule (lower is
better) that combines width and on-target coverage to give a summary of CI method performance. Since it can hide a bad coverage tail, we also report worst-case coverage (MinCov). The scenario suite is described in \S\ref{app:scenarios} with the per-method analysis in \S\ref{app:ci-methods}--\ref{sec:fwer}. 

To pick candidates to test, we drew from multiple sources: for binary proportions and their
differences, previously recommended intervals by \citet{fagerland2014recommended} (which includes \citet{bonett2012adjusted}); for numeric, the log-transform recommended by \citet{dragicevic2016fair}; hypothesis tests in NLP research~\cite{dror2018hitchhikers}; multiple comparisons corrections from a survey in bioinformatics~\cite{dudoit2003multiple}, simultaneous-CI methodology~\cite{fuchs1987simultaneous}, economics~\cite{calonico2025beyond}, and \texttt{multcomp}~\cite{bretz2016multiple}, leading us to corrections like \citet{sidak1967rectangular}, \citet{shaffer1986modified}, max-T and \citet{romano2005exact}, and \citet{westfall1993resampling}; methods that recent AI work examines~\cite{dontusetheclt,gonzalez2025repetitions,hariri2026don}; and common frequentist methods in current HCI practice~\cite{kaptein2010,dragicevic2016fair}. We also cover bootstrap methods systematically---percentile, studentized, BCa, Bayesian bootstrap, smooth bootstrap, and a joint bootstrap for simultaneous CIs---as bootstrapping CIs was previously recommended to HCI researchers by \citet{dragicevic2016fair}.

We summarize the results here via two decision trees, Figure~\ref{fig:ci-decision-tree}(a) for CI methods and (b) for $p$-value/FWER correction, which are \pkg{}' automatic defaults under \texttt{method="auto"}. Figure~\ref{fig:ci-evidence} plots a summary of evidence for each recommendation against every competing method we tested, on calibration, efficiency, and worst-case performance. Some highlights from our results include high performance for Romano-Wolf's step-down procedure~\cite{romano2005exact} for FWER correction of p-values, a technique recommended in applied economics when many tests are correlated~\cite{calonico2025beyond} (we recommend it only at $n{\geq}30$ due to few-samples instability; see \S\ref{app:rw-stability}). Our simulations also surface a new result for CIs of paired binary data: \citet{bonett2012adjusted}'s method outperforms the Bayesian paired method of \citet{dontusetheclt} on worst-case coverage and Type-I error control while running much faster. Finally, bootstrap CI methods were overconfident nearly everywhere, and on paired binary data none reached nominal coverage even at $N{=}100$. The smooth bootstrap is the one variant that holds nominal on continuous and Likert data, but it grows conservative rather than converging, which makes it a poor rule of thumb. \textbf{We strongly advise against bootstrap CIs at} $\mathbf{N{<}100}$. %

\section{How to run statistics over LLM judge scores? Mixed human-AI versions of standard hypothesis tests} \label{sec:ppi:main}

Researchers in HCI, software engineering, and AI increasingly use LLM judges to scale up data annotation and evaluate system quality~\cite{yan2024llmevaluator}. As \S\ref{sec:related} described, the usual practice is to report an IRR metric on a small human-labeled sample and then run statistics over the judge's scores directly~\cite{yan2024llmevaluator}.

Two questions follow. Does high IRR mean that LLM judge scores can be trusted as surrogates for human scores (Q1)? Second, is there a more principled approach that factors in judge quality while controlling Type I error and improving power over the human-only data, without abandoning familiar frequentist practice (Q2)?

Until now, researchers have lacked an accessible answer to Q2 for standard hypothesis tests. Thus, even if Q1 is negative, without a practical alternative, researchers using LLM judges may remain tempted to run statistics over judge scores directly. %

In this section, we address both questions. We show that high IRR does not engender trust, but often the opposite: \textbf{higher IRR is often associated with \textit{higher} false-positive risk (\S\ref{sec:ppi-q1}).} To address this issue, we make prediction-powered inference (PPI)~\cite{angelopoulos2023prediction, hullman2026human, eyre2025ppi} practical for these settings. Applying PPI, we implement LLM-judge-bias-corrected versions of nine commonly used hypothesis tests, validate their calibration on synthetic and real data (\S\ref{sec:ppi-calibration}), quantify what a mixed human-AI judge design buys in ``effective sample size'' (\S\ref{sec:label-efficiency}), and offer guidance for researchers  (\S\ref{sec:guidelines}). 

\subsubsection*{\textbf{Background on statistical frameworks to mitigate AI judge bias}} \label{background-ppi}

\begin{figure*}[t]
    \centering
    \includegraphics[width=\linewidth]{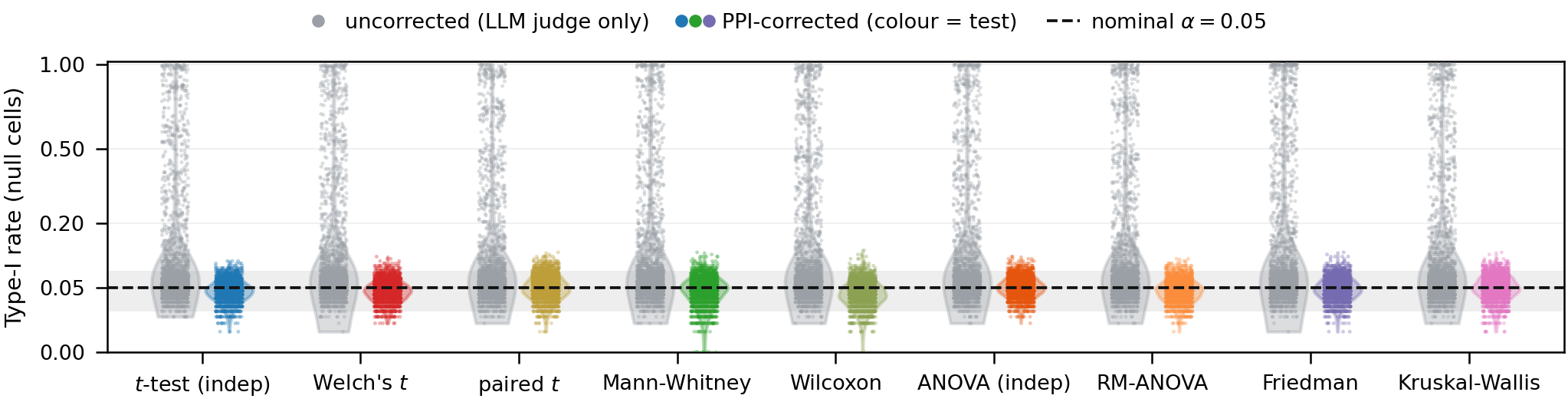}
    \caption{Type I error remains calibrated across all nine PPI-corrected tests under an aggressive factorial sweep compounding judge bias magnitude and judge noise with sample size $N$, labeled-item count $N_\mathrm{lab}$, and data type (2,046 null scenarios/test, 200 MC replicates each, $\approx$3.68M trials). PPI calibration holds even when multiple judge biases compound simultaneously. These are null scenarios under MCAR labeling; MNAR appears in Fig.~\ref{fig:typeI-factorial}. %
    Classical tests on uncorrected judge scores instead yield false positives (\textcolor{customgray}{gray} dots). The shaded band is a perfectly-calibrated test's scatter at this replicate count. %
    The $y$-axis is square-root scaled to resolve the region near $\alpha$ while keeping the uncorrected spread visible. \S\ref{sec:ppi-typeI} has full details.}
    \label{fig:ppi-type-i-error}
\end{figure*}

\textit{Prediction-powered inference} (PPI), introduced by~\citet{angelopoulos2023prediction} in \textit{Science}, combines noisy machine predictions with a smaller set $N_\mathrm{lab}$ of human labels into a single corrected estimate, offering Type I error control and, when predictions are accurate, power gains over the human-only subset. Its refinement PPI++ adds \textit{power tuning}, a data-estimated weight $\hat\lambda \in [0,1]$ on the judge scores---near $0$ when the judge adds nothing, near $1$ when it is accurate---under which, asymptotically, power never drops below the human-only subset's~\cite{angelopoulos2023ppiplusplus}. Despite this validation, PPI remains little-used in applied research, as it requires a custom implementation per test. 

Three things need attention when adopting PPI. First, PPI targets estimators built from item-wise averages, like means and regression coefficients, so applying it to widely used nonparametric tests~\cite{kaptein2016modernstatshci} like Wilcoxon requires an extra step. %
One preprint generalizes PPI to any asymptotically linear estimator~\cite{zou2026generalizedppi}, applied to binary-classifier metrics including AUC, but constructs no hypothesis tests; we are not aware of prior PPI corrections of the rank-based tests here. Concurrent work by \citet{gao2026metrics} in August 2026 PPI-corrects a paired permutation test, showing it is possible to PPI-correct a nonparametric test, albeit not a rank-based one. Second, classical PPI assumes normality via a CLT approximation~\cite{angelopoulos2023prediction}; a bootstrap variant relaxes this~\cite{zrnic2024note}, but bootstrap methods can themselves fail at small $N$. Finally, \citet{eyre2025ppi} found that PPI ``performs poorly (sometimes worse than classical inference) when very few labelled examples are available''; informed by \cite{mani2026nofreelunch}, we adopted $N_\mathrm{lab}=15$ as \pkg{}' floor for reporting LLM judge statistics.

PPI is not the only calibration framework; design-based supervised learning (DSL)~\citep{egami2023dsl} debiases inference via a different construction. We adopt PPI for three reasons: DSL is inconsistent for unknown reasons~\cite{de-ieuchon2025benchmarking}; PPI treats the judge as a black box; and PPI correction extends cleanly to familiar hypothesis tests. Other constructions such as \citet{lee2026how} focus  on point estimate CIs for binary proportions in large-sample settings, not hypothesis testing across data types.

\subsection{Validation of PPI-corrected tests} \label{sec:ppi-calibration}

\subsubsection{Monte Carlo simulations of PPI-corrected statistical tests} We implemented PPI-corrected versions of nine standard hypothesis tests: tests of two independent groups (\textit{t}-test, Welch's \textit{t}-test, Mann-Whitney U), paired/repeated measures (paired t-test, Wilcoxon signed-rank, repeated-measures one-way ANOVA, and Friedman), and $k$ independent groups (one-way ANOVA, Kruskal-Wallis).
In addition, we PPI-corrected the CI defaults of Figure~\ref{fig:ci-decision-tree}, as these are needed to PPI-correct 95\% CIs when the user passes alignment data to \texttt{compare()} (\S\ref{sec:toolkit}); we also correct the rank-biserial effect size reported alongside Wilcoxon and Mann-Whitney (\S\ref{sec:appendix-rank-based}). As PPI corrections of parametric tests are straightforward applications of standard asymptotic mean and variance PPI rectifiers~\cite{angelopoulos2023prediction, chen2026power}, we focus most of our technical details on reporting rank-based tests.\footnote{Since \texttt{compare()} requires at least $N=50$ judge-scored items for PPI correction, we do not encounter low-samples recommendations that deviate (Fig.~\ref{fig:ci-decision-tree}).}

As there are different PPI constructions, we clarify that all methods use \citet{angelopoulos2023ppiplusplus}'s power tuning (PPI++) with our bootstrap-adaptive modification, whose effect fades as $N_\mathrm{lab}$ grows (\S\ref{app:ppi:adaptive}). We validate all methods' Type I error rate under the null hypothesis and power under a battery of adversarial judge biases, effect sizes, and  data distributions, on both synthetic and real LLM judge data (\S\ref{sec:ppi-typeI} for Type-I error, \S\ref{sec:ppi-power} for power).

\begin{figure*}[t]
    \centering
    \includegraphics[width=0.9\linewidth]{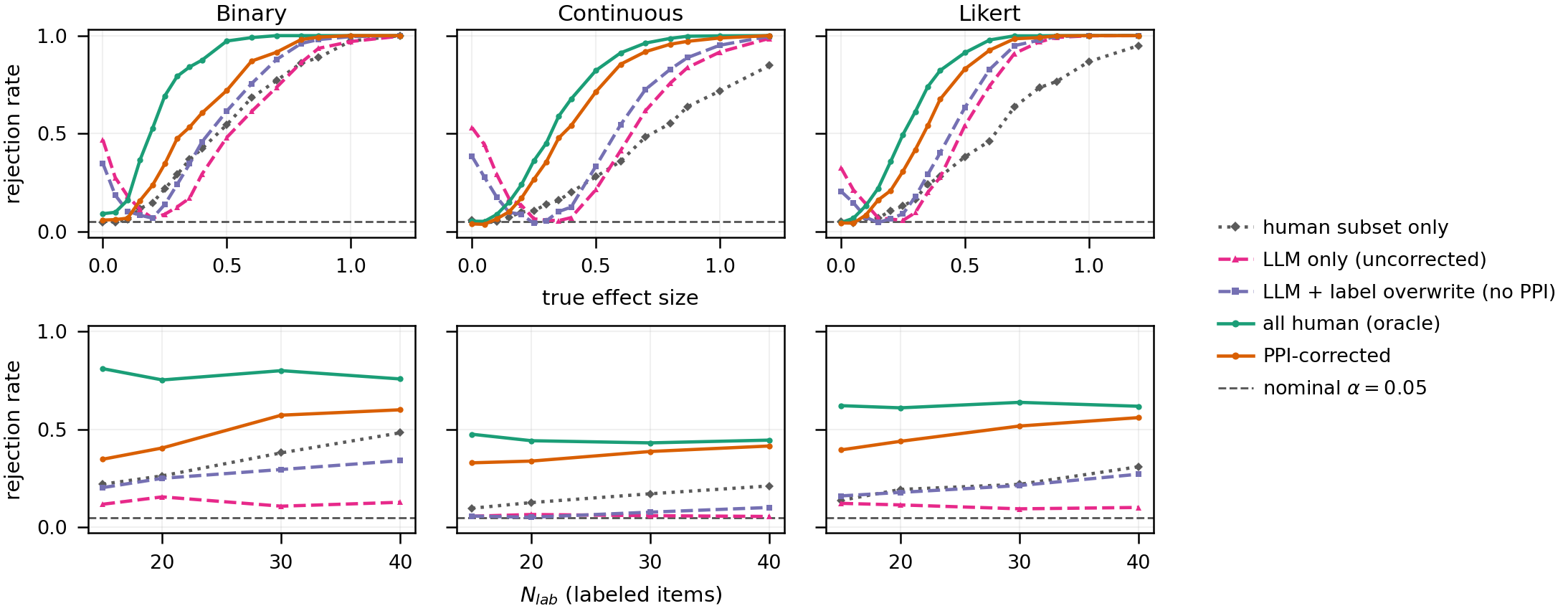}
    \caption{PPI-corrected tests (\textcolor{ppiorange}{orange}) recover power under biased LLM judges, approaching a fully-human oracle (\textcolor{humangreen}{green}) and improving on the small human-only subset (\textcolor{customgray}{gray} dotted) across all data types. \textcolor{uncorrectedpink}{Pink}: running tests over uncorrected scores show false-positive rates at zero effect; \textcolor{swappedpurple}{purple}: swapping in some human gold labels still miscalibrates. Bottom: power across human-only size $N_\mathrm{lab}$ at Cohen's $d{=}0.3$. Continuous/Likert averages five PPI-corrected tests and their classical counterparts (independent, Welch's, paired $t$-tests, Wilcoxon, Mann-Whitney U); binary averages the two $t$-tests. For more details, see \S\ref{sec:ppi-power}. 200 Monte Carlo replicates per condition (Appendix~\ref{app:ppi}).}
    \label{fig:five-way-comparison-ppi-power}
\end{figure*}

Two figures summarize our main results. Figure~\ref{fig:ppi-type-i-error} reports calibrated Type I error for all PPI tests under an aggressive factorial sweep, and Figure~\ref{fig:five-way-comparison-ppi-power} reports power for two-group tests against four alternatives, including the human-only subset and the uncorrected judge. Table~\ref{tab:ppi:typeI:pertest} provides descriptive statistics for data behind these plots. Calibration also holds for real judge data (\S\ref{app:ppi}, Fig.~\ref{fig:typeI-real-violin}) and our omnibus tests' power (Fig.~\ref{fig:five-way-comparison-omnibus}). These results  assume the human subset was selected by random sampling, meeting the missing-completely-at-random (MCAR) assumption PPI requires~\cite{angelopoulos2023prediction}; for missing-not-at-random (MNAR) regimes, see Fig.~\ref{fig:typeI-factorial}.

PPI is being steadily extended to new estimators~\cite{zrnic2024note, zou2026generalizedppi, csillag2025ppievalues}, and rank-based tests needed one such extension. Standard PPI corrects estimators built from itemwise averages~\cite{angelopoulos2023prediction}, and an item's rank depends on the whole sample, so the correction does not apply to the literal rank statistic. The usual remedy is to restate the hypothesis in terms of an estimand that is stable under subsampling, which is what we do: for Mann-Whitney, an exact $U$-statistic (dominance probability), with its power-tuning weight estimated by the PPI bootstrap~\cite{zrnic2024note}; for Wilcoxon signed-rank, a Walsh-average estimand with an analytic H\'ajek-projection; for Friedman, within-subject rank means, which need no reformulation at all; and for Kruskal-Wallis, the pairwise-dominance vector whose weighted row sums are its group rank means. To our knowledge these are the first PPI corrections of these four tests. Appendix \S\ref{sec:appendix-rank-based} breaks down each formulation, for technical readers.

Our results show good calibration for all tests, with expected under-performance when PPI's MCAR assumption is violated. Paired tests RM-ANOVA, paired-$t$, and Wilcoxon appear the most robust to mild deviations from MCAR (Fig.~\ref{fig:typeI-factorial}). Lastly, most tests assume numeric data; for binary data we validate only Welch's and paired $t$-tests, which avoid rank-based ties, and leave the rest to follow-up work.

\subsubsection{Summary of simulation sweep (\S\ref{sec:datagen:judgesynth}) for LLM judge bias scenarios.} Our simulation suite employs two complementary sweeps (Table~\ref{tab:judge-bias-sweep}): \textit{one-factor-at-a-time}, isolating a single change from a fixed baseline, and a \textit{factorial} sweep crossing several factors to catch failures that emerge only in combination, varying the data, labeling mechanism, and judge factors, each chosen to mirror a documented LLM-judge failure mode like preference for longer responses~\cite{zheng2023judging}.

\begin{figure*}[t]
  \centering
  \includegraphics[width=\linewidth]{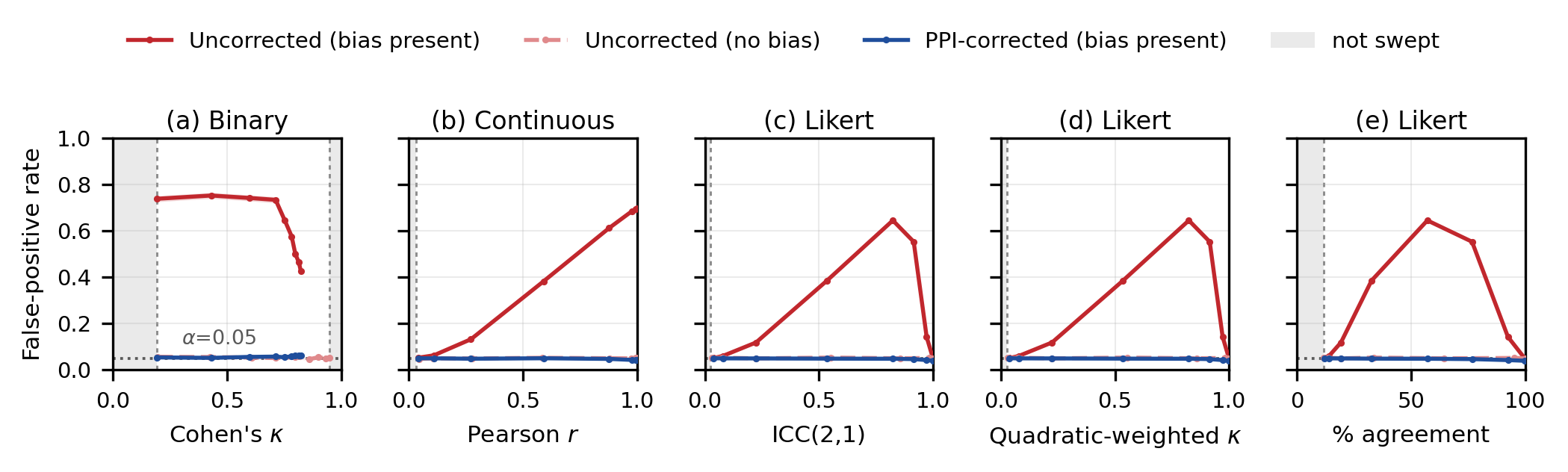}
  \caption{False-positive rate against human-LLM judge alignment on MCAR data, with five IRR metrics across three data types. Uncorrected, the false-positive rate peaks where agreement is high on Likert (c-e) and continuous data (b). PPI (blue) and a no-bias control (dashed) hold nominal $\alpha{=}.05$ throughout. Shading marks alignment our sweep did not reach. A wider sweep across 11 IRR metrics shows similar patterns (full-page plots in Appendix~\ref{app:irr-peak}). 200 MC replicates per scenario (\S\ref{app:ppi}).}
  \label{fig:alignment-panels}
\end{figure*}

\subsection{Does high inter-rater reliability certify judge trustworthiness?} \label{sec:ppi-q1}

We answer Q1---\textit{does high IRR certify judge trustworthiness?}---by showing, corroborating prior arguments~\cite{egami2023dsl, chehbouni2025llmjudgesneitherreliablenorvalid}, that a judge scoring high on standard IRR metrics can still produce badly inflated false positives. \textbf{Counter-intuitively, the risk can peak at high IRR: for most metrics on Likert data, including quadratic-weighted $\kappa$ and ICC(2,1), false-positive risk is worst around 0.80-0.90 (Fig.~\ref{fig:alignment-panels}c-e), or ``almost perfect'' agreement, remaining high through 0.90 and returning to nominal only above 0.99.}\footnote{A larger sweep crossed judge bias with noise over all four data types, up to 11 judge configurations, and up to 11 IRR metrics; full-page plots are in \S\ref{app:irr-peak}. On numeric data the takeaway holds across metrics: false-positive risk is worst at \emph{high} agreement. Only the exact peak moves. %
Exact-match metrics (percent agreement, Gwet's AC1, PABAK) are no safer, since the same number can carry either little or near-total risk at high bias. Binary Cohen's $\kappa$ is the single more trustworthy metric: higher $\kappa$ \textit{does} mean lower risk, but not nominal (e.g., $\kappa{=}0.81$ still gave an ${\approx{}}{37\%}$ false-positive rate).} On continuous data the risk rises monotonically with agreement for Pearson's $r$ (Fig.~\ref{fig:alignment-panels}b).\footnote{$r$ and Spearman's $\rho$ measure \textit{relative} agreement and can overlook systematic bias; quadratic-weighted $\kappa$ and ICC(2,1) measure \textit{absolute} agreement, yet can still be high while the judge exhibits a systematic bias.} For binary data, higher Cohen's $\kappa$ \textit{does} reduce Type I error, but does not eliminate it completely (Fig.~\ref{fig:alignment-panels}a). Thus, we conclude that LLM judge-bias-corrected statistics must be employed in all cases; it is never warranted to run statistics solely over uncorrected judge scores and trust the results~\cite{hullman2026human}.\footnote{\label{footnote:perfectly-aligned} The astute reader might wonder ``what if it's 100\% aligned?'' The answer to that is: aligned to the human sample \textit{that you have}. 100\% agreement, for instance on 30 samples, raises more questions than it answers. \pkg{} does not guard against this at time of writing, because 100\% agreement \textit{is} treated as perfect by PPI. Until we add a guard, authors might apply a conservative pseudo-count adjustment, in the spirit of Agresti \& Coull's ``add two successes and two failures'' interval for proportions~\cite{agresti1998approximate}: add two pseudo-items to the \emph{labeled} set \textit{per condition}, on which the judge misses by the smallest possible step, once in each direction (binary: one false positive and one false negative; Likert: one $+1$ and one $-1$ point; continuous: $\pm$ the smallest increment in which scores are recorded). Missing in both directions gives the correction some disagreement to learn from while leaving its bias estimate at zero. Disclose the pseudo-items.}%

\subsection{How much power do LLM judges buy you? Quantifying effective human labels} \label{sec:label-efficiency}

Establishing that PPI-corrected tests are calibrated (\S\ref{sec:ppi-calibration}) is necessary but not sufficient for adoption. A critical question remains: \textit{Is running a mixed human-AI judge design actually worth the trouble?}~\cite{hullman2026human} After mitigating for judge bias, how much statistical power does mixing human and LLM judges actually buy us, compared to just running statistics over the smaller human subset?

\begin{figure*}
    \centering
    \includegraphics[width=\linewidth]{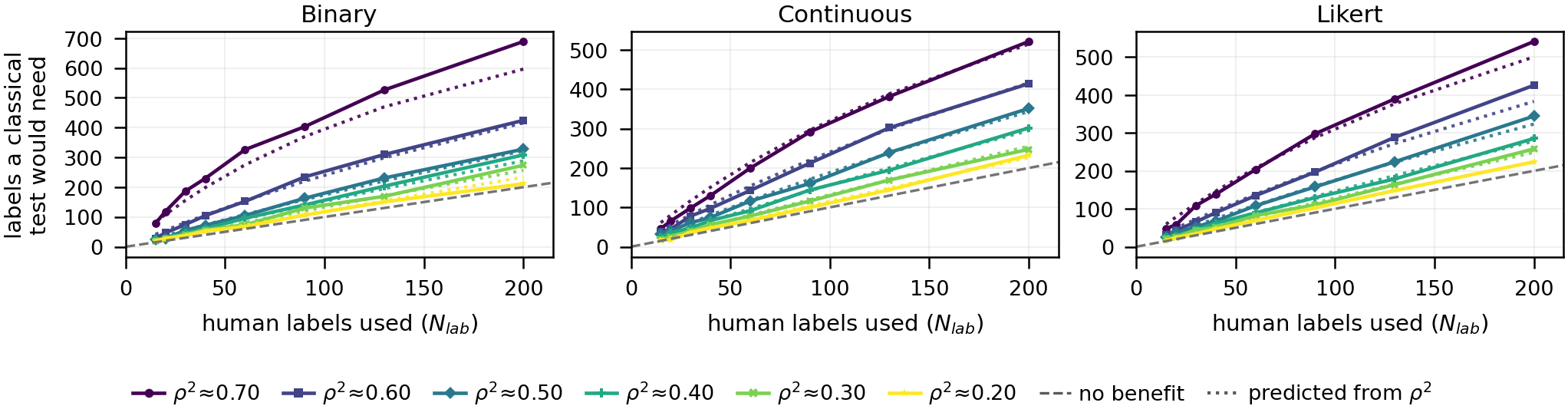}

    {\footnotesize
    \setlength{\tabcolsep}{3pt}
    \begin{tabular}{r|ccc|ccc|ccc}
    \toprule
     & \multicolumn{3}{c|}{\textbf{Binary}} & \multicolumn{3}{c|}{\textbf{Continuous}} & \multicolumn{3}{c}{\textbf{Likert}} \\
    \cmidrule(lr){2-4}\cmidrule(lr){5-7}\cmidrule(lr){8-10}
    $\rho^2$ & $n_{lab}{=}30$ & $n_{lab}{=}90$ & $n_{lab}{=}200$ & $n_{lab}{=}30$ & $n_{lab}{=}90$ & $n_{lab}{=}200$ & $n_{lab}{=}30$ & $n_{lab}{=}90$ & $n_{lab}{=}200$ \\
    \midrule
    0.70 & 5.73$\times$\,{\tiny[5.15,8.25]} & 4.48$\times$\,{\tiny[4.08,4.88]} & 3.37$\times$\,{\tiny[3.37,3.60]} & 3.74$\times$\,{\tiny[3.23,4.00]} & 3.46$\times$\,{\tiny[3.25,3.60]} & 2.68$\times$\,{\tiny[2.57,2.81]} & 3.92$\times$\,{\tiny[3.55,4.09]} & 3.55$\times$\,{\tiny[3.13,3.89]} & 2.83$\times$\,{\tiny[2.54,3.07]} \\
    0.60 & 2.44$\times$\,{\tiny[2.36,2.72]} & 2.67$\times$\,{\tiny[2.21,2.88]} & 2.15$\times$\,{\tiny[1.98,2.20]} & 2.73$\times$\,{\tiny[2.57,2.90]} & 2.54$\times$\,{\tiny[2.34,2.65]} & 2.14$\times$\,{\tiny[2.01,2.16]} & 2.20$\times$\,{\tiny[2.16,2.27]} & 2.25$\times$\,{\tiny[2.14,2.37]} & 2.04$\times$\,{\tiny[1.91,2.13]} \\
    0.50 & 1.82$\times$\,{\tiny[1.80,1.91]} & 1.81$\times$\,{\tiny[1.63,1.99]} & 1.63$\times$\,{\tiny[1.62,1.67]} & 2.10$\times$\,{\tiny[1.94,2.24]} & 1.90$\times$\,{\tiny[1.81,2.00]} & 1.78$\times$\,{\tiny[1.75,1.82]} & 1.77$\times$\,{\tiny[1.74,1.81]} & 1.72$\times$\,{\tiny[1.67,1.87]} & 1.71$\times$\,{\tiny[1.53,1.90]} \\
    0.40 & 1.54$\times$\,{\tiny[1.51,1.63]} & 1.55$\times$\,{\tiny[1.47,1.67]} & 1.54$\times$\,{\tiny[1.49,1.59]} & 1.89$\times$\,{\tiny[1.56,1.98]} & 1.65$\times$\,{\tiny[1.59,1.67]} & 1.59$\times$\,{\tiny[1.50,1.64]} & 1.47$\times$\,{\tiny[1.44,1.51]} & 1.42$\times$\,{\tiny[1.40,1.49]} & 1.42$\times$\,{\tiny[1.36,1.45]} \\
    0.30 & 1.38$\times$\,{\tiny[1.31,1.39]} & 1.45$\times$\,{\tiny[1.37,1.50]} & 1.37$\times$\,{\tiny[1.26,1.47]} & 1.39$\times$\,{\tiny[1.32,1.44]} & 1.31$\times$\,{\tiny[1.30,1.39]} & 1.28$\times$\,{\tiny[1.23,1.39]} & 1.40$\times$\,{\tiny[1.32,1.49]} & 1.27$\times$\,{\tiny[1.21,1.28]} & 1.27$\times$\,{\tiny[1.21,1.31]} \\
    0.20 & 1.27$\times$\,{\tiny[1.15,1.34]} & 1.18$\times$\,{\tiny[1.16,1.21]} & 1.06$\times$\,{\tiny[0.93,1.17]} & 1.31$\times$\,{\tiny[1.20,1.35]} & 1.18$\times$\,{\tiny[1.11,1.22]} & 1.18$\times$\,{\tiny[1.15,1.18]} & 1.23$\times$\,{\tiny[1.10,1.32]} & 1.13$\times$\,{\tiny[1.09,1.16]} & 1.12$\times$\,{\tiny[1.10,1.16]} \\
    \bottomrule\end{tabular}}
    
    \caption{How many human ratings can LLM judges save, once judge bias is corrected for? \textbf{Top}: the label-efficiency multiplier ($N_\mathrm{lab}'/N_\mathrm{lab}$) plotted across eval types, the factor by which a PPI-corrected test with $N_\mathrm{lab}$ human labels out of $N$ judge-scored items behaves like a larger human-only sample $N_\mathrm{lab}'$, for $N{=}1000$. Judge quality is $\rho^2$, the squared judge-human correlation (Pearson for parametric, Spearman for rank-based tests). Each point pools labeling budgets, effect sizes, two-sample tests, and judge-error shapes. Dotted lines show the predicted efficiency $1/(1-\rho^2(1-N_\mathrm{lab}/N))$. %
    \textbf{Below}: a table form of the same data, expressed in terms of the multiplier. $2\times$ means PPI doubled the effective sample size, acting like double the human raters. Each cell is the median over four effect sizes with a bootstrapped 95\% CI. Likert and Binary exceed their predictions slightly at high $\rho^2$ through a power-curve inversion artifact. See Appendix~\ref{app:label-eff-rt} for simulation details. }
    \label{fig:label_efficiency}
\end{figure*}

We report this as an \textit{effective human labels} curve (Figure~\ref{fig:label_efficiency}): for a fixed human budget $N_\mathrm{lab}$ and judge, how many labels $N_\mathrm{lab}'$ would a human-only setup need to match the power a PPI-corrected mixed design reaches with $N_\mathrm{lab}$ labels plus the full judge-scored dataset? (We pool rejection rates across the three $t$-tests, Mann-Whitney U, and Wilcoxon for continuous/Likert data; and Welch's and paired $t$-tests only for binary.) The $y{=}x$ diagonal in Fig.~\ref{fig:label_efficiency} marks ``no benefit from the judge,'' and the dotted line is the theoretical prediction.

Following \citet{broska2025mixed} and \citet{chen2026power}, PPI++'s label efficiency follows $1/(1-\rho_P^2(1-N_\mathrm{lab}/N))$, a special case of the general result that PPI variance reduction is governed by $(1{-}\operatorname{Corr}^2)$ between the influence functions of the labeled and predicted statistics~\cite{angelopoulos2023ppiplusplus}. Rank-based tests break linearity, so we predicted they would instead track Spearman's $\rho_S$ at small-to-moderate effect sizes; our simulations confirm this for both Wilcoxon and Mann-Whitney $U$ (\S\ref{app:label-eff-rt}). Two caveats: rank tests' effective $\rho^2$ declines as the true effect grows (negligibly below $d\le0.5$), and which test is more label-efficient depends on the shape of LLM judge noise, reversing exactly like the tests' classical counterparts do~\cite{MCKEAN2003891}---if judge noise is non-Gaussian in practice, as is plausible, rank-based tests may be more efficient. Thus, the specific $\rho$ depends on the test used---for a two-group parametric test or point estimate correction, this is Pearson's $\rho_P$; if non-parametric, Spearman's $\rho_S$. Table~\ref{tab:which-rho} provides the $\rho$ for omnibus tests.

Our simulations show that practically meaningful gains begin around $\rho^2{\ge}0.4$, regardless of metric, data type, effect size, test, or judge-noise shape (Appendix~\ref{app:label-eff-rt}, Figs~\ref{fig:le-threshold}, \ref{fig:le-lookup}, \ref{fig:le-esinv}). Specifically, the \textit{lower bound} of the estimated efficiency multiplier at $\rho^2{=}0.4$ in our simulations was $1.36\times$ (Fig.~\ref{fig:label_efficiency}), with the average at $1.57\times$ (Fig.~\ref{fig:le-threshold})---meaning that $\rho^2{=}0.4$  guarantees at least \textit{some} gains to offset the costs of using an LLM judge. Thus we answer ``How good does a judge need to be?'' with a rule-of-thumb: \textbf{Researchers should aim for at least $\rho^2{=}0.4$ human-LLM judge alignment, as at this scale, there is an ${\approx}1.5\times$ gain in effective sample size.}  At that threshold Cohen's $\kappa$ (binary) and quadratic-weighted $\kappa$ (Likert) are ${\approx}0.6$, so as a corollary, we can reframe this in terms of \citet{landis1977measurement}'s bands for inter-rater reliability: \textbf{Researchers should try to attain at least ``substantial agreement'' ($\kappa{\ge}0.61$) between their LLM judge and human raters on their validation set.} This is the \textit{minimum} alignment to target, not a suggestion for what $\rho^2$ \textit{should} be: gains rise steeply above it (${\approx}2.3\times$ at $\rho^2{=}0.6$). 

We can also address the related question of when a judge is too poor to be usable at all. That secondary threshold is around $\rho^2{\approx}0.2$, as at that point, we cannot promise any meaningful gains: the worst cell we measured returns an average of $1.06\times$ (Fig.~\ref{fig:label_efficiency}). This leads to a follow-up: \textbf{If the test-appropriate $\rho^2$ correlation between the human raters and LLM judges is below 0.2, the judge is too poor to justify its use.} At $\rho^2{\approx}0.2$, Cohen's $\kappa$ (binary) and quadratic-weighted $\kappa$ (Likert) are ${\approx}0.43$, so in \citet{landis1977measurement}'s terms for inter-rater reliability: \textbf{If $\kappa$ is below ``moderate agreement'' (${<}0.41$), the LLM judge is not worth the trouble.} %

\subsection{Guidance for running mixed human-AI judge designs and analyzing results with PPI-corrected hypothesis tests} \label{sec:guidelines}

To help authors understand how to run mixed human-AI judge designs and analyze results in a manner that respects PPI-corrected tests, we provide the following guidelines.

\textbf{Step 1: Confirm a mixed human-AI judge design is worth it.} Mixing human and LLM judges is typically worthwhile only if the full dataset is too large or costly to grade entirely by hand, and the construct is well-defined enough that human-human alignment would be ``substantial agreement'' or above; otherwise label everything by hand or with a trusted automated metric. (It is \textit{never} correct to report only LLM judge scores without alignment metrics, unless the judges are themselves the object of study.) PPI's benefits scale with how many more judge labels there are than human labels (e.g., 50 human vs.\ 1{,}000 judge). Estimate the effective sample size $N_{\text{eff}}$ with~\cite{broska2025mixed}:
\[
N_{\text{eff}} \;=\; \frac{N_{\text{lab}}}{1 - \rho^2\left(1 - \dfrac{N_{\text{lab}}}{N}\right)}
\]
where $N_{\text{lab}}$ is human-labeled units \textit{per condition} (for repeated measures, participants labeled on all $k$ conditions), $N$ is the total, and $\rho^2$ is the squared judge-human correlation on the labeled subset. Since you have no labels yet, treat this as a power analysis: posit $\rho^2$ from a prior validation of a similar judge, or sweep a plausible range as a cost analysis. 

Which $\rho$ depends on your test (Table~\ref{tab:which-rho}); with 3+ conditions, compute it via \pkg{}' \texttt{judge\_alignment} function rather than by hand. Rank-test $\rho$ declines with effect size (\S\ref{sec:label-efficiency}), so treat a rank-test $N_{\text{eff}}$ as an upper bound. 

\textbf{Step 2: Tune your LLM judge on data \emph{*disjoint*} from your human-labeled subset.} When building an LLM judge, tune it on a separate, held-out validation set \textit{completely disjoint} from data  used in PPI correction. Even informal adjustments (eyeballing disagreements and tweaking the prompt) count as cheating, like $p$-hacking: the judge overfits to the labeled items, and the PPI rectifier underestimates its error. Either randomly split off the validation set \textit{before looking at the data}, or collect it in a pilot; prefer splitting for rank-based tests, since $\rho$ there is sensitive to score spread.

\textbf{Step 3: Randomly sample the items that will get human labels.} Draw the items for human labeling $N_\mathrm{lab}$ completely at random from the full dataset, stratified by condition/measure. This is not optional: the guarantees of PPI correction depend on it. For within-subjects designs, sampling must be \emph{coupled across conditions}: the same items need a human label in \textit{every} condition, as our paired corrections only work when an item's full row has human scores. For between-subjects designs, sample each group independently. If true random sampling is not possible (MNAR), PPI-corrected tests are still worthwhile overall, but disclose the violation and prefer paired tests (paired $t$-test, Wilcoxon signed-rank, RM-ANOVA), the most robust under MNAR in our simulations (Fig.~\ref{fig:typeI-factorial}).
 
\textbf{Step 4: Collect enough human labels.} Aim for at least 30 human labels per condition/measure ($N_\mathrm{lab}{=}30$). The floor is 15: \pkg{} reports no PPI statistics below it, since PPI can perform poorly with very few labels~\cite{eyre2025ppi}. %
 
\textbf{Step 5: Check your judge's alignment with human labels.} Compute the test-appropriate $\rho^2$ (Step 1) between judge and human labels. Aim for $\ge0.4$; below $\approx0.2$, a corrected test will likely not outperform the human-only subset (\S\ref{sec:label-efficiency}). Other IRR metrics are welcome too, but $\rho^2$ must be reported, since it directly quantifies PPI's efficiency; \pkg{}' \texttt{judge\_alignment} reports it when given the planned test (\texttt{test=}). If the LLM judge lands below $0.2$ at this stage, two options remain: run classical statistics over the human-only subset and scrap the judge data; or improve the judge and re-validate $\rho^2$ on a held-out set, never the study data itself, since reusing it would violate the MCAR assumption (Step 2).
 
\textbf{Step 6: Apply PPI correction regardless of how good the IRR looks.} Correction is \textit{mandatory}, not conditional on alignment quality (\S\ref{sec:ppi-q1}): a high IRR metric does not license running statistics directly on LLM judge scores. (If for some reason you achieved ``100\% alignment,'' see Footnote~\ref{footnote:perfectly-aligned}.)
 
\textbf{Step 7: Report the full setup alongside your result.} We offer a checklist as a guideline:

\vspace{0.2cm}

\begin{center}
\fbox{\parbox{0.94\linewidth}{\small
\textbf{Box 1: Reporting checklist for mixed human-AI judge designs.} When publishing results, report:
\begin{itemize}[leftmargin=1.4em, itemsep=1pt, label={\large$\square$}]
  \item The judge's exact model version, prompt, and configuration
  \item How units were sampled for labeling (at random, ideally; disclose deviations, per Step~3)
  \item How the judge was built and validated, with no contamination ensured between validation and study data
  \item Human rater recruitment and process; with multiple raters, human-human IRR and the aggregation rule (mean or median)
  \item Judge-human alignment: the test-appropriate $\rho^2$ (Table~\ref{tab:which-rho}), per condition/measure, from \pkg{}' \texttt{judge\_alignment}; %
  and any other IRR metrics expected in your setting
  \item Which PPI-corrected method variant was applied, its result (point estimate and CI, or $p$-value), \textbf{and} the effective sample size $N_{\text{eff}}$ (i.e., for each CI, and if using hypothesis tests, for the omnibus and each pairwise test).
\end{itemize}}}
\end{center}

\vspace{0.2cm}

\pkg{} automatically produces statistical quantities on this checklist when researchers call \texttt{judge\_alignment} and mark a metric as an LLM judge in \texttt{compare()} (the latter shown by Figure~\ref{fig:terminal-model-comparison}).

\begin{figure*}[h]
  \centering
  \includegraphics[width=\linewidth]{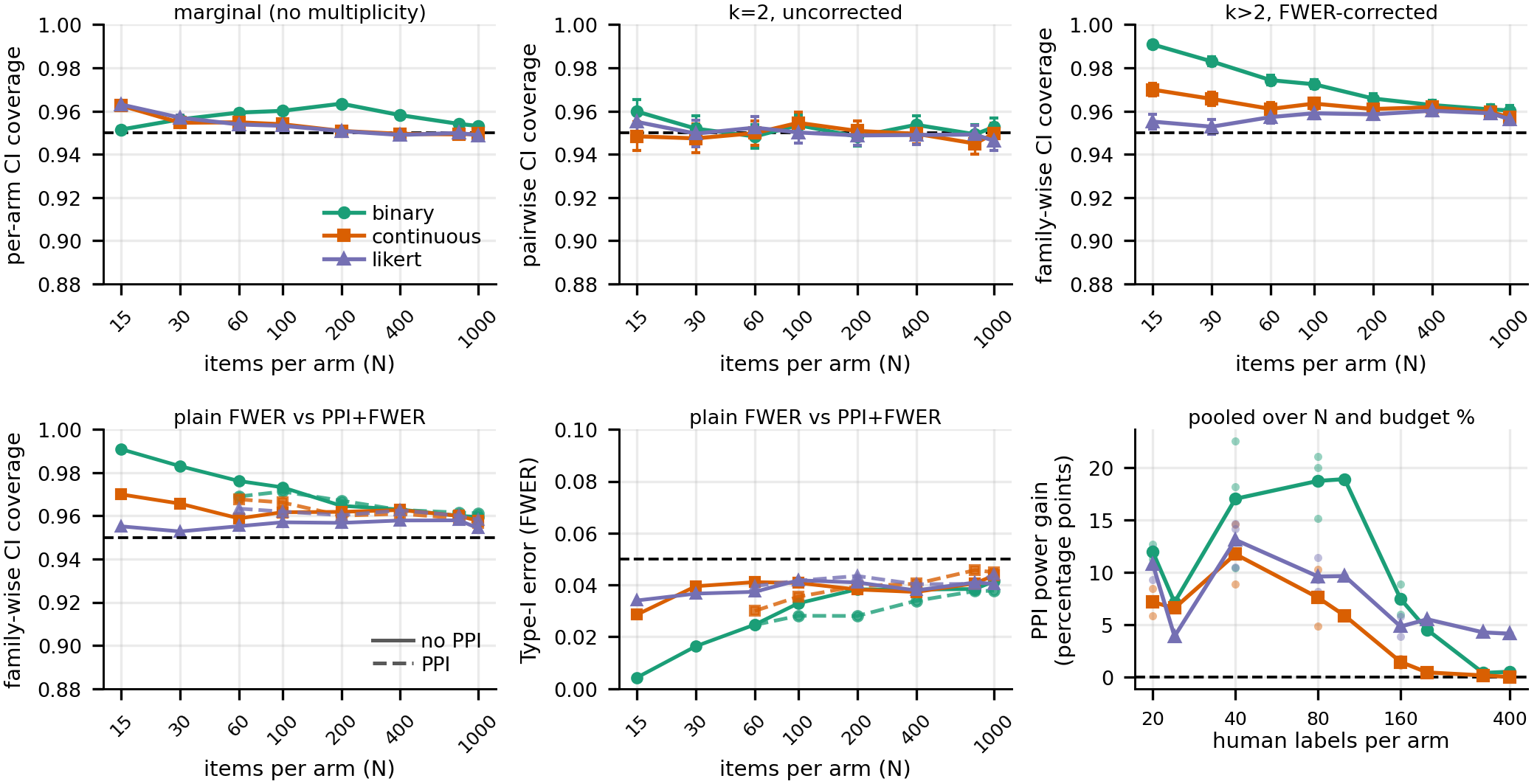}
  \caption{End-to-end validation of \pkg{}' \texttt{compare()} with every method choice left to
  \texttt{"auto"}, across 7{,}176 cells
  with 100 MC repetitions. \textbf{Top row}: CI method coverage at increasing multiplicity:
  per-arm CIs, pairwise CIs at $k{=}2$, and FWER-corrected simultaneous CIs at $k>2$. \textbf{Bottom
  row}: family-wise CI coverage and $p$-value Type-I error split by non-PPI and PPI scenarios (dashed). Bottom-right shows the power that PPI tests recover after FWER-correction for $k{=}3$ groups, over
  analyzing the human-labeled subset alone. Dashed black lines mark nominal ($95\%$ coverage, $\alpha{=}.05$, and zero gain). %
  Coverage is nominal at every level, Type-I is controlled with and without PPI, and the PPI gains hold. For sweep details behind this plot, see \S\ref{app:e2e}.}
  \label{fig:e2e}
\end{figure*}

\newpage
\section{End-to-end validation of \texttt{compare()}} \label{sec:e2e}

So far, we have provided recommendations and tests. However, our simulations were run in isolation; %
unknown deficiencies could emerge when methods interact. %
Thus, we ran a final end-to-end test, where we fed our  simulations' data generation suite directly into \pkg{}' \texttt{compare()}, leaving method choices to \texttt{method="auto"}. Sweeping %
across binary, continuous, and Likert data, varying sample size $N{=}15$ to $1000$, number of conditions $k{=}2$ to $10$, and labeling budget, across 717{,}600 calls, estimates are near-nominal at every level for every datatype, from per-arm CIs through FWER-corrected simultaneous CIs, and Type-I error of $p$-values remains controlled (Figure~\ref{fig:e2e}). Critically, FWER correction does not destroy our PPI test's gains, leaving coverage intact and costing a fraction of Type-I error while recovering up to ${\approx}14$ (on average ${\approx}9$) percentage
points of power over running tests over the human-labeled subset alone at $k{=}3$. %
See \S\ref{app:e2e} for full details.

\section{Demonstrations} \label{sec:demos}

Having established that \pkg{}' \texttt{compare()} is well-calibrated, what can we do with it? Toolkits are commonly evaluated by a demonstration, which walks a reader through what the package makes possible, coupled with a second evaluation such as technical validations~\cite{ledo2018toolkitevaluating}. Thus, to demonstrate \pkg{}, here we follow a student building an AI-powered job interview practice companion. We conclude with a different example of a researcher analyzing mixed human-AI judge data for a between-subjects study and preparing it for publication. Reported values are from real calls to \texttt{compare()}, on data that is either real or, where noted in a footnote, synthetic, but the narrative around the data has been reframed to keep the story consistent.

\subsection{Choosing a model and prompt for the dialogue component}

Imagine a student is building an AI-powered job interview companion. It has two LLM components: a dialogue component that plays the interviewer, and a feedback component that offers constructive feedback on the user's performance (\S\ref{sec:demo:judge}). Even with the dialogue component, the student faces three decisions: which prompt to use, whether the model behind it is stable across repeated runs, and whether the prompt responds fast enough for live conversation.

\textbf{Which prompt?} Before the dialogue component replies, it sorts the user's last message into a category and answers accordingly. Latency matters here, so the student chooses \texttt{gemma3:1b}, an on-device, small model. The student writes eight candidate prompts, ranging from a bare instruction, to a persona, to few-shot examples, chain of thought, structured JSON output, and negative phrasing. They create a 20-item eval set, hand-labeling ground truth, and score all eight prompts against those labels. (This step uses real data, with a known ground truth.)\footnote{We made a support-ticket classification eval, which is distributed in the \pkg{} repository (\texttt{examples/support\_ticket\_prompts.py}; analysis in \texttt{simulations/demo\_tickets\_smalln.py}). All eight prompts were run over 120 tickets, five times each, over \texttt{gemma3:1b}. 20 items are drawn from those tickets at random, and the full 120 tickets serve as ground truth for what a 20-item eval could have concluded.}

Let us imagine what someone without \pkg{} may do. The status quo in AI evaluation is to eyeball performance without any uncertainty estimates~\cite{bean2025measuring}, as systems for LLM evaluation, whether research prototypes or industry tools, rarely offer it. 

If the student eyeballs a bar chart to determine their prompt in this example, they will choose a sub-optimal prompt 43\% of the time (5,000 draws of 20 items from the full 120).

If the student adds error bars as standard errors---also a common practice in AI and science~\cite{evans2024addingerrorbarsevals, belia2005researchers}---and eyeball whether the bars overlap, they find nothing, since the bars separate on only $0.1\%$ of the time.

Third, if they try a bootstrapped CI, which they heard somewhere was robust, and do pairwise comparisons for every unique prompt pair (28 total), the percentile bootstrap's intervals will be too narrow, holding the true difference on only 88\% of draws rather than the expected 95\%. The BCa bootstrap is worse, separating truly tied prompts 12\% of the time rather than the nominal 5\%, and holding the true difference on only 82\%.

A careful reader might object that the student has not corrected CIs for multiple comparisons. Suppose they do. Under the same \v{S}id\'ak correction \pkg{} applies, the bootstrap separates tied prompts far less often, and a second problem takes its place. Across all 28 intervals at once, at least one of them misses capturing its true mean difference about 90\% of the time---nearly always showing a miscalibrated result.

Now, consider the same data going through \pkg{}:

\begin{lstlisting}[language=Python]
import (*\pkgcode*) as es
evaldata = es.load_from(df)  # prompt, item, score
result = es.compare(evaldata, factors="prompt", score_range=(0, 1))
result.summary()
\end{lstlisting}

\noindent\includegraphics[width=\linewidth]{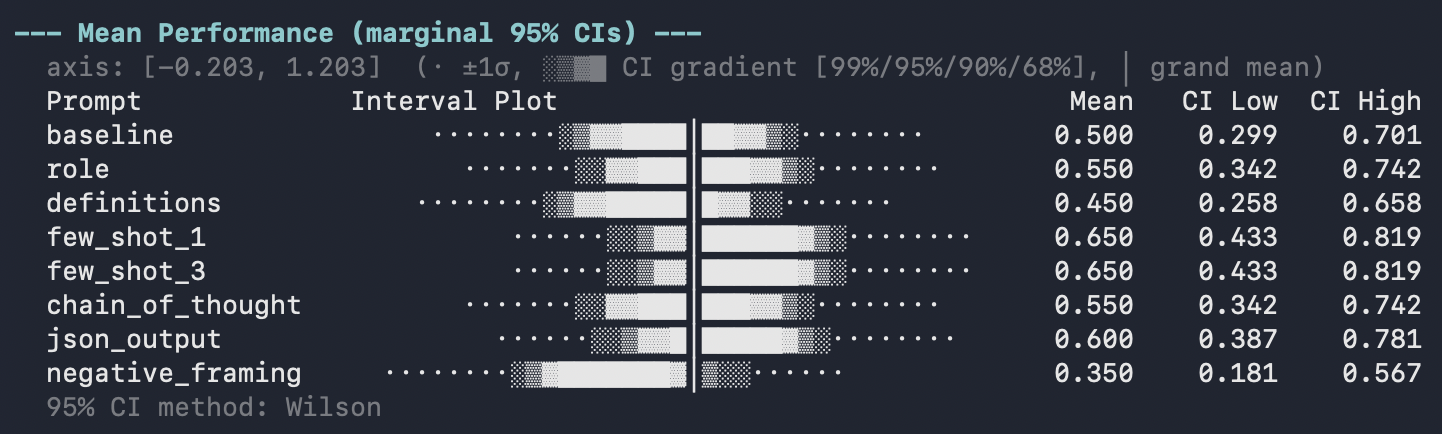}

\pkg{} runs pairwise comparisons over Bonett-Price CIs, \v{S}id\'ak-corrected for FWER control (not shown, since 28 comparisons would take up the page). Compared to the bootstrap's roughly 90\% family-wise miss rate, at least one of its pairwise intervals only misses the true difference 0.6\% of the time. However, at 20 tickets almost nothing is significant (it rules out the worst prompt on 3\% of draws). That is the correct reading of 20 items spread over 28 comparisons. It suggests, however, that roughly 80 tickets would help isolate the data more. If the student scaled up to an 80-item eval, it would rule out the worst prompt 97\% of the time and about 2/7 others. Across 1,500 draws it never discards a truly winning prompt.

This is real data, and \pkg{} is deliberately conservative here. Binary scores are the hardest case for a paired interval, with our simulations (\S\ref{sec:techvalid:main}) showing that most methods lose calibration at small sample sizes, which is why \pkg{} selects the Bonett-Price interval for binary data. The cost is that it will rarely name a winning condition for a 20-item eval set and 28  comparisons. The benefit is that its reported intervals very likely contain the truth---in other words, its output is well-calibrated.

\subsection{What if the metric is an LLM judge?} \label{sec:demo:judge}

With the dialogue component chosen, the student turns to the second LLM component that drafts written feedback for the user after each practice session. They think the feedback content is pretty good, but the style is overly long and sounds like AI slop. %
To address this, they add two dozen style rules to the component's system prompt and call the result \emph{``plain-writing
variant.''} To check whether this skill helps or not, they devise an eval set composed of the transcripts of 60 practice sessions from a public dataset.\footnote{The skill and its 25-rule rubric are real and public~\cite{plainwritingskill}, but the eval data here is synthetic. The response-length confound is simulated explicitly, using the same per-item nuisance-covariate mechanism as the paper's judge-bias sweep (\S\ref{sec:datagen:judgesynth}): length correlates with true quality at $\rho{=}0.3$, baseline sits ${\approx}0.4$\,SD above the mean and plain-writing mode ${\approx}0.4$\,SD below it, and the judge adds $1.3$ points per SD of length to almost exactly cancel the true $1.5$-point gain.} %

They run their eval set through the interface twice, once per system prompt variant, and score every output against the rubric with an LLM judge, yielding a score out of 25 per item per condition. %
Eyeballing a few outputs, they think the judge seems pretty good. They pass the results to \texttt{compare()}:

\begin{lstlisting}[language=Python]
evaldata = es.load_from(df)  # item, condition, score
result = es.compare(
    evaldata, factors="condition", metric="score",
    score_range=(0, 25), design="paired")
\end{lstlisting}

Across their 60-item eval set the judge sees no real difference: plain-writing mode
scores $0.08$ points \emph{lower} on average, with a 95\% CI of $[-1.15, 0.98]$, and
\pkg{} reports the result a tie.

What the student does not know is that their judge is biased in an important way: it rewards length, a well-documented tendency of LLM judges~\cite{zheng2023judging}, and plain-writing mode makes text shorter (the judge's scores correlate $r{=}0.43$ with response length in our constructed setup). %
Catching this unaided would mean suspecting it first, then probing for it. The student has no reason to suspect anything; the feature simply looks like it did not work.

They are disappointed, sure it was helping, but before shelving it they check whether the judge itself is trustworthy, as they heard somewhere that validating a judge against human labels is recommended industry best practice~\cite{yan2024llmevaluator}. Using \pkg{}, they get a simple random sample of 15 items per condition and, blinded to condition, manually label them against their judge rubric:

\begin{lstlisting}[language=bash]
$ (*\pkgcode*) label eval_results.csv --factor condition --metric score \
    --n-lab 15 --interactive
\end{lstlisting}

They then pass the labeled spreadsheet into \texttt{judge\_alignment}:
\begin{lstlisting}[language=Python]
ar = es.judge_alignment(
    evaldata, llm_metric="score", 
    human_groundtruth="human_score",
    selection="random")
\end{lstlisting}

The alignment report is reassuring: absolute agreement ICC(2,1)${=}0.83$, very high correlation. By that standard, there is nothing left to do. But the astute student still passes the alignment data to \texttt{compare()}:

\begin{lstlisting}[language=Python]
result = es.compare(
    evaldata, factors="condition", metric="score",
    score_range=(0, 25), design="paired", 
    alignment={"score": ar})
\end{lstlisting}

The corrected estimate tells a different story. Plain-writing mode is ahead by $1.43$ points (95\% CI $[0.53, 2.32]$), potentially significant.\footnote{The draw shown here is typical: the generative model was fixed in advance, and across 60 random draws the uncorrected judge recovers a mean effect of $+0.24$ against a true $+1.5$ while the PPI-corrected estimate recovers $+1.46$, flipping the verdict in ${\approx}63\%$ of draws.} Correcting for the judge's condition-dependent length bias reversed the verdict, from ``no measurable effect'' to ``the feature helps,'' sparing the student a misleading answer.

Note that the LLM judge here is no less biased than it was: PPI does not need it to be unbiased, since the labeled sample tells the correction how it errs. Nor does PPI require the student to detect how, exactly, the judge fails. What PPI needs is random sampling for human labeling, and for human labels to be good quality. This example also showed a practitioner setting, with an independent rating assumption relaxed; for a scientific setting, we recommend more rigorous collection of the human gold labels.

\subsection{Preventing a spurious finding from reaching publication: Analysis of mixed human-AI judge designs} \label{sec:demo:mixed-rater}

Between-subjects designs are common in human-subjects research, including HCI and UX. \pkg{}' PPI-corrected tests and marginal CIs work for them, and data can be analyzed by setting \texttt{design="unpaired"} on the \texttt{compare()} function. %
However, as we do not validate pairwise CI methods for between-subjects designs in this paper (\S\ref{sec:techvalid:main}), this example reports only the results for PPI tests and point estimates, which \textit{are} well-calibrated. \textbf{This demonstration uses real data.}\footnote{Specifically, we use four topics from the ELLIPSE corpus of argumentative essays by US students learning English~\cite{crossley2023english}, chosen because their trained-rater scores are close to null ($n=209$, $297$, $176$, $178$; true means $3.06$--$3.15$). Each \texttt{human\_label} averages two trained raters scoring a language-proficiency rubric. Six judges (\texttt{gemini-2.5-flash}, \texttt{gpt-4o-mini}, \texttt{nova-lite-v1}, \texttt{command-r7b}, \texttt{llama-3.1-8b}, \texttt{qwen3-30b-a3b}) score at temperature $0$ on a $1$--$9$ scale mapped to the raters' $1$--$5$ half-point lattice; their ensemble departs from the raters mainly by up-rating \emph{Being busy} ($+0.15$) and down-rating \emph{Distance learning} ($-0.06$). All numbers in this section are real; only the narrative is lightly reframed. The script reproducing this finding is available on request.}

Consider a researcher studying online learning platforms, preparing a conference submission. They would like to know if the way students responded to the platform's discussion feature differs by topic. The class split students into cohorts to discuss four topic units across the course---\emph{Being busy}, \emph{Distance learning}, \emph{Impact of technology}, and \emph{Positive attitudes}. Enrollment fluctuations left the cohort sizes unequal, with $860$ written responses in all. 

Grading that many discussion responses by hand is infeasible, so the researcher builds an LLM judge to score each response for writing quality on a $1$-$5$ scale. Following what they have read about good practice, they use a judge ensemble: they run six LLM judges from six providers and average their scores, and calibrate each with human-expert-rated examples, stratified by score level, drawn from a held-out set of responses their platform collected on other topics. In internal testing on a hand-made validation set, they achieve Pearson's $\rho{=}0.76$, and deem their judge well-calibrated.

\paragraph{Without \pkg{}.} The researcher validates their LLM judge in the best way they know how: they and a coauthor each independently score $30$ randomly sampled responses per
condition, blinded to which condition an item came from, average the two ratings per item as ground truth, and compute inter-rater reliability against the panel. The result is quadratic-weighted Cohen's $\kappa$ of $0.66$, comfortably ``substantial
agreement'' by \citet{landis1977measurement}'s standards. Reassured, they run familiar NHST statistical tests over the raw LLM judge scores. A Kruskal-Wallis omnibus rejects the null that the four conditions score equally ($p=1.9\times10^{-4}$), and post-hoc Mann-Whitney $U$ comparisons, Holm-corrected, leave three of six pairs significant, with Distance learning sitting $0.17$ points below Positive attitudes ($p=0.0003$). They
write this up as a finding and submit it for peer review. Reviewers see an inter-rater reliability that looks high and conclude that the LLM judge panel, and therefore the result, can be trusted.

Every one of those $p$-values is wildly wrong. In the real data behind this example there is no significant difference between conditions ($p=0.47$): the effect is an artifact of the judge, and the
$\kappa{=}0.66$ is not evidence to trust the results. %
Nothing in the researcher's procedure was careless. They sampled at random, blinded their raters,
aggregated ratings, reported IRR, and used an ensembled LLM judge with few-shot examples from a held-out validation set. The issue is that they stopped short of propagating human-LLM alignment into downstream inference.

\paragraph{With \pkg{}.} Now suppose the same researcher reaches for \pkg{} at the outset. To help ensure their sample for human labeling is random, they draw it with the CLI's \texttt{label} helper, which samples \emph{within} each condition and so satisfies the missing-completely-at-random assumption PPI requires:\footnote{\pkg{} also checks the labeled subset against the full pool, using the distributional and label-position tests detailed below. These remain warnings, and cannot catch everything.}

\begin{lstlisting}[language=bash]
$ (*\pkgcode*) label discussions.csv --factor unit \
    --metric quality_score --n-lab 30
\end{lstlisting}

\noindent{} This adds a "human\_score" column to their spreadsheet, and marks the randomly-selected cells for human labeling (the rest remain NaN). With some finagling, they upload a blinded version to Google Sheets and have their raters give separate ratings, performing the same two-rater averaging as before. With their gold labels collected and the final spreadsheet filled out, they then compute alignment, declaring how items were selected:

\begin{lstlisting}[language=Python]
al = es.judge_alignment(
    evaldata, llm_metric="quality_score",
    human_groundtruth="human_score", selection="random")
al.summary()
\end{lstlisting}

\begin{lstlisting}[basicstyle=\ttfamily\scriptsize]
Judge alignment report
------------------------------------------------------------
Alignment set  : 120 of 860 items have human labels (14.0%)
Representativeness: labeled items look like the full pool
  Pearson r     0.71  [0.60, 0.79]  large positive correlation
  Spearman r    0.68  [0.56, 0.77]  large positive correlation
  ICC(2,1)      0.67  [0.55, 0.75]  moderate absolute agreement
------------------------------------------------------------
\end{lstlisting}

\noindent The labeled subset seems random overall, with the LLM judge generally agreeing with humans in the pooled IRR. %

Now, the researcher proceeds to the comparison. \textbf{All they need to do is pass the alignment data straight into \texttt{compare()} via a single option, which marks the metric untrusted:}

\begin{lstlisting}[language=Python]
result = es.compare(evaldata, factors="unit", 
    metric="quality_score", design="unpaired", 
    alignment={"quality_score": al},
    score_range=(1, 5))
\end{lstlisting}

The corrected analysis finds nothing. The Kruskal--Wallis omnibus no longer rejects
($p{=}0.31$), no pairwise comparison survives correction, and all four conditions land in a single
statistically indistinguishable band---which is what the ground truth, unknown to the researcher in this story, actually says.\footnote{Across $300$ random draws of which $30$ items per condition get human labels, the draw shown sits at the $27$th percentile of corrected omnibus $p$; its $\kappa$ of $0.66$ is on the high side, at the $76$th percentile, against a median draw's $0.63$, remaining ``substantial agreement.'' The effect was also small even before correction (largest pairwise rank-biserial $-0.21$). Correction did not so much shrink it ($0.024$) as expose its uncertainty.} 

In this alternate universe, both the authors and the published record are spared a misleading result. The researcher writes up the results, reporting the null finding:

\begin{quote}\itshape
Judge-human agreement on the human-labeled calibration set was quadratic-weighted $\kappa{=}0.66$ and raw Spearman $\rho_S{=}0.68$. When running statistics over judge scores, we corrected for judge bias using prediction-powered inference~\textbackslash cite\{...\}. The test-appropriate $\rho^2$ governing the PPI-corrected Kruskal-Wallis test was $\rho^2{=}0.42$, for an estimated effective sample size of $47$ human ratings per condition ($1.6\times$ our labeled $30$). The PPI-Kruskal-Wallis test found no significant difference in discussion quality across the four units ($p{=}0.31$). Post-hoc PPI-corrected Mann-Whitney $U$ tests, FWER-controlled via Shaffer's procedure, likewise found no difference.
\end{quote}

How often would a researcher have been misled without judge bias calibration, and reported a false result? Every time, at least given the same LLM judge and labeling budget they chose for its calibration. Since the $30$ items randomly selected for the human raters can change, let us repeat the study across $300$ random draws of this subset.\footnote{Each draw samples $30$ responses per condition uniformly
at random and re-runs both the corrected analysis and the hand-scored-only baseline on
it. The corpus is close to null but not exactly null---the trained raters put the four
units at $p{=}0.47$---so these are rates at which a finding survives re-labeling,
rather than exact Type-I error rates.} Across draws, quadratic-weighted
Cohen's $\kappa$ stays at a reassuring median of $0.63$, or ``substantial agreement.'' The researcher may continue to be misled. By contrast, \pkg{}' PPI-corrected Kruskal-Wallis test rejects the null only $5.0\%$ of the time (at nominal $\alpha{=}.05$, 15/300), and the PPI-corrected MWU post-hoc tests report at least one significant pair in only 11/300. Since writing up a finding needs both omnibus and pairwise tests to be significant, a true false positive happens only $3.7\%$ of the time had the researcher used \pkg{}. Had they opted for no LLM judges at all, randomly selected $30$ items for human scoring, and run statistics over human scores alone, they would have reported a false positive, coincidentally, also only $3.7\%$ of the time.

\section{Discussion} \label{sec:discussion}

Our work addresses gaps of statistical know-how identified in \S\ref{sec:motivation} and targeted by proposed guidance from the Canadian AI Safety Institute and NIST~\cite{keller2026practices, caisi2026evaluators}, which would be required by researchers attempting to reconcile guidelines on LLM evaluation reporting~\cite{navarro2026reporting, baltes2025guidelines} with calls to quantify uncertainty~\cite{dragicevic2016fair, kaptein2012rethinking, agarwal2021deep}. We provide guidance and validated tests for running statistics over LLM judge scores in mixed human-AI judge designs, and concrete recommendations and tooling for statistical methods in small-sample AI evaluation regimes. These regimes are inter-related, as evals often invoke LLM judges, but also distinct. Researchers and practitioners can use \pkg{} today for common scenarios. Here, we reflect on the broader implications of our work and its limitations.

\subsection{Implications for researchers}

Previously, researchers may have wanted to use LLM judges to score data, but hesitated or, even if using them, wondered how best to trust the results. They now have calibrated options, for a practice that will likely only grow. As \S\ref{sec:intro} noted, some previously published papers in HCI, AI, and NLP run statistics over raw LLM judge scores after reporting IRR alone~(e.g., \cite{gao2025homeworkwars, zhou-etal-2024-real, panda2025accesseval, fanous2025syceval}).\footnote{We reached out to authors of some of these papers, but the underlying data could not be made available to us due to ethics constraints.} We stress that no individual authors nor reviewers are at fault here: the corrective methods were simply not yet well-known in these communities. And, in the much larger LLM-as-a-judge literature in NLP, judge scores can be routinely compared across models with no propagation of judge uncertainty~\cite{lee2026how, fiedler2026bias}. Our concern is rather the practice---recall that, for some common IRR metrics, \textit{higher} IRR is associated with \textit{higher} false-positive risk (\S\ref{sec:ppi-q1}). Now that corrections are available and known, \textbf{LLM-judge-bias-aware inference is not an optional ``nice-to-have''}---it should be an expectation going forward. Whether authors correct via PPI or \pkg{} \textit{specifically} is a different question~\cite{hullman2026human}; alternatives like design-based supervised
learning exist~\cite{egami2023dsl, lee2026how, calderon2025alternative}.

One might still wonder if judge bias could simply be detected and subtracted before running tests. For a mean, that \emph{is} what PPI already does. But PPI also ensures the correction incorporates the uncertainty estimated from the $N_\mathrm{lab}$ items into the variance. A simple subtraction would miss this, as authors would still be treating all $N$ adjusted judge scores as surrogates for human ones, and thus report confidence much narrower than the data supports.

Our small-sample simulations also hold takeaways beyond AI evaluation. Classical NHST $p$-values remained calibrated, if conservative at low $N$, but the bootstrap CI variants we tested (\S\ref{sec:techvalid:main}) failed nominal coverage at small sample sizes, even around $N{=}50$, with the main exception of the smooth bootstrap (which is itself kind of a hack). \textbf{We caution researchers against bootstrap CI estimation in small-sample regimes.} Instead, we point researchers adopting estimation statistics~\cite{dragicevic2016fair, cumming2014new} to our recommendations (Fig.~\ref{fig:ci-decision-tree}).

Although \pkg{} is suited to LLM evals-like data, there is nothing \textit{too} special about its setup. It takes in results, and does not care about data provenance---technically speaking, it works over non-AI evals data, all the same. Comparing 3 interfaces across 24 participants in a within-subjects study is, statistically, the same shape of problem as comparing three models across 24 benchmark items, and \pkg{}' CI method and $p$-value choices remain reasonable, as we validated against a wide range of scenarios. Two caveats apply to user studies specifically---\pkg{} does not yet validate pairwise CI methods for between-subjects comparisons and will refuse to output statistics below its cut-off of $N{=}15$ (\S\ref{sec:philosophy})---but within those bounds, it offers, perhaps unwittingly, a practical toolkit implementing the estimation statistics advocated by \citet{dragicevic2016fair}, albeit with concessions made to NHST by still allowing users to print $p$-values~\cite{masson2023statslator}.

\subsection{Making uncertainty-aware interfaces for AI evaluation possible}

Prior interfaces for AI evaluation do not support statistical inference, despite often operating in small samples and LLM judge regimes (\S\ref{sec:related}); even if designers had wanted to add inference, it would be unclear what methods they should choose. \pkg{} now makes well-calibrated inference possible. For instance, MetricMate~\cite{gebreegziabher2025metricmate} could surface a PPI-corrected estimate whenever a user checks their evaluator against a few human-graded examples; ChainForge~\cite{arawjo2024chainforge} could flag a ``winning'' prompt on $N{=}30$ as statistically indistinguishable from the runner-up; and DocWrangler~\cite{shankar2025docwrangler} could use paired human-LLM judgments to power judge-bias-aware inference over a full annotated document set. Systems adopting \pkg{} can also inherit its gradient-plot visualizations, shown to produce better-calibrated inferential judgments than the error bars and bar charts rampant across AI evaluation~\cite{correll2014error, padilla2022uncertainty}. A natural next step is a graphical interface that helps practitioners plan, run, analyze, and report mixed human-AI judge designs.

\subsection{Responsible setup and statistical analysis of mixed human-AI judge designs}

While \pkg{}' PPI tests enable researchers to run calibrated statistics over LLM judges, they are not a panacea. For PPI-corrected methods to deduce a signal from an \textit{untrustworthy} judge, the smaller subset of human labels must be \textit{trustworthy} themselves. Neither LLM judges nor PPI is a license to abdicate our responsibility to collect quality data from human subjects and design the careful experiments needed to obtain it. The fewer human labels $N_\mathrm{lab}$ feeding a PPI-corrected method, the more critical their quality.

Researchers presented with early versions of this work asked whether our PPI-corrected tests can accommodate disagreeing human raters. In principle, yes: PPI's guarantees only require the labeled subsample be selected randomly and independently, not noise-free raters, and authors can aggregate across multiple independent raters to improve label quality (we leave formal multi-rater extensions to future work; one promising direction is judge validation under rating indeterminacy~\cite{guerdan2025ratingjudge}). But PPI cannot manufacture agreement where none exists, or substitute for careful thought about what is being measured.

Our $\rho^2{\ge}0.4$ rule of thumb is also just that: researchers must interpret ``cost'' and ``gains'' in their own context, since the gain is conditional on the $N_\mathrm{lab}$ one started with. Given 30 human labels and 300 judge scores per condition, alignment at the low end of ``substantial agreement'' ($\rho^2{\approx}0.4$) buys only about 17 more effective human samples---so mixed designs may only be worthwhile with a highly aligned judge or a very large dataset, which cautions against LLM judges where human raters cannot feasibly reach consensus. %

\subsection{PPI tests outside of LLM judge scenarios} \label{sec:discussion:syntheticusers}

Researchers might wonder whether our PPI tests work in ``synthetic users'' regimes~\cite{salminen2025syntheticusers}, outside grading or annotation. PPI reduces variance when the LLM's score covaries with the human's at the individual-unit level, which requires the judge to score a real artifact (essay, transcript, trace) a human also scored. That assumption fails for synthetic users that experience a condition and self-report (e.g., an agent filling out SUS): the agent sees only the condition and produces the same prediction for every participant assigned to it, so the rectifier recovers the classical estimate with no gain. \citet{broska2025mixed} shows PPI \textit{could} apply if each synthetic subject is conditioned on a real participant's profile or trace, but this needs separate validation we leave to future work. Our nonparametric PPI tests may be especially useful here, since LLM judge scores are often discrete or bounded, and many HCI metrics already favor nonparametric tests~\cite{kaptein2010}. 

PPI-corrected tests also apply to \textit{any} noisy predictor, not just AI models~\cite{angelopoulos2023prediction}. Crowdworker ratings are a recurring concern in HCI, less reliable than in-lab or expert raters~\cite{hube2019crowdsourcing, draws2022crowdworkerbias, he2024ifinacrowdsource} and increasingly confounded by unsanctioned AI usage~\cite{veselovsky2025aicrowdworkers}. Prior work seeks to improve the human annotations (e.g., \cite{OPPENHEIMER2009867, hube2019crowdsourcing}), rather adjusting inference for annotation error. Our PPI tests complement this, and could yield calibrated inference over a full crowdsourced set given a random subset rated by trusted experts.

\subsection{Why not Bayesian?}
\label{sec:discussion:bayesian}

While adopting an estimation statistics philosophy~\cite{dragicevic2016fair, cumming2014new}, we ultimately commit to a frequentist framing. It is natural to wonder what a fully Bayesian \pkg{} would look like. Bayesian methods are appealing in some ways: more principled inference in small-$n$ studies~\cite{kay2016researcher}, prior information, and posterior probability statements that map more directly onto how practitioners already (mis)read confidence intervals~\cite{belia2005researchers}. We agree, and believe a Bayesian \pkg{} is an exciting direction for future work. One of our own CI recommendations (Fig.~\ref{fig:ci-decision-tree}) is evidence in favor of Bayesian methods. However, %
going Bayesian comes with its own difficulties: priors can be complex for users to engineer~\cite{jun2026priorweaver}, and some Bayesian intervals we tested erred conservative (Appendix~\ref{app:ci-methods}); many methods also demand heavier computation (e.g., MCMC); and several of our target audiences may expect frequentist output (\S\ref{sec:philosophy}).

\subsection{Limitations}
\label{sec:discussion:limitations}

Notably missing are recommendations for \textit{pairwise} CIs on \textit{unpaired} data ($k$ independent samples, common to between-subjects designs, but uncommon in AI evaluation settings), as well as %
reporting validation of \pkg{}' multi-run CI methods (\S\ref{sec:toolkit}). %
We also focus on means rather than medians, as means dominate AI evaluation and capture tailed behavior that medians mask. Like~\citet{mani2026nofreelunch}, our shrinkage of the power-tuning weight (\S\ref{app:ppi:adaptive}) is also not free: with strongly aligned judges it can cost a few points of power relative to classical PPI++. Future work might add multi-metric analysis, stratified designs, other metrics (F1, win-rates), and PPI corrections for  ART~\cite{wobbrock2011art}, ordinal regression~\cite{likertchi2026}, Chi-squared test, linear mixed models, $e$-values~\cite{csillag2025ppievalues}, or multi-run data. Our simulation harness makes this easier: researchers can hook up new tests rather than reimplement from scratch.

We also did not conduct a user study, which raises an interesting question: even given well-calibrated statistics, would users still ignore uncertainty when making decisions, and how could we improve that? These questions are separate from statistical calibration, but deserve careful study in their own right. Prior work suggests that uncertainty visualizations lead to improved decision-making~\cite{correll2014error, padilla2022uncertainty, kay2016bus}. %

Finally, statistics is a field rife with controversy, exemplified by the heated debates between frequentist and Bayesian methods and NHST versus estimation statistics~\cite{besan2019dichotomous, mayo2022statistical}. Critics might thus question some of \pkg{}' decisions, such as defaulting to empirically-tested yet less familiar CI methods, or still printing $p$-values alongside 95\% CIs. Our stance, like \citet{masson2023statslator}'s Statslator, is to defer to the user's judgement and support both NHST and estimation statistics. (Our PPI corrections also work on CIs; Appendix~\ref{app:ppi:ci-methods}.) While we can, and should, quibble over the specifics, the risk of having \textit{no} uncertainty quantification at all, or running stats over raw LLM judge scores directly---both the status quo in the wild west of AI evaluation---is far too great. We believe reasonable uncertainty quantification, grounded in simulations and prior research, is a good start, and look forward to other work that shares our goal, but differs in philosophy or approach.

\section{AI Usage Acknowledgment} \label{sec:ai}

The PPI-corrected methods in this work were built with the help of Opus~5 and Sonnet~5, with Fable~5 in an advisor role for hard mathematical problems, and validated in our simulation harness on real and synthetic judge data. Plots and simulation code were prepared with the coding help of Claude Code. Appendix~\ref{sec:appendix-rank-based}, which documents the rank-test constructions, was drafted with Fable~5 from the implementation, then checked line by line against the code by Fable~5, by the author, and double-checked by GPT-5.6 Luna and Gemini 3.1 Pro. All other text was written by the author, with minimal AI help for line-editing, clarity, and conciseness. The author takes full responsibility for any errors that may appear. %

\bibliographystyle{ACM-Reference-Format}
\bibliography{sample-base}

\appendix

\setcounter{dbltopnumber}{2}%
\section{Monte Carlo simulations comparing statistical methods in LLM evals-like data settings} \label{appendix:techval}

All tests in this Appendix can be run from the root of the \pkg{} repository (\url{https://github.com/ianarawjo/evalstats}) by running:

\begin{verbatim}
    python -m simulations.harness.cli --official-tests
\end{verbatim}

from the root directory, then selecting the test. Many are resource-intensive and therefore all are parallelized across CPUs; even so, each can take many hours to run. All tests were performed on a MacBook Pro 2024 with 64 GB RAM and an M4 Max chip.

Note that we tried our best to represent all information necessary to validate main text claims in this appendix. Beautiful violin plots for CI and $p$-value methods did not always fit the bill, unfortunately, as they largely repeat information presented in tables. Rather than lose them entirely, we put them in a separate Supplementary Material, available as an ancillary file with this arXiv submission, for readers who are curious. Our tables color-code cells to enhance glance-ability: under-coverage or poor performance is shaded \textcolor{red}{red}, overly conservative results are \textcolor{blue}{blue}, and near-nominal results are left alone.

\subsection{Data distributions and scenario suite} \label{app:scenarios}

\subsubsection{Synthetic data generation} \label{sec:datagen:synth}

For consistency and simplicity, we use the same  synthetic data generation suite across our CI, $p$-value, FWER, and end-to-end \texttt{compare()} simulations (the end-to-end simulations use its standard subset). Table~\ref{tab:synthetic-shapes} summarizes the data generating processes and sweeps. %
For group comparisons ($k{\ge}2$ arms), a further effect size (Cohen's $d$) shifts one or more groups, which is realized in units of the specific shape's  standard deviation at the given ICC (i.e., so that $d{=}0.3$ denotes the same gap for every shape and eval type regardless of its scale or skew). Note that our PPI correction scenarios use a separate generator for judge bias, see \S\ref{appendix:ppi}. %

\begin{table*}[t]
\centering
\small
\renewcommand{\arraystretch}{1.2}
\begin{fitwide}
\begin{tabular}{|p{4.3cm}|>{\raggedright\arraybackslash}p{6.2cm}|>{\raggedright\arraybackslash}p{6.0cm}|}
\hline
\textbf{Eval type} & \textbf{Shape family (parameters)} & \textbf{Rationale} \\
\hline
\multirow{2}{*}{Binary} &
Pass rate $p$: 0.02, 0.05, 0.10, 0.20, 0.30, 0.50, 0.70, 0.80, 0.90, 0.92,
0.95, 0.98 (12 shapes) &
Spans rare-success to near-ceiling regimes, where some intervals can misbehave; $p=0.5$ gives the maximum-variance case. \\
\cline{2-3}
&
10 shapes of lopsided paired agreement, as (P(A-only pass), P(B-only pass), P(both
pass)) triples, at 5 magnitude
levels: extreme (0.001, 0.384, 0.000), ultra (0.000, 0.520, 0.000),
strong (0.020, 0.300, 0.050), sparse (0.001, 0.090, 0.030), and moderate
(0.050, 0.220, 0.150). Each are also mirrored with A/B swapped. To these we add two standalone near-ceiling (0.000, 0.080, 0.900) and near-floor
(0.080, 0.000, 0.020) scenarios. &
\textit{Used only in pairwise comparisons}. Stress-tests paired binary comparisons against highly lopsided disagreement patterns (wins concentrated almost entirely in one direction) that a symmetric
ICC/effect-size sweep would not produce. Mirrored A/B pairs check that paired tests behave symmetrically
regardless of which model wins. \\
\hline
\multirow{3}{*}{Continuous [0,1]} &
Symmetric/boundary Beta$(a,b)$: $(1,1)$ uniform, $(0.5,0.5)$ U-shaped,
$(6,6)$ center-peaked, $(0.6,0.6)$ \& $(0.3,0.3)$ boundary-piled (5 shapes) &
Sweeps mass concentration of the distribution from flat to strongly bimodal, and tests sensitivity to boundary clustering (e.g., many near-perfect or
near-zero scores). \\
\cline{2-3}
&
Skewed Beta$(a,b)$: $(2,8)/(8,2)$ moderate, $(2,5)$ mild,
$(0.35,6)/(6,0.35)$ extreme skew (5 shapes) &
Tests robustness to the skewed score distributions common in LLM benchmarks (e.g., many failures, few high scores). \\
\cline{2-3}
&
Non-Beta: logit-Normal$(-0.35,1.35)$; zero-inflated (70\% zeros +
Beta$(2,4)$); one-inflated (70\% ones + Beta$(4,2)$); two-population
mixture (55\% Beta$(0.5,4)$ / 45\% Beta$(5.5,1.2)$) (4 shapes) &
Stress-tests non-smooth densities that violate  parametric assumptions, such as boundary spikes (refusals, perfect scores) or genuinely distinct sub-populations. \\
\hline
\multirow{3}{*}{Likert (1--5)} &
Location/spread sweep, latent Normal(mean, SD) before rounding, as
(mean, SD) pairs: (3.0, 1.2), (2.2, 1.2), (3.8, 1.2), (3.0, 2.0),
(2.0, 1.1), (4.0, 1.1), (3.0, 0.55), (3.0, 1.4) (8 shapes) &
Covers combinations of central tendency and dispersion on a coarse
ordinal scale. \\
\cline{2-3}
&
Floor/ceiling, latent Normal(mean, SD) before rounding, as (mean, SD)
pairs: (1.8, 1.2) floor, (1.5, 0.65) near-floor, (4.5, 0.65) near-ceiling
(3 shapes) &
Models rubric scores that pile up against a scale endpoint. \\
\cline{2-3}
&
Bimodal: 50/50 mix of Normal$(1.5,0.65)$ and Normal$(4.5,0.65)$, and a more
extreme Normal$(1.3,0.50)$ and Normal$(4.7,0.50)$ variant (2 shapes) &
Models rater disagreement (``love it or hate it'') that violates unimodality. \\
\hline
\multicolumn{3}{|l|}{\textbf{Cross-cutting sweep (applied on top of every shape above)}} \\
\hline
Reliability, pairwise/\allowbreak multiplicity &
ICC $\in \{0.01, 0.30, 0.50, 0.65, 0.75, 0.85, 0.95\}$ &
\textit{Used only in pairwise comparisons.} Weighted towards the empirical per-item ICC distribution measured on
our real Inspect-AI benchmark corpus (median $\approx 0.75$). $0.01$ is kept to test a near-chance extreme. \\
\hline
Effect size &
Cohen's $d \in \{0 \text{ (null)}, 0.2, 0.4\}$ &
Zero effect sizes checks Type-I error control. $d{=}0.2$/$0.4$ cover small-to-moderate real-world effects. These are always in units of each shape's own realized standard deviation. \\
\hline
\end{tabular}
\end{fitwide}
\caption{Overview of our synthetic data generation suite's distributions and how we sweep over them. This suite is shared across all three simulation studies (point estimate CI methods, pairwise comparison CI methods, FWER-correction methods, and pairwise p-value hypothesis tests). Each shape defines a distribution family for one evaluation type. The reliability (ICC) and effect-size (Cohen's $d$) rows vary measurement noise and true differences on top of those shapes for pairwise scenarios ($k\ge2$ groups). Our sweep is not designed to be exhaustive, but to have good coverage of a range of real-world, evals-like scenarios. Full parameter values and generating code are available in the \pkg{} codebase via its simulations harness.}
\label{tab:synthetic-shapes}
\end{table*}

\subsubsection{Real evaluation data sourcing} \label{sec:datagen:real}

To approximate real, small-sample evals datasets, we randomly sample from larger benchmarks and take their population-level point estimate (mean) as the ground truth. We then run the same tests as in the synthetic case. This is not free of problems---the benchmark results can themselves be imperfect predictors of underlying performance---however, they resemble the scenario of making a small eval to approximate larger performance.

Specifically, we compiled our real-data corpora from two sources: (a)~OpenEval~\cite{jiang2026position}, choosing a diversity of 15 unique model~$\times$~benchmark pairs; and (b)~running benchmarks manually via the Inspect AI harness, six models across nine benchmarks (Table~\ref{tab:openeval-real}). %
We report real data results only for binary and continuous data, as we struggled to find large benchmarks that used Likert scores. %

\begin{table*}[htbp]
\centering
\scriptsize
\setlength{\tabcolsep}{3pt}
\caption{The two real evals corpora. OpenEval pairs one benchmark with each model; Inspect AI crosses every model with every benchmark. Unmarked benchmarks are binary, \textsuperscript{c} marks continuous.}
\label{tab:openeval-real}
\begin{minipage}[t]{0.555\linewidth}
{\itshape OpenEval}~\cite{jiang2026position} --- one benchmark per model, single run\\[2pt]
\begin{tabular}{@{}ll@{\hspace{7pt}}ll@{}}
\toprule
Model & Benchmark & Model & Benchmark \\
\midrule
falcon-7b-instruct    & mmlu-pro      & gpt-4.1-mini          & hi-tom \\
gpt-4o                & culturalbench & gemma-3-27b-it        & ifeval\textsuperscript{c} \\
o4-mini               & opentom       & falcon-40b-instruct   & imdb \\
llama-65b             & bbq           & qwen-2.5-72b-instruct & omni-math \\
vicuna-13b-v1.3       & cnndm\textsuperscript{c} & kimi-k2    & salad-bench \\
DeepSeek-V3-0324      & do-not-answer\textsuperscript{c} & llama-2-70b & xsum\textsuperscript{c} \\
DeepSeek-R1           & emobench      & grok-4                & truthfulqa\textsuperscript{c} \\
qwen-3-80b-instruct   & gpqa          & & \\
\bottomrule
\end{tabular}
\end{minipage}\hfill
\begin{minipage}[t]{0.425\linewidth}
{\itshape Inspect AI} --- every model $\times$ every benchmark, 5 runs each at $T{=}0.7$\\[2pt]
\begin{tabular}{@{}l@{\hspace{5pt}}c@{\hspace{5pt}}p{0.42\linewidth}@{}}
\toprule
Model & & Benchmarks \\
\midrule
gemma-3n-e4b-it\textsuperscript{$\dagger$} & \multirow{6}{*}{\normalsize$\times$}
  & \multirow{6}{=}{arc, bbq, boolq, gsm8k, hellaswag, mmlu, piqa, truthfulqa, winogrande} \\
granite-4.1-8b        & & \\
ministral-8b-2512     & & \\
gpt-4o-mini           & & \\
llama-3.1-8b-instruct & & \\
qwen3.5-35b-a3b       & & \\
\bottomrule
\end{tabular}
\end{minipage}
\begin{flushleft}\scriptsize
\textsuperscript{c}Continuous metrics: ROUGE-L (cnndm, xsum), BLEU-max
(truthfulqa), rescaled $[0,6]\!\to\![0,1]$ (do-not-answer).
\textsuperscript{$\dagger$}omits boolq, on which it performed at near-zero accuracy.
\end{flushleft}
\end{table*}

\subsection{Which confidence interval (CI) methods perform optimally in small-samples eval settings?} \label{app:ci-methods}

As \pkg{} foregrounds estimation statistics, 95\% CIs $[L, U]$ are the central measure of uncertainty. The classic measures for CIs are coverage rate and interval width. To this we add minimum coverage (MinCov, the worst single scenario/$n$ cell), Type-I error rate, power, and mean wall-clock runtime per call. We also report the \textbf{interval score}~\cite{gneiting2007strictly}, which combines coverage and width $U{-}L$ into one number for a more glance-able metric of CI ``performance'':
\[
S_{\alpha}(L,U;\theta)=(U-L)+\tfrac{2}{\alpha}(L-\theta)\mathbf{1}({\theta<L})+\tfrac{2}{\alpha}(\theta-U)\mathbf{1}({\theta>U}),
\]

where $\theta$ is the true mean the CI is trying to capture. However, interval score can somewhat hide a bad coverage tail; thus, MinCov and Pen(alty), which reports only the penalty part of this score, are important to consider too.

We consider 95\% CI methods across for the mean point estimate and the mean pairwise difference. We report a summary for each below, with summary tables of descriptive statistics and our overall recommendation.

\subsubsection{Mean point estimates, single-run.} 

Table~\ref{tab:ci_single:sim} shows the Jeffreys and Wilson intervals achieve the best interval scores for synthetic binary data and maintain good coverage across $n$ on synthetic and real data. For numeric data, logit-$t$ consistently holds nominal coverage. Based on these results, from $n=10$ to $100$, we recommend:

\begin{table*}[t]
\centering
\scriptsize
\setlength{\tabcolsep}{2pt}
\begin{fitwide}
\begin{tabular}{lrrrrrrrrrrr!{\hspace{4pt}\vrule\hspace{4pt}}rrrrrrrr}
\toprule
 & \multicolumn{6}{c}{Overall} & \multicolumn{5}{c}{Coverage by $n$} & \multicolumn{8}{c}{Real evals data} \\
\cmidrule(lr){2-7}\cmidrule(lr){8-12}\cmidrule(lr){13-20}
Method & Cov & MinCov & Width & Pen $\downarrow$ & Score $\downarrow$ & T (ms) & $15$ & $30$ & $50$ & $80$ & $100$ & Cov & Width & Score $\downarrow$ & $15$ & $30$ & $50$ & $80$ & $100$ \\
\midrule
\multicolumn{12}{l}{\textit{Binary}} & \multicolumn{8}{c}{} \\
bootstrap & \cellcolor{red!57}.825 & \cellcolor{red!65}.166 & .1834 & .2521 & .4355 & 1.8 & \cellcolor{red!65}.694 & \cellcolor{red!65}.797 & \cellcolor{red!39}.878 & \cellcolor{red!30}.904 & \cellcolor{red!27}.914 & \cellcolor{red!28}.910 & .2412 & .4234 & \cellcolor{red!47}.855 & \cellcolor{red!28}.909 & \cellcolor{red!19}.936 & .949 & .953 \\
bca & \cellcolor{red!57}.824 & \cellcolor{red!65}.166 & .2005 & .2632 & .4637 & 2.3 & \cellcolor{red!65}.678 & \cellcolor{red!65}.792 & \cellcolor{red!38}.879 & \cellcolor{red!27}.912 & \cellcolor{red!22}.929 & \cellcolor{red!30}.905 & .2671 & .4610 & \cellcolor{red!54}.832 & \cellcolor{red!29}.907 & \cellcolor{red!18}.939 & .952 & .959 \\
bayes\_bootstrap & \cellcolor{red!52}.838 & \cellcolor{red!65}.152 & .1819 & .2478 & .4297 & 3.8 & \cellcolor{red!65}.714 & \cellcolor{red!58}.822 & \cellcolor{red!38}.879 & \cellcolor{red!28}.909 & \cellcolor{red!21}.931 & \cellcolor{red!26}.917 & .2369 & .4082 & \cellcolor{red!46}.858 & \cellcolor{red!25}.918 & \cellcolor{red!20}.935 & .951 & .955 \\
smooth\_bootstrap & \cellcolor{red!49}.849 & \cellcolor{red!65}.166 & .2129 & .2369 & .4498 & 4.3 & \cellcolor{red!65}.722 & \cellcolor{red!53}.836 & \cellcolor{red!32}.898 & \cellcolor{red!23}.924 & \cellcolor{red!17}.942 & \cellcolor{red!21}.930 & .2769 & .4342 & \cellcolor{red!39}.876 & \cellcolor{red!20}.933 & \cellcolor{red!16}.947 & .960 & .965 \\
bootstrap\_t & \cellcolor{red!65}.745 & \cellcolor{red!65}.166 & .1837 & .3779 & .5616 & 2.6 & \cellcolor{red!65}.598 & \cellcolor{red!65}.740 & \cellcolor{red!65}.772 & \cellcolor{red!51}.841 & \cellcolor{red!40}.875 & \cellcolor{red!42}.870 & .2520 & .5617 & \cellcolor{red!65}.732 & \cellcolor{red!46}.856 & \cellcolor{red!26}.916 & \cellcolor{red!17}.943 & .955 \\
t\_interval & \cellcolor{red!53}.837 & \cellcolor{red!65}.166 & .2053 & .2489 & .4542 & .074 & \cellcolor{red!65}.722 & \cellcolor{red!53}.835 & \cellcolor{red!39}.878 & \cellcolor{red!30}.903 & \cellcolor{red!27}.914 & \cellcolor{red!26}.915 & .2667 & .4409 & \cellcolor{red!45}.861 & \cellcolor{red!25}.918 & \cellcolor{red!19}.937 & \cellcolor{red!16}.945 & .953 \\
wilson & .953 & \cellcolor{red!29}.908 & .2090 & \underline{.0352} & \underline{.2442} & .048 & .956 & .959 & .952 & .960 & \cellcolor{red!16}.946 & .958 & .2458 & \underline{.2851} & .961 & .956 & .958 & .961 & .962 \\
jeffreys & .956 & \cellcolor{red!35}.890 & .2015 & .0376 & \textbf{.2391} & .088 & .963 & .950 & \cellcolor{red!16}.947 & .954 & .951 & .956 & .2444 & .2891 & .953 & .952 & .958 & .961 & .962 \\
wald & \cellcolor{red!56}.826 & \cellcolor{red!65}.166 & .1817 & .2595 & .4413 & .040 & \cellcolor{red!65}.691 & \cellcolor{red!64}.803 & \cellcolor{red!40}.875 & \cellcolor{red!31}.901 & \cellcolor{red!27}.913 & \cellcolor{red!29}.906 & .2404 & .4312 & \cellcolor{red!49}.849 & \cellcolor{red!29}.907 & \cellcolor{red!22}.929 & \cellcolor{red!18}.940 & .949 \\
clopper\_pearson & \cellcolor{blue!25}.976 & .950 & .2303 & \textbf{.0181} & .2484 & .072 & \cellcolor{blue!42}.986 & \cellcolor{blue!32}.980 & \cellcolor{blue!23}.975 & \cellcolor{blue!17}.971 & .968 & \cellcolor{blue!23}.975 & .2724 & .2937 & \cellcolor{blue!37}.983 & \cellcolor{blue!23}.975 & \cellcolor{blue!17}.971 & \cellcolor{blue!18}.972 & \cellcolor{blue!22}.974 \\
bayes\_indep & .949 & \cellcolor{red!62}.810 & .2094 & .0358 & .2452 & .419 & .967 & \cellcolor{red!17}.944 & .952 & .951 & \cellcolor{red!16}.947 & .959 & .2467 & \textbf{.2849} & .962 & .957 & .957 & .961 & .962 \\
\midrule
\multicolumn{12}{l}{\textit{Continuous}} & \multicolumn{8}{c}{} \\
bootstrap & \cellcolor{red!29}.907 & \cellcolor{red!65}.374 & .1265 & .0603 & .1868 & 2.0 & \cellcolor{red!44}.862 & \cellcolor{red!30}.905 & \cellcolor{red!24}.923 & \cellcolor{red!20}.935 & \cellcolor{red!18}.940 & \cellcolor{red!19}.938 & .1362 & .1834 & \cellcolor{red!26}.916 & \cellcolor{red!21}.932 & \cellcolor{red!16}.945 & .955 & .964 \\
bca & \cellcolor{red!25}.918 & \cellcolor{red!65}.345 & .1297 & .0499 & .1796 & 2.5 & \cellcolor{red!40}.874 & \cellcolor{red!26}.917 & \cellcolor{red!20}.934 & \cellcolor{red!17}.944 & \cellcolor{red!16}.947 & \cellcolor{red!16}.945 & .1395 & .1779 & \cellcolor{red!24}.922 & \cellcolor{red!17}.943 & .952 & .959 & .966 \\
bayes\_bootstrap & \cellcolor{red!29}.907 & \cellcolor{red!65}.370 & .1240 & .0594 & .1834 & 4.1 & \cellcolor{red!45}.859 & \cellcolor{red!29}.906 & \cellcolor{red!23}.926 & \cellcolor{red!19}.937 & \cellcolor{red!17}.942 & \cellcolor{red!18}.939 & .1332 & .1793 & \cellcolor{red!26}.916 & \cellcolor{red!21}.931 & \cellcolor{red!16}.945 & .955 & .965 \\
smooth\_bootstrap & \cellcolor{red!22}.928 & \cellcolor{red!65}.377 & .1437 & .0454 & .1891 & 5.0 & \cellcolor{red!35}.888 & \cellcolor{red!22}.927 & \cellcolor{red!17}.944 & .952 & .954 & .959 & .1545 & .1876 & \cellcolor{red!16}.946 & .956 & .965 & .969 & \cellcolor{blue!23}.975 \\
bootstrap\_t & \cellcolor{red!25}.918 & \cellcolor{red!65}.327 & .1503 & .0474 & .1977 & 2.9 & \cellcolor{red!35}.888 & \cellcolor{red!30}.904 & \cellcolor{red!24}.922 & \cellcolor{red!17}.943 & .949 & .956 & .1567 & .2112 & \cellcolor{red!20}.934 & .962 & .963 & .966 & \cellcolor{blue!18}.972 \\
t\_interval & \cellcolor{red!27}.914 & \cellcolor{red!65}.376 & .1382 & .0532 & .1915 & .085 & \cellcolor{red!38}.879 & \cellcolor{red!27}.912 & \cellcolor{red!22}.927 & \cellcolor{red!19}.937 & \cellcolor{red!18}.940 & \cellcolor{red!15}.948 & .1486 & .1895 & \cellcolor{red!20}.934 & \cellcolor{red!18}.940 & \cellcolor{red!15}.948 & .958 & .966 \\
beta & \cellcolor{red!20}.935 & \cellcolor{red!52}.840 & .1338 & .0400 & .1738 & 3.8 & \cellcolor{red!22}.929 & \cellcolor{red!19}.937 & \cellcolor{red!19}.937 & \cellcolor{red!18}.939 & \cellcolor{red!17}.942 & \cellcolor{red!17}.944 & .1395 & .1753 & \cellcolor{red!22}.927 & \cellcolor{red!19}.936 & \cellcolor{red!16}.947 & .957 & .964 \\
logit\_t & .951 & \cellcolor{red!30}.904 & .1436 & \textbf{.0251} & \underline{.1688} & .098 & .954 & .953 & .951 & .950 & .951 & .961 & .1492 & \textbf{.1683} & .959 & .957 & .959 & .964 & .970 \\
nig & .961 & \cellcolor{red!29}.908 & .1401 & \underline{.0283} & \textbf{.1684} & .058 & .963 & .962 & .961 & .959 & .961 & .963 & .1453 & \underline{.1702} & .960 & .958 & .962 & .967 & \cellcolor{blue!18}.972 \\
el & \cellcolor{red!27}.914 & \cellcolor{red!65}.362 & .1264 & .0536 & .1800 & 1.8 & \cellcolor{red!42}.868 & \cellcolor{red!27}.913 & \cellcolor{red!21}.932 & \cellcolor{red!17}.942 & \cellcolor{red!17}.944 & \cellcolor{red!16}.945 & .1360 & .1763 & \cellcolor{red!22}.927 & \cellcolor{red!18}.941 & .950 & .959 & .967 \\
\midrule
\multicolumn{12}{l}{\textit{Likert}} & \multicolumn{8}{c}{} \\
bootstrap & \cellcolor{red!20}.935 & \cellcolor{red!54}.834 & .6888 & .2182 & .9070 & 2.0 & \cellcolor{red!24}.921 & \cellcolor{red!20}.935 & \cellcolor{red!18}.940 & \cellcolor{red!16}.945 & \cellcolor{red!16}.946 & -- & -- & -- & -- & -- & -- & -- & -- \\
bca & \cellcolor{red!20}.935 & \cellcolor{red!62}.809 & .6914 & .2074 & .8988 & 2.5 & \cellcolor{red!24}.921 & \cellcolor{red!20}.934 & \cellcolor{red!18}.941 & \cellcolor{red!16}.945 & \cellcolor{red!16}.946 & -- & -- & -- & -- & -- & -- & -- & -- \\
bayes\_bootstrap & \cellcolor{red!20}.934 & \cellcolor{red!65}.739 & .6757 & .2208 & .8965 & 4.1 & \cellcolor{red!25}.919 & \cellcolor{red!20}.935 & \cellcolor{red!18}.941 & \cellcolor{red!16}.946 & \cellcolor{red!16}.946 & -- & -- & -- & -- & -- & -- & -- & -- \\
smooth\_bootstrap & .959 & \cellcolor{red!20}.934 & .7809 & .1322 & .9131 & 4.9 & .953 & .960 & .961 & .961 & .961 & -- & -- & -- & -- & -- & -- & -- & -- \\
bootstrap\_t & .956 & \cellcolor{red!34}.892 & .7698 & \underline{.1267} & .8965 & 2.8 & .966 & .959 & .956 & .954 & .953 & -- & -- & -- & -- & -- & -- & -- & -- \\
t\_interval & \cellcolor{red!16}.947 & \cellcolor{red!22}.927 & .7509 & .1679 & .9188 & .083 & \cellcolor{red!16}.945 & \cellcolor{red!16}.947 & \cellcolor{red!16}.946 & .949 & .950 & -- & -- & -- & -- & -- & -- & -- & -- \\
beta & \cellcolor{red!18}.941 & \cellcolor{red!27}.913 & .7041 & .1914 & .8955 & 4.5 & \cellcolor{red!20}.933 & \cellcolor{red!18}.940 & \cellcolor{red!17}.942 & \cellcolor{red!16}.947 & \cellcolor{red!16}.947 & -- & -- & -- & -- & -- & -- & -- & -- \\
logit\_t & .956 & \cellcolor{red!19}.938 & .7359 & \textbf{.1172} & \textbf{.8531} & .096 & .961 & .957 & .953 & .953 & .952 & -- & -- & -- & -- & -- & -- & -- & -- \\
nig & \cellcolor{red!16}.945 & \cellcolor{red!32}.898 & .6934 & .1727 & \underline{.8660} & .055 & \cellcolor{red!17}.943 & \cellcolor{red!17}.944 & \cellcolor{red!16}.945 & .949 & \cellcolor{red!15}.948 & -- & -- & -- & -- & -- & -- & -- & -- \\
el & \cellcolor{red!18}.940 & \cellcolor{red!65}.741 & .6865 & .1930 & .8796 & 1.8 & \cellcolor{red!22}.928 & \cellcolor{red!17}.942 & \cellcolor{red!17}.944 & .949 & .949 & -- & -- & -- & -- & -- & -- & -- & -- \\
\bottomrule
\end{tabular}
\end{fitwide}
\caption{CI methods for mean point estimates, single-run (nominal 95\%, 2000 MC reps
per cell).}
\label{tab:ci_single:sim}
\end{table*}

\begin{itemize}[nosep]
    \item \textbf{Binary}: Wilson interval, as the   Jeffreys interval is Bayesian and we prefer a frequentist method for consistency.
    \item \textbf{Numeric}: Logit-transformed t-interval (which here, has a Clopper-Pearson fallback in zero-width cases.) %
\end{itemize}

In our internal testing, these recommendations remained reasonable past $n{=}100$; thus, \pkg{} currently chooses them for mean point estimates regardless of $n$.

\subsubsection{Pairwise comparisons, single-run.} \label{sec:ci_paired:single}

Table~\ref{tab:ci_paired:single:synth} reports coverage and interval scores for single-run pairwise-difference CIs. (Note that a few of these methods, like Wald, are technically for independent-samples settings, but we included them as a check.) For binary data, the choice is tough: a floored May-Johnson~\cite{may1997confidence} interval is the clear standout across synthetic and real data, with the tightest, on-target CI. However, its minimum coverage is 0.8, which we deem an unacceptable price; thus, we choose the runner-up with good MinCov, Bonett-Price~\cite{bonett2012adjusted}, previously recommended by~\citet{fagerland2014recommended}. For continuous data, logit-$t$ achieves the best interval score, while for Likert, the Bayesian NIG method performs better, offering tighter intervals. Thus, we recommend:

\begin{table*}[t]
\centering
\scriptsize
\setlength{\tabcolsep}{2pt}
\begin{fitwide}
\begin{tabular}{lrrrrrrrrrrrrr!{\hspace{4pt}\vrule\hspace{4pt}}rrrrrrrr}
\toprule
 & \multicolumn{8}{c}{Overall} & \multicolumn{5}{c}{Coverage by $n$} & \multicolumn{8}{c}{Real evals data} \\
\cmidrule(lr){2-9}\cmidrule(lr){10-14}\cmidrule(lr){15-22}
Method & Cov & MinCov & Width & Pen $\downarrow$ & Score $\downarrow$ & T-I & Pow $\uparrow$ & T (ms) & $15$ & $30$ & $50$ & $80$ & $100$ & Cov & Width & Score $\downarrow$ & $15$ & $30$ & $50$ & $80$ & $100$ \\
\midrule
\multicolumn{14}{l}{\textit{Binary}} & \multicolumn{8}{c}{} \\
bootstrap & \cellcolor{red!65}.781 & \cellcolor{red!65}.010 & .1855 & .2016 & .3871 & .023 & .256 & 1.9 & \cellcolor{red!65}.629 & \cellcolor{red!65}.764 & \cellcolor{red!56}.828 & \cellcolor{red!41}.872 & \cellcolor{red!36}.886 & \cellcolor{red!29}.906 & .2818 & .4210 & \cellcolor{red!55}.830 & \cellcolor{red!28}.910 & \cellcolor{red!20}.935 & .949 & .955 \\
bca & \cellcolor{red!65}.730 & \cellcolor{red!65}.010 & .2101 & .2546 & .4648 & .072 & .414 & 2.4 & \cellcolor{red!65}.551 & \cellcolor{red!65}.703 & \cellcolor{red!65}.779 & \cellcolor{red!53}.836 & \cellcolor{red!46}.856 & \cellcolor{red!39}.878 & .2923 & .4846 & \cellcolor{red!65}.782 & \cellcolor{red!39}.876 & \cellcolor{red!26}.915 & \cellcolor{red!19}.937 & \cellcolor{red!16}.945 \\
bayes\_bootstrap & \cellcolor{red!65}.771 & \cellcolor{red!65}.000 & .1871 & .2121 & .3991 & .279 & .527 & 3.5 & \cellcolor{red!65}.611 & \cellcolor{red!65}.753 & \cellcolor{red!59}.819 & \cellcolor{red!43}.865 & \cellcolor{red!38}.879 & \cellcolor{red!32}.897 & .2807 & .4359 & \cellcolor{red!62}.808 & \cellcolor{red!32}.897 & \cellcolor{red!21}.930 & \cellcolor{red!16}.946 & .953 \\
smooth\_bootstrap & \cellcolor{red!57}.824 & \cellcolor{red!65}.030 & .2154 & .1525 & .3679 & .022 & .253 & 4.5 & \cellcolor{red!65}.684 & \cellcolor{red!61}.811 & \cellcolor{red!42}.869 & \cellcolor{red!30}.904 & \cellcolor{red!26}.915 & \cellcolor{red!18}.941 & .3226 & .4069 & \cellcolor{red!33}.895 & \cellcolor{red!15}.948 & .961 & .965 & .969 \\
bootstrap\_t & \cellcolor{red!65}.760 & \cellcolor{red!65}.030 & .1827 & .2195 & .4021 & .086 & .357 & 2.7 & \cellcolor{red!65}.591 & \cellcolor{red!65}.724 & \cellcolor{red!63}.805 & \cellcolor{red!44}.863 & \cellcolor{red!38}.879 & \cellcolor{red!30}.904 & .2945 & .4403 & \cellcolor{red!57}.823 & \cellcolor{red!31}.902 & \cellcolor{red!20}.934 & .949 & .955 \\
newcombe\_mover & \cellcolor{blue!28}.978 & \cellcolor{red!28}.910 & .2740 & .0189 & .2929 & .016 & .229 & .102 & \cellcolor{blue!43}.987 & \cellcolor{blue!33}.981 & \cellcolor{blue!25}.976 & \cellcolor{blue!18}.972 & .969 & .970 & .3217 & .3546 & \cellcolor{blue!25}.976 & \cellcolor{blue!17}.971 & .967 & .965 & .967 \\
mj\_floor & .962 & \cellcolor{red!65}.800 & .2136 & .0311 & \textbf{.2447} & .017 & .234 & .082 & \cellcolor{blue!22}.974 & .964 & .958 & .957 & .955 & .957 & .2818 & \textbf{.3304} & .955 & .956 & .955 & .958 & .961 \\
tango\_scc & \cellcolor{blue!25}.976 & \cellcolor{red!29}.907 & .2909 & .0204 & .3112 & .015 & .225 & .147 & \cellcolor{blue!52}.992 & \cellcolor{blue!35}.982 & \cellcolor{blue!17}.971 & .965 & .961 & .969 & .3396 & .3714 & \cellcolor{blue!38}.984 & .969 & .960 & .962 & .964 \\
bayes\_indep\_comp & \cellcolor{blue!48}.990 & \cellcolor{red!29}.907 & .3119 & \textbf{.0098} & .3216 & .007 & .146 & .713 & \cellcolor{blue!53}.993 & \cellcolor{blue!50}.991 & \cellcolor{blue!47}.989 & \cellcolor{blue!45}.988 & \cellcolor{blue!43}.987 & \cellcolor{blue!40}.985 & .3587 & .3766 & \cellcolor{blue!38}.984 & \cellcolor{blue!38}.984 & \cellcolor{blue!38}.984 & \cellcolor{blue!40}.985 & \cellcolor{blue!42}.986 \\
bayes\_paired\_comp & \cellcolor{blue!22}.974 & \cellcolor{red!35}.890 & .2667 & .0229 & .2897 & .023 & .244 & 5.3 & \cellcolor{blue!42}.986 & \cellcolor{blue!32}.980 & \cellcolor{blue!20}.973 & .964 & .957 & .965 & .3128 & .3519 & \cellcolor{blue!17}.971 & .966 & .961 & .961 & .962 \\
wald\_indep & \cellcolor{red!26}.917 & \cellcolor{red!65}.150 & .2829 & .0730 & .3559 & .011 & .165 & .072 & \cellcolor{red!57}.824 & \cellcolor{red!26}.916 & .951 & .970 & \cellcolor{blue!25}.976 & .969 & .3583 & .4080 & \cellcolor{red!16}.945 & \cellcolor{blue!22}.974 & \cellcolor{blue!32}.980 & \cellcolor{blue!37}.983 & \cellcolor{blue!40}.985 \\
tango\_exact & \cellcolor{blue!17}.971 & \cellcolor{red!30}.903 & .2799 & .0241 & .3040 & .023 & .258 & .112 & \cellcolor{blue!48}.990 & \cellcolor{blue!25}.976 & .966 & .961 & .957 & .964 & .3299 & .3673 & \cellcolor{blue!30}.979 & .961 & .957 & .959 & .960 \\
mj\_unfloored & \cellcolor{red!59}.817 & \cellcolor{red!65}.030 & .1803 & .1585 & .3389 & .023 & .258 & .053 & \cellcolor{red!65}.683 & \cellcolor{red!63}.807 & \cellcolor{red!45}.860 & \cellcolor{red!33}.896 & \cellcolor{red!29}.907 & \cellcolor{red!21}.930 & .2706 & .3662 & \cellcolor{red!37}.884 & \cellcolor{red!20}.933 & \cellcolor{red!15}.948 & .954 & .959 \\
bonett\_price & \cellcolor{blue!30}.979 & \cellcolor{red!26}.917 & .2609 & \underline{.0169} & \underline{.2778} & .012 & .204 & .067 & \cellcolor{blue!52}.992 & \cellcolor{blue!37}.983 & \cellcolor{blue!23}.975 & .970 & .967 & \cellcolor{blue!18}.972 & .3234 & \underline{.3513} & \cellcolor{blue!35}.982 & \cellcolor{blue!23}.975 & .969 & .965 & .968 \\
\midrule
\multicolumn{14}{l}{\textit{Continuous}} & \multicolumn{8}{c}{} \\
bootstrap & \cellcolor{red!20}.934 & \cellcolor{red!57}.823 & .1096 & .0351 & .1446 & .068 & .713 & 2.0 & \cellcolor{red!26}.917 & \cellcolor{red!20}.935 & \cellcolor{red!17}.942 & \cellcolor{red!17}.944 & \cellcolor{red!16}.945 & \cellcolor{red!19}.938 & .1540 & .2015 & \cellcolor{red!29}.906 & \cellcolor{red!21}.931 & .955 & .959 & .961 \\
bca & \cellcolor{red!21}.931 & \cellcolor{red!65}.787 & .1104 & .0383 & .1488 & .074 & .720 & 2.5 & \cellcolor{red!28}.911 & \cellcolor{red!21}.932 & \cellcolor{red!18}.940 & \cellcolor{red!17}.943 & \cellcolor{red!16}.945 & \cellcolor{red!20}.935 & .1564 & .2039 & \cellcolor{red!31}.902 & \cellcolor{red!21}.931 & .951 & .960 & .960 \\
bayes\_bootstrap & \cellcolor{red!22}.929 & \cellcolor{red!65}.783 & .1080 & .0397 & .1477 & .075 & .723 & 3.6 & \cellcolor{red!29}.906 & \cellcolor{red!21}.931 & \cellcolor{red!18}.940 & \cellcolor{red!17}.943 & \cellcolor{red!17}.944 & \cellcolor{red!20}.933 & .1516 & .2030 & \cellcolor{red!32}.898 & \cellcolor{red!22}.928 & .951 & .958 & .961 \\
smooth\_bootstrap & .958 & \cellcolor{red!42}.870 & .1240 & \underline{.0195} & .1435 & .042 & .665 & 4.9 & .953 & .961 & .963 & .961 & .962 & .960 & .1746 & .2015 & \cellcolor{red!17}.944 & .954 & \cellcolor{blue!22}.974 & \cellcolor{blue!27}.977 & \cellcolor{blue!18}.972 \\
bootstrap\_t & \cellcolor{red!16}.946 & \cellcolor{red!47}.853 & .1279 & .0269 & .1548 & .059 & .693 & 2.8 & \cellcolor{red!17}.942 & \cellcolor{red!16}.947 & .950 & .949 & .949 & .952 & .1753 & .2066 & \cellcolor{red!18}.941 & \cellcolor{red!16}.946 & .962 & .965 & .964 \\
t\_interval & .949 & \cellcolor{red!43}.867 & .1191 & .0227 & .1418 & .050 & .680 & .087 & .949 & .950 & .951 & .950 & .950 & .951 & .1678 & .1999 & \cellcolor{red!19}.937 & \cellcolor{red!17}.944 & .961 & .964 & .965 \\
logit\_t & .950 & \cellcolor{red!43}.867 & .1181 & .0209 & \underline{.1389} & .049 & .677 & .083 & .951 & .951 & .951 & .950 & .950 & .954 & .1672 & \underline{.1946} & \cellcolor{red!16}.945 & \cellcolor{red!16}.946 & .963 & .967 & .965 \\
nig & .967 & \cellcolor{red!44}.863 & .1193 & .0227 & .1420 & .032 & .594 & .056 & .970 & .969 & .968 & .964 & .964 & .957 & .1596 & \textbf{.1914} & .949 & .953 & .965 & .968 & .963 \\
el & \cellcolor{red!20}.935 & \cellcolor{red!65}.787 & .1119 & .0360 & .1479 & .068 & .711 & 1.9 & \cellcolor{red!26}.916 & \cellcolor{red!18}.939 & \cellcolor{red!16}.945 & \cellcolor{red!16}.947 & \cellcolor{red!16}.947 & \cellcolor{red!18}.939 & .1565 & .2025 & \cellcolor{red!29}.906 & \cellcolor{red!19}.936 & .954 & .962 & .962 \\
logit\_t\_dither & .950 & \cellcolor{red!43}.867 & .1180 & .0209 & \textbf{.1389} & .049 & .677 & .093 & .951 & .951 & .951 & .950 & .950 & .952 & .1652 & .1947 & \cellcolor{red!18}.939 & \cellcolor{red!16}.945 & .961 & .966 & .963 \\
smooth\_bootstrap\_dither & .958 & \cellcolor{red!42}.870 & .1239 & \textbf{.0195} & .1434 & .042 & .665 & 4.8 & .953 & .961 & .963 & .961 & .962 & .960 & .1727 & .1990 & \cellcolor{red!17}.943 & .957 & \cellcolor{blue!17}.971 & \cellcolor{blue!25}.976 & \cellcolor{blue!20}.973 \\
\midrule
\multicolumn{14}{l}{\textit{Likert}} & \multicolumn{8}{c}{} \\
bootstrap & \cellcolor{red!21}.932 & \cellcolor{red!65}.703 & .4980 & .1767 & .6746 & .048 & .609 & 2.1 & \cellcolor{red!27}.913 & \cellcolor{red!21}.930 & \cellcolor{red!18}.939 & \cellcolor{red!17}.944 & \cellcolor{red!16}.946 & -- & -- & -- & -- & -- & -- & -- & -- \\
bca & \cellcolor{red!24}.923 & \cellcolor{red!65}.563 & .5064 & .2098 & .7163 & .069 & .683 & 2.6 & \cellcolor{red!34}.893 & \cellcolor{red!22}.927 & \cellcolor{red!20}.934 & \cellcolor{red!18}.940 & \cellcolor{red!17}.943 & -- & -- & -- & -- & -- & -- & -- & -- \\
bayes\_bootstrap & \cellcolor{red!21}.931 & \cellcolor{red!65}.687 & .4913 & .1865 & .6778 & .102 & .672 & 3.7 & \cellcolor{red!28}.911 & \cellcolor{red!21}.930 & \cellcolor{red!18}.939 & \cellcolor{red!17}.944 & \cellcolor{red!16}.946 & -- & -- & -- & -- & -- & -- & -- & -- \\
smooth\_bootstrap & .957 & \cellcolor{red!65}.703 & .5653 & .1029 & .6682 & .040 & .586 & 5.1 & .950 & .960 & .961 & .962 & .963 & -- & -- & -- & -- & -- & -- & -- & -- \\
bootstrap\_t & \cellcolor{red!19}.938 & \cellcolor{red!65}.373 & .5469 & .1643 & .7112 & .064 & .630 & 2.9 & \cellcolor{red!24}.923 & \cellcolor{red!16}.945 & .949 & .951 & .951 & -- & -- & -- & -- & -- & -- & -- & -- \\
t\_interval & \cellcolor{red!16}.947 & \cellcolor{red!65}.740 & .5433 & .1195 & .6628 & .047 & .603 & .091 & \cellcolor{red!17}.942 & \cellcolor{red!15}.948 & .949 & .951 & .951 & -- & -- & -- & -- & -- & -- & -- & -- \\
logit\_t & .950 & \cellcolor{red!35}.890 & .5458 & \textbf{.1006} & \underline{.6464} & .046 & .601 & .086 & .949 & .949 & .949 & .951 & .951 & -- & -- & -- & -- & -- & -- & -- & -- \\
nig & .952 & \cellcolor{red!41}.873 & .5152 & .1032 & \textbf{.6184} & .045 & .598 & .059 & .953 & .950 & .950 & .951 & .951 & -- & -- & -- & -- & -- & -- & -- & -- \\
el & \cellcolor{red!19}.937 & \cellcolor{red!65}.687 & .5077 & .1702 & .6779 & .097 & .661 & 1.9 & \cellcolor{red!25}.920 & \cellcolor{red!18}.939 & \cellcolor{red!17}.944 & \cellcolor{red!15}.948 & .949 & -- & -- & -- & -- & -- & -- & -- & -- \\
logit\_t\_dither & \cellcolor{red!15}.948 & \cellcolor{red!35}.890 & .5802 & .1156 & .6958 & .049 & .541 & .097 & \cellcolor{red!15}.948 & \cellcolor{red!16}.947 & \cellcolor{red!15}.948 & \cellcolor{red!15}.948 & \cellcolor{red!16}.947 & -- & -- & -- & -- & -- & -- & -- & -- \\
smooth\_bootstrap\_dither & .957 & \cellcolor{red!38}.880 & .6066 & \underline{.1011} & .7077 & .041 & .521 & 5.0 & .951 & .958 & .960 & .959 & .960 & -- & -- & -- & -- & -- & -- & -- & -- \\
\bottomrule
\end{tabular}
\end{fitwide}
\caption{CI methods for pairwise comparisons, single-run (nominal 95\%, 300 MC reps per cell).}
\label{tab:ci_paired:single:synth}
\end{table*}

\begin{itemize}
    \item \textbf{Binary, pairwise, single-run}:  \citet{bonett2012adjusted}. It never drops below .90 coverage in any cell on either synthetic or real data, keeps pooled Type-I error under nominal, and is extremely fast.
    \item \textbf{Continuous, pairwise, single-run}: Logit-transformed t-interval (with Clopper-Pearson fallback). %
    \item \textbf{Likert / discrete, pairwise, single-run}: Bayesian Normal-Inverse Gamma (NIG). Specifically, we use the weakly-informed prior: $m_0{=}0.5$, $\kappa_0{=}1$, $\alpha_0{=}2$, $\beta_0{=}0.0625/4$.
\end{itemize}

\subsection{Which p-value methods for pairwise comparisons perform optimally in small-samples eval settings?} \label{app:pvalue:pairwise}

We evaluate hypothesis tests for a single pairwise comparison, and consider them across the same three data types and our synthetic and real data suite. For synthetic, we sweep a null condition (no true difference between A and B, used to measure Type-I error) and two effect sizes (Cohen's $d \in \{0.2, 0.4\}$). For our real eval corpus (\S\ref{sec:datagen:real}), we draw genuine A-vs-B differences for power, and build the null for Type-I error by randomly swapping each item's A and B scores. Our metrics here are Type I error, power (ability to detect a real effect), and the Max Type I error (the worst error across the entire sweep, analogous to MinCov in our CI simulations).

Table~\ref{tab:pvalues:pairwise:synth} shows the results. Although there is not a clear standout, Wilcoxon signed-rank and a classic paired t-test maintain good control and power. Of resampling-based methods, only the smooth bootstrap and the permutation test stayed near or below nominal across data types and sizes. Based on these results, we recommend:

\begin{table*}[t]
\centering
\scriptsize
\setlength{\tabcolsep}{2pt}
\begin{tabular}{lrrrrrrrr!{\hspace{4pt}\vrule\hspace{4pt}}rrrrrr}
\toprule
 & \multicolumn{3}{c}{Overall} & \multicolumn{5}{c}{Type-I error by $n$} & \multicolumn{6}{c}{Real evals data} \\
\cmidrule(lr){2-4}\cmidrule(lr){5-9}\cmidrule(lr){10-15}
Method & Type-I & Max & Power & $20$ & $30$ & $50$ & $75$ & $100$ & Type-I & Max & Power & $20$ & $50$ & $100$ \\
\midrule
\multicolumn{9}{l}{\textit{Binary}} & \multicolumn{6}{c}{} \\
bootstrap & \cellcolor{blue!17}.029 & .110 & .370 & \cellcolor{blue!33}.019 & \cellcolor{blue!23}.025 & \cellcolor{blue!22}.026 & .040 & .040 & .047 & .127 & .368 & .037 & .045 & .049 \\
bayes\_bootstrap & .043 & .143 & \textbf{.408} & .047 & .041 & .043 & .046 & .046 & \cellcolor{red!19}.064 & .173 & \textbf{.402} & \cellcolor{red!23}.074 & \cellcolor{red!18}.059 & \cellcolor{red!17}.056 \\
smooth\_bootstrap & \cellcolor{blue!32}.020 & .070 & .342 & \cellcolor{blue!43}.013 & \cellcolor{blue!37}.017 & \cellcolor{blue!28}.022 & \cellcolor{blue!17}.029 & .030 & .033 & .093 & .341 & \cellcolor{blue!18}.028 & .035 & .040 \\
bootstrap\_t & \cellcolor{red!19}.062 & .190 & \underline{.374} & .050 & .033 & \cellcolor{red!23}.076 & \cellcolor{red!21}.069 & \cellcolor{red!20}.065 & \cellcolor{red!18}.059 & .213 & \underline{.368} & \cellcolor{red!15}.052 & .051 & .050 \\
mcnemar & \cellcolor{blue!48}.010 & .053 & .284 & \cellcolor{blue!60}.003 & \cellcolor{blue!53}.007 & \cellcolor{blue!47}.011 & \cellcolor{blue!35}.018 & \cellcolor{blue!35}.018 & \cellcolor{blue!37}.017 & .063 & .290 & \cellcolor{blue!52}.008 & \cellcolor{blue!30}.021 & .030 \\
mcnemar\_midp & \cellcolor{blue!33}.019 & .077 & .332 & \cellcolor{blue!50}.009 & \cellcolor{blue!40}.015 & \cellcolor{blue!27}.023 & .030 & .032 & .032 & .087 & .335 & \cellcolor{blue!30}.021 & .039 & .047 \\
permutation & \cellcolor{blue!48}.010 & .053 & .284 & \cellcolor{blue!60}.003 & \cellcolor{blue!53}.007 & \cellcolor{blue!47}.011 & \cellcolor{blue!35}.018 & \cellcolor{blue!35}.018 & \cellcolor{blue!37}.017 & .063 & .290 & \cellcolor{blue!52}.008 & \cellcolor{blue!30}.021 & .030 \\
sign\_test & \cellcolor{blue!48}.010 & .053 & .284 & \cellcolor{blue!60}.003 & \cellcolor{blue!53}.007 & \cellcolor{blue!47}.011 & \cellcolor{blue!35}.018 & \cellcolor{blue!35}.018 & \cellcolor{blue!37}.017 & .063 & .290 & \cellcolor{blue!52}.008 & \cellcolor{blue!30}.021 & .030 \\
bayes\_binary & .042 & .193 & .374 & \cellcolor{blue!30}.021 & \cellcolor{blue!23}.025 & .038 & \cellcolor{red!19}.064 & \cellcolor{red!29}.094 & .051 & .307 & .364 & .040 & \cellcolor{red!15}.052 & \cellcolor{red!23}.075 \\
wilcoxon & \cellcolor{blue!20}.027 & .083 & .354 & \cellcolor{blue!33}.019 & \cellcolor{blue!22}.026 & .034 & .040 & .042 & .038 & .090 & .345 & .036 & .047 & .051 \\
paired\_t & \cellcolor{blue!18}.028 & .083 & .369 & \cellcolor{blue!33}.019 & \cellcolor{blue!22}.026 & .034 & .040 & .042 & .042 & .090 & .359 & .037 & .047 & .051 \\
\midrule
\multicolumn{9}{l}{\textit{Continuous}} & \multicolumn{6}{c}{} \\
bootstrap & \cellcolor{red!20}.066 & .133 & \underline{.581} & \cellcolor{red!21}.070 & \cellcolor{red!19}.063 & \cellcolor{red!17}.056 & \cellcolor{red!16}.055 & \cellcolor{red!16}.055 & \cellcolor{red!18}.061 & .100 & .754 & \cellcolor{red!21}.070 & \cellcolor{red!15}.052 & .044 \\
bayes\_bootstrap & \cellcolor{red!21}.070 & .147 & \textbf{.588} & \cellcolor{red!23}.074 & \cellcolor{red!20}.065 & \cellcolor{red!17}.058 & \cellcolor{red!17}.056 & \cellcolor{red!17}.056 & \cellcolor{red!20}.066 & .123 & \textbf{.763} & \cellcolor{red!23}.076 & \cellcolor{red!16}.054 & .043 \\
smooth\_bootstrap & .040 & .083 & .520 & .040 & .039 & .037 & .039 & .039 & .038 & .070 & .716 & .042 & .034 & .031 \\
permutation & .048 & .093 & .541 & .050 & .050 & .049 & .050 & .051 & .045 & .073 & .724 & .048 & .048 & .041 \\
sign\_test & .032 & .070 & .439 & \cellcolor{blue!18}.028 & \cellcolor{blue!17}.029 & .034 & .039 & .039 & \cellcolor{blue!18}.028 & .050 & .635 & \cellcolor{blue!20}.027 & .032 & .030 \\
wilcoxon & .048 & .097 & .537 & .048 & .050 & .050 & .050 & .051 & .046 & .070 & .696 & .047 & .049 & .041 \\
paired\_t & .049 & .097 & .542 & .049 & .050 & .048 & .050 & \cellcolor{red!15}.052 & .047 & .070 & .729 & .051 & .046 & .041 \\
\midrule
\multicolumn{9}{l}{\textit{Likert}} & \multicolumn{6}{c}{} \\
bootstrap & \cellcolor{red!17}.057 & .127 & .515 & \cellcolor{red!17}.056 & \cellcolor{red!17}.058 & \cellcolor{red!15}.052 & \cellcolor{red!15}.052 & .049 & -- & -- & -- & -- & -- & -- \\
bca & \cellcolor{red!17}.057 & .127 & \underline{.515} & \cellcolor{red!17}.056 & \cellcolor{red!17}.058 & \cellcolor{red!15}.052 & \cellcolor{red!15}.052 & .049 & -- & -- & -- & -- & -- & -- \\
bayes\_bootstrap & \cellcolor{red!22}.071 & .167 & \textbf{.543} & \cellcolor{red!24}.078 & \cellcolor{red!21}.068 & \cellcolor{red!17}.057 & \cellcolor{red!16}.055 & \cellcolor{red!16}.053 & -- & -- & -- & -- & -- & -- \\
smooth\_bootstrap & .039 & .103 & .471 & .039 & .040 & .038 & .035 & .037 & -- & -- & -- & -- & -- & -- \\
permutation & \cellcolor{blue!20}.027 & .080 & .432 & \cellcolor{blue!28}.022 & \cellcolor{blue!18}.028 & .032 & .034 & .037 & -- & -- & -- & -- & -- & -- \\
sign\_test & \cellcolor{blue!23}.025 & .063 & .408 & \cellcolor{blue!33}.019 & \cellcolor{blue!22}.026 & \cellcolor{blue!17}.029 & .031 & .034 & -- & -- & -- & -- & -- & -- \\
wilcoxon & .041 & .087 & .472 & .045 & .047 & .048 & .049 & .049 & -- & -- & -- & -- & -- & -- \\
paired\_t & .047 & .090 & .492 & .049 & .049 & .048 & .049 & .049 & -- & -- & -- & -- & -- & -- \\
\bottomrule
\end{tabular}
\caption{Pairwise p-value methods (nominal $\alpha{=}0.05$, 300 MC reps per cell), on synthetic and real evals data.}
\label{tab:pvalues:pairwise:synth}
\end{table*}

\begin{itemize}
    \item \textbf{Numeric data (continuous, Likert)}: Wilcoxon signed-ranks, for three reasons: it is slightly more conservative at low $n$, where evals likely suffer construct validity issues anyway; it makes no parametric assumptions; and it is already well-known as a pairwise method in many scientific fields.
    \item \textbf{Binary data}: McNemar's mid-$p$, aligned with \citet{fagerland2014recommended}. Oddly enough, Wilcoxon appears admissible and slightly more powerful. However, we chose mid-$p$ for two practical reasons: 1) when Wilcoxon was placed beside the Bonett-Price CI interval in \pkg{}, it disagrees with significance roughly three times as often, which would hand users inconsistent answers; and 2) Wilcoxon may raise concerns to stakeholders, as it is technically not meant for binary data.
\end{itemize}

\subsection{Which FWER correction methods perform optimally for CIs and p-values?} \label{sec:fwer}

Eval statistics require comparing more than two conditions at once (many models, prompts, or agents), which raises the multiple comparisons problem. Because evals are repeated-measures setups over shared items, pairwise test statistics are correlated rather than independent, the situation that resampling-based corrections like Westfall-Young and Romano-Wolf exploit~\cite{calonico2025beyond}. We also include Friedman-Nemenyi post-hoc, the repeated-comparison design of older ML benchmarking practice~\cite{demvsar2006statistical}.

We evaluate methods that control the \emph{family-wise error rate} (FWER, the probability of any false positive across all comparisons), both for confidence intervals widened to hold \emph{simultaneously} across comparisons (``simultaneous CIs'') and for p-values, across binary, continuous, and Likert 1-5 data, sweeping $k \in \{3, 5, 10, 20\}$ groups. For CIs we report \textbf{family-wise coverage}, the proportion of replicates in which \emph{every} pairwise CI contains its true difference (ideal $1-\alpha$); for p-values we report the \textbf{FWER} itself (ideal $\alpha$) and \textbf{best-arm selection power}, the proportion in which the true best arm is marked significantly different from every other. Width and run time are as defined above. Both checks are also validated on real data (\S\ref{sec:datagen:real}), with $k$ limited by the number of models available per benchmark.

\paragraph{Simultaneous CIs.} \label{sec:simult_cis}

We compare four methods: Bonferroni, \v{S}id\'ak, a joint-bootstrap that computes an effective alpha for the underlying CI method (Supplementary~\S{}S5), and the studentized bootstrap max-T~\cite{goeman2014multiple}, included because it is recommended for high-covariance, many-comparison settings~\cite{rink2025post, goeman2014multiple, bretz2016multiple}. For the first three, the underlying per-comparison CI is whichever method \pkg{} recommends for that eval type (Bonett-Price for binary, logit-$t$ for continuous, Normal-Inverse-Gamma for Likert), so these sweeps widen exactly the interval a user would otherwise see; max-T instead requires a bootstrap pairwise CI.

Tables~\ref{tab:fwer:simult:synth} and \ref{tab:fwer:simult:real} report the synthetic and real data cases, respectively. max-T, whose parametric single-step form is the default in the \texttt{multcomp} R package~\cite{bretz2016multiple} (we test a bootstrap version), struggles with discrete or bounded data at small $n$ and yields no  performance gains over the simpler \v{S}id\'ak correction in small-sample regimes (though around $n{=}200$, max-T does eventually yield better on-target coverage, as expected). Thus:

\begin{table*}[t]
\centering
\scriptsize
\setlength{\tabcolsep}{2pt}
\begin{fitwide}
\begin{tabular}{lrrrrrrrrrrrrrrrrr}
\toprule
CI method & Cov(null) & MinCov(null) & Width(null) & Pen(null) & Score(null) & Cov(alt) & MinCov(alt) & Width(alt) & Pen(alt) & Score(alt) & Type & n=15 & n=30 & n=50 & n=100 & k=3 & k=20 \\
\midrule
bonferroni (bin) & \cellcolor{blue!30}0.979 & \cellcolor{red!23}0.923 & 0.3738 & 0.0136 & 0.3873 & \cellcolor{blue!24}0.975 & \cellcolor{red!27}0.913 & 0.3745 & 0.0154 & 0.3898 & bin & \cellcolor{blue!45}0.993 & \cellcolor{blue!42}0.986 & \cellcolor{blue!30}0.979 & \cellcolor{blue!18}0.972 & 0.970 & \cellcolor{blue!38}0.984 \\
max\_t (bin) & 0.954 & \cellcolor{red!65}0.450 & 0.3081 & 0.0205 & \textbf{0.3285} & \cellcolor{red!43}0.864 & \cellcolor{red!65}0.033 & 0.3120 & 0.1311 & 0.4431 & bin & \cellcolor{red!65}0.780 & \cellcolor{red!37}0.882 & \cellcolor{red!22}0.926 & 0.953 & \cellcolor{red!29}0.905 & \cellcolor{red!29}0.906 \\
sidak (bin) & \cellcolor{blue!33}0.981 & \cellcolor{red!23}0.923 & 0.3559 & 0.0109 & 0.3668 & \cellcolor{blue!33}0.981 & \cellcolor{red!22}0.927 & 0.3570 & 0.0117 & \underline{0.3687} & bin & \cellcolor{blue!45}0.995 & \cellcolor{blue!45}0.989 & \cellcolor{blue!36}0.982 & \cellcolor{blue!28}0.978 & \cellcolor{blue!27}0.977 & \cellcolor{blue!39}0.984 \\
boot (bin) & 0.961 & \cellcolor{red!65}0.500 & 0.3288 & 0.0157 & \underline{0.3445} & 0.962 & \cellcolor{red!65}0.483 & 0.3304 & 0.0167 & \textbf{0.3471} & bin & \cellcolor{red!19}0.936 & 0.967 & \cellcolor{blue!21}0.974 & \cellcolor{blue!15}0.970 & \cellcolor{red!16}0.944 & \cellcolor{blue!20}0.973 \\
\midrule
bonferroni (cont) & 0.965 & \cellcolor{red!22}0.927 & 0.2331 & 0.0134 & 0.2465 & 0.964 & \cellcolor{red!24}0.920 & 0.2332 & 0.0150 & 0.2482 & cont & \cellcolor{blue!20}0.973 & 0.965 & 0.962 & 0.963 & 0.957 & \cellcolor{blue!18}0.972 \\
max\_t (cont) & 0.961 & \cellcolor{red!25}0.917 & 0.2519 & 0.0136 & 0.2655 & 0.960 & \cellcolor{red!25}0.917 & 0.2559 & 0.0148 & 0.2708 & cont & \cellcolor{blue!44}0.987 & 0.967 & 0.956 & 0.953 & 0.955 & 0.966 \\
sidak (cont) & 0.967 & \cellcolor{red!22}0.927 & 0.2281 & 0.0116 & \underline{0.2398} & 0.965 & \cellcolor{red!24}0.920 & 0.2283 & 0.0132 & \underline{0.2415} & cont & \cellcolor{blue!28}0.978 & 0.968 & 0.964 & 0.964 & 0.958 & \cellcolor{blue!23}0.975 \\
boot (cont) & 0.955 & \cellcolor{red!27}0.913 & 0.2180 & 0.0169 & \textbf{0.2349} & 0.953 & \cellcolor{red!27}0.913 & 0.2181 & 0.0189 & \textbf{0.2370} & cont & 0.963 & 0.954 & 0.952 & 0.953 & 0.951 & 0.958 \\
\midrule
bonferroni (lik) & 0.964 & \cellcolor{red!23}0.923 & 0.9469 & 0.0604 & 1.0073 & 0.963 & \cellcolor{red!28}0.910 & 0.9468 & 0.0630 & 1.0098 & lik & \cellcolor{blue!16}0.971 & 0.964 & 0.963 & 0.961 & 0.958 & 0.969 \\
max\_t (lik) & 0.958 & \cellcolor{red!24}0.920 & 0.9709 & 0.0623 & 1.0332 & 0.957 & \cellcolor{red!32}0.897 & 0.9748 & 0.0649 & 1.0397 & lik & \cellcolor{blue!29}0.979 & 0.963 & 0.955 & 0.950 & 0.954 & 0.962 \\
sidak (lik) & 0.958 & \cellcolor{red!27}0.913 & 0.8651 & 0.0741 & \underline{0.9391} & 0.957 & \cellcolor{red!32}0.897 & 0.8654 & 0.0772 & \underline{0.9426} & lik & 0.958 & 0.957 & 0.956 & 0.957 & 0.952 & 0.963 \\
boot (lik) & \cellcolor{red!17}0.943 & \cellcolor{red!37}0.883 & 0.8272 & 0.1096 & \textbf{0.9368} & \cellcolor{red!17}0.942 & \cellcolor{red!47}0.853 & 0.8274 & 0.1146 & \textbf{0.9420} & lik & \cellcolor{red!20}0.933 & \cellcolor{red!18}0.938 & \cellcolor{red!17}0.942 & \cellcolor{red!16}0.944 & \cellcolor{red!16}0.945 & \cellcolor{red!18}0.940 \\
\bottomrule
\end{tabular}
\end{fitwide}
\caption{Simultaneous CI methods on our synthetic suite (nominal 95\%, 300 MC reps per cell).}
\label{tab:fwer:simult:synth}
\vspace{1.2em}
{\scriptsize\setlength{\tabcolsep}{1.5pt}
\begin{fitwide}
\begin{tabular}{lrrrrrrrrrrrrrrrrrrrrr}
\toprule
CI method & Cov(null) & MinCov(null) & Width(null) & Pen(null) & Score(null) & Cov(alt) & MinCov(alt) & Width(alt) & Pen(alt) & Score(alt) & Type & n=10 & n=15 & n=20 & n=30 & n=50 & n=75 & n=100 & k=3 & k=5 \\
\midrule
bonferroni & \cellcolor{blue!25}0.976 & \cellcolor{red!28}0.910 & 0.5150 & 0.0277 & 0.5427 & 0.970 & \cellcolor{red!30}0.903 & 0.4870 & 0.0327 & 0.5197 & bin & \cellcolor{blue!45}0.988 & \cellcolor{blue!50}0.991 & \cellcolor{blue!32}0.980 & \cellcolor{blue!28}0.978 & 0.969 & 0.964 & 0.960 & \cellcolor{blue!18}0.972 & \cellcolor{blue!37}0.983 \\
max\_t & \cellcolor{red!17}0.942 & \cellcolor{red!65}0.490 & 0.4328 & 0.0515 & \underline{0.4843} & \cellcolor{red!58}0.821 & \cellcolor{red!65}0.037 & 0.4105 & 0.2911 & 0.7016 & bin & \cellcolor{red!45}0.860 & \cellcolor{red!22}0.928 & 0.957 & 0.968 & 0.962 & 0.962 & 0.958 & \cellcolor{red!22}0.929 & 0.963 \\
sidak & \cellcolor{blue!33}0.981 & \cellcolor{red!31}0.900 & 0.4866 & 0.0187 & 0.5053 & \cellcolor{blue!37}0.983 & \cellcolor{red!19}0.937 & 0.4681 & 0.0160 & 0.4841 & bin & \cellcolor{blue!58}0.996 & \cellcolor{blue!50}0.991 & \cellcolor{blue!50}0.991 & \cellcolor{blue!40}0.985 & \cellcolor{blue!20}0.973 & 0.968 & 0.967 & \cellcolor{blue!28}0.978 & \cellcolor{blue!43}0.987 \\
boot & 0.962 & \cellcolor{red!65}0.583 & 0.4556 & 0.0248 & \textbf{0.4804} & 0.962 & \cellcolor{red!65}0.617 & 0.4351 & 0.0239 & 0.4590 & bin & \cellcolor{red!20}0.934 & 0.964 & \cellcolor{blue!17}0.971 & \cellcolor{blue!22}0.974 & 0.970 & 0.964 & 0.960 & 0.954 & \cellcolor{blue!27}0.977 \\
\bottomrule
\end{tabular}
\end{fitwide}
}
\caption{Simultaneous CI methods on our real evals data (binary only, $k \in \{3, 5\}$; resampling used 5000 reps).}
\label{tab:fwer:simult:real}
\end{table*}

\textbf{For FWER correction of simultaneous CIs at small $N{<}100$, we recommend \v{S}id\'ak.} It never drops below 0.95 average coverage for any eval type and $n$, and its worst cell is 0.897. The joint bootstrap looks slightly narrower, but it is inconsistent, with lower worst-case coverage and collapse on sparse or lopsided binary data.

\paragraph{P-value correction.}

We compare eleven strategies: no correction, Holm~\cite{holm1979simple}, Bonferroni, Benjamini-Hochberg FDR control~\cite{benjamini1995controlling}, Hochberg~\cite{hochberg1988sharper}, Shaffer~\cite{shaffer1986modified}, Friedman-Nemenyi~\cite{demvsar2006statistical}, the max-T bootstrap~\cite{goeman2014multiple} and its step-down refinement Romano-Wolf~\cite{romano2005exact}, the permutation-based Westfall-Young~\cite{westfall1993resampling}, and a joint-bootstrap correction. The first six are agnostic to the underlying test, so we feed them \pkg{}' single-comparison defaults, the McNemar mid-$p$ for binary and the Wilcoxon signed-rank $p$-value for numeric data (\S\ref{app:pvalue:pairwise}). max-T, Romano-Wolf, and Westfall-Young instead build their own pairwise p-values. Friedman-Nemenyi is likewise self-contained.

\begin{figure}[tbp]
\centering
\centering
\includegraphics[width=\linewidth]{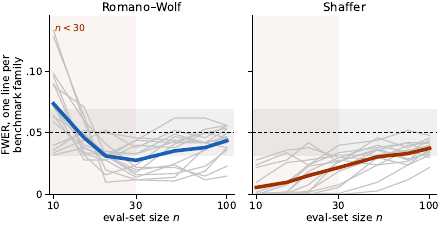}
\caption{Family-wise error rate on real benchmark data, one gray line per set. Shading is the Monte-Carlo interval for a perfectly
calibrated method at the reps behind this plot. Romano--Wolf is unstable below $n{\approx}20$; Shaffer is
conservative throughout. 500 replicates per cell.}
\label{fig:rw-stability}
\end{figure}

Tables~\ref{tab:fwer:pvalue:synth} and~\ref{tab:fwer:pvalue:real} report the synthetic and real cases. Romano-Wolf is the best choice at $n\ge30$ (Figure~\ref{fig:rw-stability}), with well-calibrated Type~I control at slightly greater power than max-T (holding this performance to $n=1000$ on synthetic data). However, Romano-Wolf's failure mode, shared with the other resampling methods, is extremely lopsided binary data at low $n$ and $k$. Westfall-Young sits on nominal at larger $n$ but exceeds it in places (0.069 at $n=50$ on binary data) and runs much slower, so we do not recommend it. Among methods that correct rather than replace the pairwise $p$-value, Shaffer's procedure is the safer choice. \label{app:rw-stability} 
We recommend:

\begin{table*}[t]
\centering
\vspace{1.2em}
\scriptsize
\setlength{\tabcolsep}{2pt}
\begin{tabular}{lrrrrrrrrrrrrrrr}
\toprule
Correction & FWER & Best-arm power & Time (ms) & Type & n=15 & n=30 & n=50 & n=100 & n=200 & n=500 & n=1000 & k=3 & k=5 & k=10 & k=20 \\
\midrule
holm (bin) & \cellcolor{blue!27}0.023 & 0.219 & 0.0 & bin & \cellcolor{blue!62}0.002 & \cellcolor{blue!48}0.010 & \cellcolor{blue!35}0.018 & \cellcolor{blue!22}0.026 & 0.031 & 0.036 & 0.038 & \cellcolor{blue!17}0.029 & \cellcolor{blue!27}0.023 & \cellcolor{blue!28}0.022 & \cellcolor{blue!35}0.018 \\
bonferroni (bin) & \cellcolor{blue!27}0.023 & 0.210 & 0.0 & bin & \cellcolor{blue!62}0.002 & \cellcolor{blue!48}0.010 & \cellcolor{blue!35}0.018 & \cellcolor{blue!22}0.026 & 0.031 & 0.036 & 0.038 & \cellcolor{blue!17}0.029 & \cellcolor{blue!27}0.023 & \cellcolor{blue!28}0.022 & \cellcolor{blue!35}0.018 \\
fdr\_bh (bin) & \cellcolor{blue!23}0.025 & \textbf{0.260} & 0.0 & bin & \cellcolor{blue!60}0.003 & \cellcolor{blue!47}0.011 & \cellcolor{blue!32}0.020 & \cellcolor{blue!18}0.028 & 0.033 & 0.040 & 0.041 & 0.031 & \cellcolor{blue!22}0.026 & \cellcolor{blue!25}0.024 & \cellcolor{blue!30}0.021 \\
hochberg (bin) & \cellcolor{blue!27}0.023 & 0.220 & 0.0 & bin & \cellcolor{blue!62}0.002 & \cellcolor{blue!48}0.010 & \cellcolor{blue!35}0.018 & \cellcolor{blue!22}0.026 & 0.031 & 0.037 & 0.038 & \cellcolor{blue!17}0.029 & \cellcolor{blue!25}0.024 & \cellcolor{blue!28}0.022 & \cellcolor{blue!35}0.018 \\
shaffer (bin) & \cellcolor{blue!27}0.023 & 0.228 & 0.1 & bin & \cellcolor{blue!62}0.002 & \cellcolor{blue!48}0.010 & \cellcolor{blue!35}0.018 & \cellcolor{blue!22}0.026 & 0.031 & 0.036 & 0.038 & \cellcolor{blue!17}0.029 & \cellcolor{blue!27}0.023 & \cellcolor{blue!28}0.022 & \cellcolor{blue!35}0.018 \\
friedman\_nemenyi (bin) & \cellcolor{blue!65}0.000 & 0.078 & 12.7 & bin & \cellcolor{blue!65}0.000 & \cellcolor{blue!65}0.000 & \cellcolor{blue!65}0.000 & \cellcolor{blue!65}0.000 & \cellcolor{blue!65}0.000 & \cellcolor{blue!65}0.000 & \cellcolor{blue!65}0.000 & \cellcolor{blue!65}0.000 & \cellcolor{blue!65}0.000 & \cellcolor{blue!65}0.000 & \cellcolor{blue!65}0.000 \\
max\_t (bin) & 0.035 & 0.219 & 0.0 & bin & 0.030 & \cellcolor{blue!30}0.021 & 0.030 & 0.032 & 0.038 & 0.045 & 0.049 & 0.047 & 0.037 & \cellcolor{blue!17}0.029 & \cellcolor{blue!20}0.027 \\
romano\_wolf (bin) & 0.035 & 0.228 & 1.2 & bin & 0.030 & \cellcolor{blue!30}0.021 & 0.030 & 0.032 & 0.038 & 0.045 & 0.049 & 0.047 & 0.037 & \cellcolor{blue!17}0.029 & \cellcolor{blue!20}0.027 \\
westfall\_young (bin) & 0.046 & \underline{0.229} & 44.3 & bin & \cellcolor{blue!35}0.018 & 0.039 & \cellcolor{red!21}0.069 & 0.049 & 0.046 & 0.051 & 0.049 & 0.046 & 0.046 & 0.046 & 0.046 \\
boot (bin) & \cellcolor{blue!20}0.027 & 0.217 & 0.1 & bin & \cellcolor{blue!48}0.010 & \cellcolor{blue!50}0.009 & \cellcolor{blue!40}0.015 & \cellcolor{blue!25}0.024 & 0.034 & 0.045 & 0.051 & 0.036 & \cellcolor{blue!20}0.027 & \cellcolor{blue!27}0.023 & \cellcolor{blue!28}0.022 \\
\midrule
holm (cont) & 0.031 & 0.405 & 0.0 & cont & \cellcolor{blue!42}0.014 & \cellcolor{blue!27}0.023 & 0.032 & 0.035 & 0.038 & 0.039 & 0.039 & 0.042 & 0.034 & \cellcolor{blue!20}0.027 & \cellcolor{blue!27}0.023 \\
bonferroni (cont) & 0.031 & 0.393 & 0.0 & cont & \cellcolor{blue!42}0.014 & \cellcolor{blue!27}0.023 & 0.032 & 0.035 & 0.038 & 0.039 & 0.039 & 0.042 & 0.034 & \cellcolor{blue!20}0.027 & \cellcolor{blue!27}0.023 \\
fdr\_bh (cont) & 0.035 & \textbf{0.455} & 0.0 & cont & \cellcolor{blue!40}0.015 & \cellcolor{blue!22}0.026 & 0.035 & 0.039 & 0.043 & 0.043 & 0.043 & 0.044 & 0.038 & 0.031 & \cellcolor{blue!22}0.026 \\
hochberg (cont) & 0.032 & 0.406 & 0.0 & cont & \cellcolor{blue!42}0.014 & \cellcolor{blue!27}0.023 & 0.032 & 0.035 & 0.039 & 0.039 & 0.039 & 0.042 & 0.034 & \cellcolor{blue!20}0.027 & \cellcolor{blue!27}0.023 \\
shaffer (cont) & 0.031 & \underline{0.416} & 0.1 & cont & \cellcolor{blue!42}0.014 & \cellcolor{blue!27}0.023 & 0.032 & 0.035 & 0.038 & 0.039 & 0.039 & 0.042 & 0.034 & \cellcolor{blue!20}0.027 & \cellcolor{blue!27}0.023 \\
friedman\_nemenyi (cont) & \cellcolor{blue!33}0.019 & 0.364 & 15.6 & cont & \cellcolor{blue!37}0.017 & \cellcolor{blue!33}0.019 & \cellcolor{blue!32}0.020 & \cellcolor{blue!33}0.019 & \cellcolor{blue!32}0.020 & \cellcolor{blue!32}0.020 & \cellcolor{blue!32}0.020 & \cellcolor{blue!27}0.023 & \cellcolor{blue!30}0.021 & \cellcolor{blue!35}0.018 & \cellcolor{blue!38}0.016 \\
max\_t (cont) & 0.040 & 0.404 & 0.3 & cont & \cellcolor{blue!45}0.012 & \cellcolor{blue!17}0.029 & 0.041 & 0.050 & 0.050 & 0.051 & 0.049 & 0.046 & 0.043 & 0.039 & 0.034 \\
romano\_wolf (cont) & 0.040 & 0.415 & 13.6 & cont & \cellcolor{blue!45}0.012 & \cellcolor{blue!17}0.029 & 0.041 & 0.050 & 0.050 & 0.051 & 0.049 & 0.046 & 0.043 & 0.039 & 0.034 \\
westfall\_young (cont) & 0.049 & 0.416 & 528.0 & cont & 0.049 & 0.047 & 0.049 & 0.051 & 0.050 & 0.051 & 0.049 & 0.050 & 0.050 & 0.049 & 0.048 \\
boot (cont) & 0.030 & 0.394 & 0.3 & cont & \cellcolor{blue!63}0.001 & \cellcolor{blue!50}0.009 & \cellcolor{blue!27}0.023 & 0.036 & 0.045 & 0.048 & 0.049 & 0.037 & 0.033 & \cellcolor{blue!18}0.028 & \cellcolor{blue!27}0.023 \\
\midrule
holm (lik) & \cellcolor{blue!17}0.029 & 0.394 & 0.0 & lik & \cellcolor{blue!50}0.009 & \cellcolor{blue!33}0.019 & \cellcolor{blue!17}0.029 & 0.033 & 0.036 & 0.041 & 0.036 & 0.039 & 0.032 & \cellcolor{blue!23}0.025 & \cellcolor{blue!32}0.020 \\
bonferroni (lik) & \cellcolor{blue!17}0.029 & 0.382 & 0.0 & lik & \cellcolor{blue!50}0.009 & \cellcolor{blue!33}0.019 & \cellcolor{blue!17}0.029 & 0.033 & 0.036 & 0.041 & 0.036 & 0.039 & 0.032 & \cellcolor{blue!23}0.025 & \cellcolor{blue!32}0.020 \\
fdr\_bh (lik) & 0.032 & \textbf{0.444} & 0.0 & lik & \cellcolor{blue!48}0.010 & \cellcolor{blue!28}0.022 & 0.032 & 0.036 & 0.040 & 0.045 & 0.041 & 0.041 & 0.036 & \cellcolor{blue!18}0.028 & \cellcolor{blue!27}0.023 \\
hochberg (lik) & \cellcolor{blue!17}0.029 & 0.395 & 0.0 & lik & \cellcolor{blue!50}0.009 & \cellcolor{blue!33}0.019 & \cellcolor{blue!17}0.029 & 0.033 & 0.036 & 0.041 & 0.037 & 0.039 & 0.032 & \cellcolor{blue!23}0.025 & \cellcolor{blue!32}0.020 \\
shaffer (lik) & \cellcolor{blue!17}0.029 & 0.405 & 0.1 & lik & \cellcolor{blue!50}0.009 & \cellcolor{blue!33}0.019 & \cellcolor{blue!17}0.029 & 0.033 & 0.036 & 0.041 & 0.036 & 0.039 & 0.032 & \cellcolor{blue!23}0.025 & \cellcolor{blue!32}0.020 \\
friedman\_nemenyi (lik) & \cellcolor{blue!55}0.006 & 0.319 & 14.0 & lik & \cellcolor{blue!58}0.004 & \cellcolor{blue!57}0.005 & \cellcolor{blue!55}0.006 & \cellcolor{blue!53}0.007 & \cellcolor{blue!57}0.005 & \cellcolor{blue!55}0.006 & \cellcolor{blue!55}0.006 & \cellcolor{blue!55}0.006 & \cellcolor{blue!55}0.006 & \cellcolor{blue!55}0.006 & \cellcolor{blue!57}0.005 \\
max\_t (lik) & 0.043 & 0.395 & 0.2 & lik & \cellcolor{blue!35}0.018 & 0.036 & 0.047 & 0.048 & 0.049 & \cellcolor{red!16}0.053 & 0.048 & 0.046 & 0.045 & 0.041 & 0.038 \\
romano\_wolf (lik) & 0.043 & 0.405 & 12.2 & lik & \cellcolor{blue!35}0.018 & 0.036 & 0.047 & 0.048 & 0.049 & \cellcolor{red!16}0.053 & 0.048 & 0.046 & 0.045 & 0.041 & 0.038 \\
westfall\_young (lik) & 0.048 & \underline{0.406} & 474.0 & lik & 0.041 & 0.048 & 0.051 & 0.049 & 0.049 & \cellcolor{red!16}0.053 & 0.048 & 0.048 & 0.049 & 0.048 & 0.048 \\
boot (lik) & \cellcolor{blue!17}0.029 & 0.384 & 0.3 & lik & \cellcolor{blue!63}0.001 & \cellcolor{blue!52}0.008 & \cellcolor{blue!30}0.021 & 0.032 & 0.041 & 0.050 & 0.046 & 0.036 & 0.032 & \cellcolor{blue!22}0.026 & \cellcolor{blue!30}0.021 \\
\bottomrule
\end{tabular}
\caption{p-value FWER correction methods' Type I error rate and true effect discovery power across our synthetic data suite (500 MC reps per cell). Corrections are applied to the base $p$-value \pkg{} reports for the data type---mid-$p$ on binary, Wilcoxon otherwise---except max-T, Romano--Wolf and Westfall--Young, which build their own joint statistic.}
\label{tab:fwer:pvalue:synth}
\vspace{1.2em}
\begin{tabular}{lrrrrrrrrrrrrr}
\toprule
Correction & FWER & Best-arm power & Time (ms) & Type & n=10 & n=15 & n=20 & n=30 & n=50 & n=75 & n=100 & k=3 & k=5 \\
\midrule
holm & \cellcolor{blue!32}0.020 & 0.134 & 0.0 & bin & \cellcolor{blue!58}0.004 & \cellcolor{blue!53}0.007 & \cellcolor{blue!43}0.013 & \cellcolor{blue!33}0.019 & \cellcolor{blue!17}0.029 & 0.032 & 0.037 & \cellcolor{blue!27}0.023 & \cellcolor{blue!40}0.015 \\
bonferroni & \cellcolor{blue!32}0.020 & 0.120 & 0.0 & bin & \cellcolor{blue!58}0.004 & \cellcolor{blue!53}0.007 & \cellcolor{blue!43}0.013 & \cellcolor{blue!33}0.019 & \cellcolor{blue!17}0.029 & 0.032 & 0.037 & \cellcolor{blue!27}0.023 & \cellcolor{blue!40}0.015 \\
fdr\_bh & \cellcolor{blue!30}0.021 & \textbf{0.153} & 0.0 & bin & \cellcolor{blue!58}0.004 & \cellcolor{blue!52}0.008 & \cellcolor{blue!43}0.013 & \cellcolor{blue!30}0.021 & 0.031 & 0.034 & 0.039 & \cellcolor{blue!25}0.024 & \cellcolor{blue!38}0.016 \\
hochberg & \cellcolor{blue!32}0.020 & 0.135 & 0.0 & bin & \cellcolor{blue!58}0.004 & \cellcolor{blue!53}0.007 & \cellcolor{blue!43}0.013 & \cellcolor{blue!32}0.020 & \cellcolor{blue!17}0.029 & 0.032 & 0.037 & \cellcolor{blue!25}0.024 & \cellcolor{blue!40}0.015 \\
shaffer & \cellcolor{blue!32}0.020 & \underline{0.151} & 0.0 & bin & \cellcolor{blue!58}0.004 & \cellcolor{blue!53}0.007 & \cellcolor{blue!43}0.013 & \cellcolor{blue!33}0.019 & \cellcolor{blue!17}0.029 & 0.032 & 0.037 & \cellcolor{blue!27}0.023 & \cellcolor{blue!40}0.015 \\
friedman\_nemenyi & \cellcolor{blue!65}0.000 & 0.060 & 2.3 & bin & \cellcolor{blue!63}0.001 & \cellcolor{blue!63}0.001 & \cellcolor{blue!63}0.001 & \cellcolor{blue!65}0.000 & \cellcolor{blue!63}0.001 & \cellcolor{blue!65}0.000 & \cellcolor{blue!65}0.000 & \cellcolor{blue!63}0.001 & \cellcolor{blue!65}0.000 \\
max\_t & 0.042 & 0.127 & 0.0 & bin & \cellcolor{red!24}0.078 & 0.046 & 0.031 & \cellcolor{blue!23}0.025 & 0.034 & 0.036 & 0.043 & 0.047 & 0.033 \\
romano\_wolf & 0.042 & 0.143 & 1.7 & bin & \cellcolor{red!24}0.078 & 0.046 & 0.031 & \cellcolor{blue!23}0.025 & 0.034 & 0.036 & 0.043 & 0.047 & 0.033 \\
westfall\_young & 0.038 & 0.144 & 20.9 & bin & \cellcolor{blue!20}0.027 & \cellcolor{blue!33}0.019 & \cellcolor{blue!17}0.029 & 0.039 & \cellcolor{red!15}0.052 & 0.044 & \cellcolor{red!15}0.052 & 0.037 & 0.039 \\
boot & \cellcolor{blue!33}0.019 & 0.110 & 0.1 & bin & \cellcolor{blue!65}0.000 & \cellcolor{blue!37}0.017 & \cellcolor{blue!40}0.015 & \cellcolor{blue!45}0.012 & \cellcolor{blue!30}0.021 & \cellcolor{blue!17}0.029 & 0.036 & \cellcolor{blue!27}0.023 & \cellcolor{blue!45}0.012 \\
\bottomrule
\end{tabular}
\caption{p-value FWER correction methods' Type I error rate and true effect discovery power on real evals data (binary only). Per-$n$ and per-$k$ FWER columns are collapsed across the other dimension. (95\% MC band omitted for space; mean half-width $\pm$0.001.)}
\label{tab:fwer:pvalue:real}
\end{table*}

\begin{itemize}
    \item \textbf{All eval types, $n\ge30$}: Romano-Wolf~\cite{romano2005exact}, noting it \emph{replaces} rather than corrects $p$-values of pairwise tests.
    \item \textbf{All eval types, $n<30$, or when pairwise test p-values are required}: Shaffer's procedure, which corrects the Wilcoxon or McNemar $p$-value directly. (Benjamini-Hochberg has the highest power but controls FDR, not FWER, and is included for reference only.)
    \item \textbf{Exception to Romano-Wolf for extremely lopsided binary data}: if \textit{any} compared group has fewer than 5 observations of its rarer outcome, prefer Shaffer's procedure for \textit{all} comparisons; Romano-Wolf can exhibit inflated worst-case FWER here.
\end{itemize}

\newpage {~}
\newpage
\newpage {~}
\newpage
\newpage {~}
\newpage
\section{Validation of PPI-corrected tests} \label{app:ppi}

\subsection{Bootstrap-adaptive power tuning} \label{app:ppi:adaptive}

PPI++~\cite{angelopoulos2023ppiplusplus} refines PPI with a weight $\lambda$ on the corrected estimate's judge terms, chosen so that the estimate's asymptotic variance is minimized: for a mean, $\lambda^\star = \operatorname{Cov}(Y, f)/\big((1 + N_\mathrm{lab}/N_\mathrm{unlab})\operatorname{Var}(f)\big)$. In practice, PPI++ estimates $\lambda^\star$ with a plug-in $\hat\lambda$ computed on the labeled items, optionally clipped to $[0,1]$, and forms intervals as if $\hat\lambda$ were known. Its guarantees, including never losing power to the human-only estimate, are asymptotic. With 15-50 labels, $\hat\lambda$ is noisy, and PPI++ can underperform the human-only estimate~\cite{mani2026nofreelunch}---meaning the very regime that most practitioners want to use PPI for, i.e. few human-labeled samples, is the regime it fails at.

To remedy this, one approach is to stabilize $\hat\lambda$ by shrinking it toward a fixed value---in practice $0$ or $1$---but both came with trade-offs when tested in our simulations. Toward $1$ (regular PPI) lost the benefits of PPI++, causing poor judges to falter below running statistics over the human labels alone. Toward $0$ discarded power from a strong judge (well-aligned to human raters) at small $N_{\mathrm{lab}}$, and also broke down in MNAR regimes due to the over-reliance on the (now-biased) labeled sample. In short, PPI++'s weight is fine, but its certainty about that weight is not; shrinkage can fix it but needs a target, and a fixed target fails one way or the other. 

\pkg{} therefore makes two changes: we let the labeled data choose the target, by resampling it and asking how often it says the judge is informative, and we add $\hat\lambda$'s own sampling variance to the interval. This kept Type I error near nominal at small $N_\mathrm{lab}$ without giving up power for a strong judge.

\paragraph{Shrinkage toward an estimated target.} We clip $\hat\lambda$ to $[0,1]$ and recompute it on $B{=}800$ resamples of the labeled pairs; the unlabeled sample is large enough that its variance enters as a constant, so only the labeled pairs are resampled. We then set $\tau = \tfrac{1}{B}\sum_{b=1}^{B}\mathbf{1}[\hat\lambda_b \ge 0.5]$, the share of resamples in which the judge earns at least half the weight. We use that probability directly as the shrinkage target. The weight used is:
\[
\lambda = w\hat\lambda + (1-w)\tau, \qquad w = \frac{N_\mathrm{lab}}{N_\mathrm{lab} + 20}.
\]
With few labels, $\lambda$ moves toward $1$ when resamples consistently find an informative judge and toward $0$ when they do not; as $N_\mathrm{lab}$ grows, $w \to 1$ and $\lambda$ approaches the PPI++ estimate. The pseudo-count $20$ puts $w{=}\tfrac{1}{2}$ at $N_\mathrm{lab}{=}20$, so the data-driven estimate dominates once labeling reaches the $30$ we recommend (\S\ref{sec:guidelines}). 

For per-condition CIs on numeric data, a second blend guards against non-random labeling (MNAR). Let $r$ be the difference between the judge's labeled and unlabeled mean scores, let $z = r/\mathrm{SE}(r)$, let $e = \max(0, z^2 - 1)$, and let $g = e/(e+3)$. The weight then used is: $\lambda_{\mathrm{guard}} = (1-g)\,\lambda + g$. Under the sampling noise of truly random labeling, $z$ is approximately standard normal and $z^2$ averages $1$. An inflated $e$ is therefore evidence of a shift beyond that chance level, and $g$ reaches $\tfrac12$ at $|z|{=}2$. A detected shift pushes the weight back toward $1$, recovering much of the coverage otherwise lost under non-random labeling.

\paragraph{Variance of the estimated weight.} Because $\lambda$ is estimated from the same labels it weights, we add its sampling variance to that of the corrected estimate. For an estimate $\hat\theta^{H}_{\mathrm{lab}} + \lambda r$, where $\hat\theta^{H}_{\mathrm{lab}}$ is the human-label estimate on the labeled items, this is the delta-method term $r^2\,\widehat{\operatorname{Var}}(\hat\lambda)$, with the variance taken over the bootstrap replicates. The standard error is taken after this term is added, so the interval and the $p$-value both widen. For bootstrap-based intervals, the same variance is added as noise to the replicates. Appendix~\ref{sec:appendix-rank-based} gives the form for rank tests.

\paragraph{Related approaches.} \citet{mani2026nofreelunch} address the same plug-in $\hat\lambda$ problem with cross-fitting, which splits an already small labeled set. In our ablations, their method was slightly closer to nominal Type I error in MCAR, but at wider intervals; was much more fragile in MNAR regimes, causing Type I error spikes; and held no power advantage at matched Type I error. \citet{li2025adaptiveshrinkage} shrink power-tuning parameters across many estimation tasks by empirical Bayes, and \citet{shoham2026prediction} tune $\lambda$ online while training a model; neither targets a single test with few labels. Despite the name, bootstrap-adaptive power tuning differs from PPBoot~\cite{zrnic2024note}, which resamples labeled and unlabeled data to form the interval itself and plugs in a single bootstrap estimate of $\lambda$. Here, resampling instead measures how reliable $\hat\lambda$ is, which sets its shrinkage target and adds its variance; our tests whose $\lambda$ comes from PPBoot replicates apply the same shrinkage to it (e.g., Mann-Whitney U). 

\subsection{LLM judge-bias simulation data} \label{appendix:ppi}

\subsubsection{Synthetic judge-bias generation} \label{sec:datagen:judgesynth}

Our synthetic simulations for PPI-corrected methods sweep LLM judge behaviors alongside dataset size $N$ and human-labeled subset size $N_\mathrm{lab}$, on top of the synthetic data generators of \S\ref{sec:datagen:synth}. We use two sweep designs that complement each other: \textit{One-factor-at-a-time} (OFAT) varies a single judge-behavior factor from a fixed baseline; and \textit{factorial} crosses several factors at once for each data type (bias magnitude, sample size, label count, label mechanism, effect size, and bias direction, with judge noise varied in null-effect cells only) to catch calibration failures that emerge only in combination.

Each factor we sweep mimics a documented way an LLM judge misbehaves (Table~\ref{tab:judge-bias-sweep}): additive \emph{bias}~\cite{zheng2023judging}, separated into constant (the same shift across groups, which cancels in a comparison) and differential (group-specific); \emph{confounding covariates} such as response length; per-item \emph{noise}, either Gaussian or ``contaminated'' (mostly on-target but occasionally very wrong/right); \emph{scale miscalibration}, modeled as a slope around an anchor; and \emph{non-random (MNAR) labeling} at two strengths alongside the MCAR default, kept separate in plots since PPI is known to fail under MNAR.

In the factorial sweep, bias, noise, and effect-size magnitudes are fractions of each eval type's population SD (Table~\ref{tab:pop-sd}) rather than raw units, so ``moderate'' means the same severity across eval types; the OFAT sweep mostly uses native-scale values, and binary judges use flip probabilities. Crossing every factor with every truth shape in Table~\ref{tab:synthetic-shapes} is intractable, so truth is drawn from one representative shape per eval type, with a handful of dedicated scenarios testing pathological shapes (zero-inflated, bimodal, mixture). %

\begin{table}[t]
\centering
\small
\begin{tabular}{|l|r|}
\hline
\textbf{Eval type} & \textbf{Population SD} \\
\hline
Binary $\{0,1\}$ & 0.5000 \\
Continuous $[0,1]$ & 0.1206 \\
Likert (1--5) & 1.1447 \\
\hline
\end{tabular}
\caption{Population standard deviation of each eval type's representative truth distribution, used as the standardization unit for all judge-bias/noise/effect-size magnitudes in Table~\ref{tab:judge-bias-sweep} and \S\ref{sec:ppi-calibration}. This ensures that we can compare power across data types.}
\label{tab:pop-sd}
\end{table}

\begin{table*}[t]
\centering
\small
\renewcommand{\arraystretch}{1.15}
\begin{fitwide}
\begin{tabular}{|l|>{\raggedright\arraybackslash}p{5.6cm}|>{\raggedright\arraybackslash}p{6.7cm}|}
\hline
\textbf{Factor} & \textbf{Values swept} & \textbf{Notes} \\
\hline
Sample size $N$ & 60, 100, 200, 400 & \\
\hline
Group balance & 1:1, 2:1, 4:1 & Model comparisons are rarely balanced in practice (e.g.\ an established baseline with many logged runs vs.\ a new system with few) \\
\hline
Label fraction $N_\mathrm{lab}/N$ & 5\%, 10\%, 20\%, 40\% (one-factor-at-a-time); absolute $N_\mathrm{lab} \in \{15, 30, 80\}$ (factorial) & Spans a handful of labels to a moderate portion. The same grid is used for real data (\S\ref{sec:datagen:realjudge}). \\
\hline
Label mechanism & MCAR; MNAR (mild, strength 0.8); MNAR (strong, strength 1.6) & MNAR tests robustness to the realistic case where a human labels only suspicious/borderline items rather than a random sampling properly. \\
\hline
Judge noise SD (population-SD fraction) & For factorial sweep: 31-point geometric grid, 0.025--28.4 (null cells only); For OFAT sweep: $\{0, 0.10, 0.35, 0.70\}$ & Spans a near-perfect judge to one barely better than chance. \\
\hline
Noise family & Gaussian (uniformly uncertain); ``contaminated'' (90\% small-noise items, 10\% catastrophic-noise items) & ``Contaminated'' models a judge that is mostly right but occasionally very badly wrong, while Gaussian is uniformly noisy around a mean. \\
\hline
Bias type & None; constant (same shift every group); differential (group-specific shift) & Favoritism toward one group in a comparison over another (differential); and constant bias regardless of situation. \\
\hline
Bias magnitude (population-SD fraction) & Mild (0.03), moderate (0.07), severe (0.30) & The degree of systemic, additive bias of the judge. For instance, 0.30 SD means the judge rates items that much higher/lower than humans, on average (in one direction). \\
\hline
Scale (slope) miscalibration & Compress (0.80$\times$), expand (1.20$\times$), per-group differential & Models a judge reluctant to use scale extremes, or one that overuses them, independent of any additive bias. For instance, judges preferring scores closer to 3 on a 1-5 Likert scale, or preferring the opposite, piling up at extremes. \\
\hline
Shared error across conditions for the same item & 0.0, 0.3, 0.7 & Repeated-measures tests only. Higher means the judge's error attaches to the item and carries across conditions; lower means the error is redrawn each time. \\
\hline
Confound & Pure-nuisance and truth-correlated variants, several magnitudes & Judge score depends on a variable (e.g., response length, formatting) that is unrelated (nuisance) or correlated (quality-correlated) with true quality. \\
\hline
\hline
\multicolumn{3}{|l|}{\textbf{Factorial cross (combined-factor grid, used for alignment-bucketed and per-test Type-I views)}} \\
\hline
\multicolumn{2}{|l|}{$N \times N_\mathrm{lab} \times \text{label mechanism} \times \text{effect size} \times \text{bias magnitude} \times \text{bias direction} \times \text{judge noise}$} & Bias magnitude $\in\{$none, moderate, severe$\}$; effect size $\in\{$null, moderate ($d{=}0.5$), large ($d{=}0.8$)$\}$; bias direction $\in\{$opposing, reinforcing$\}$ the injected real effect --- tests whether a favorable judge bias can mask a real effect's absence, or an unfavorable one can obscure a real effect's presence \\
\hline
\end{tabular}
\end{fitwide}
\caption{Judge-bias-specific factors swept on top of our synthetic data suite (\S\ref{sec:datagen:synth}, Table~\ref{tab:synthetic-shapes}). Factors are chosen to model a concrete, realistic way an LLM judge can misbehave. Each factor is varied one at a time from a fixed baseline, except the factorial cross (bottom), which combines seven factors simultaneously at a coarser per-factor resolution, varying judge noise only in null-effect cells.}
\label{tab:judge-bias-sweep}
\end{table*}

\subsubsection{Real LLM judge data sourcing} \label{sec:datagen:realjudge}

We collected five real-world (human\_label, judge\_score) corpora, each pairing a human label with one or more independently-collected LLM judge scores on the same items, plus a within-item paired variant of one corpus (Table~\ref{tab:real-judge-data}). Unlike \S\ref{sec:datagen:real}'s binary and continuous real-data corpora (we struggled to find large benchmarks with Likert scores), these span binary,  continuous, and Likert-style data.

Four corpora (Arena, WMT DA, App Store, Privacy Judge) have several independent judges for the same items, so we test the most adversarial scenario, using as \textit{different} judge's scores per group under comparison for the same items. Two corpora are narrower and feed a single check each---ICLR Metareview has one judge, and WMT DA (paired) is our only corpus with genuine within-item paired structure (the same segment translated by ${\ge}2$ systems). Table~\ref{tab:real-judge-data} shows details per corpus. For every corpus we reveal $\mathrm{label\_frac} \in \{5\%,10\%,20\%,40\%\}$ of the human labels (at least 15) uniformly at random (MCAR), the same convention as our synthetic sweep (Table~\ref{tab:judge-bias-sweep}). We do not simulate MNAR labeling here too, as PPI is already expected to falter under MNAR~\cite{angelopoulos2023ppiplusplus}.

\subsection{Type I error rates of bias-corrected tests versus running statistics over uncorrected judge scores}
\label{sec:ppi-typeI}

\begin{figure*}[t]
  \centering
  \includegraphics[width=\linewidth]{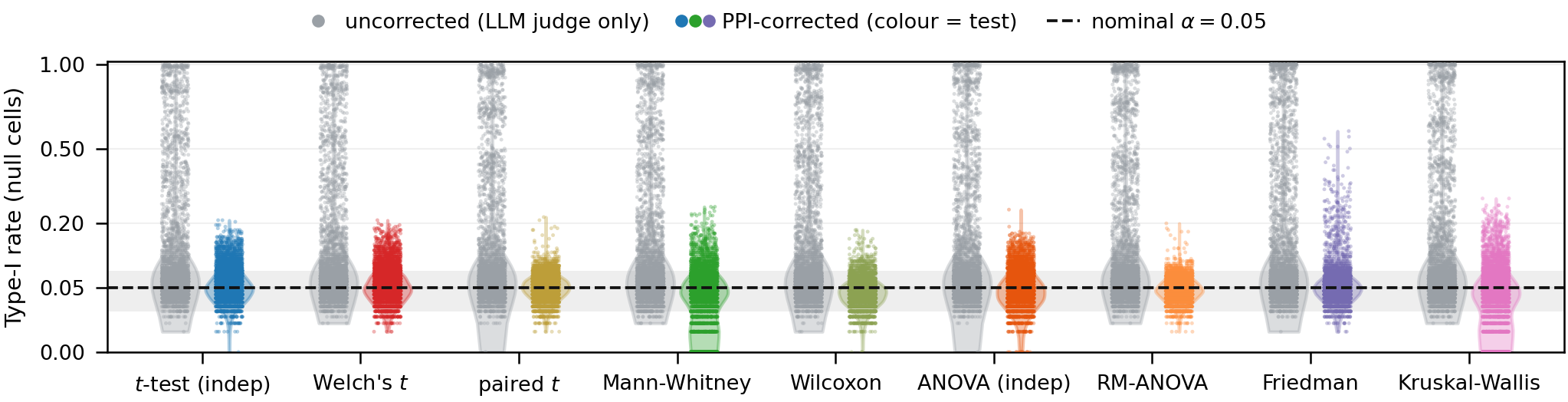}
  \caption{PPI-corrected Type-I error for hypothesis tests under missing-not-at-random (MNAR) labeling, where items for human labeling were not randomly selected, violating PPI's assumptions (e.g., labeling only the ``easy'' items). Computed from our aggressive factorial synthetic data sweep, which compounds multiple bias mechanisms per scenario. Spans 4{,}092 null scenarios (2{,}046 each at mild/strong MNAR strength) across all nine tests and 200 Monte Carlo repetitions each, for a total of $\approx$7.3M simulated trials. Rejection rates for paired t-test, Wilcoxon, and RM-ANOVA remain near nominal even as judge biases compound, while others exhibit instability. Friedman, MWU, and K-W rank-based tests appear to be the least reliable if MCAR labeling is violated.}
  \label{fig:typeI-factorial}
\end{figure*}

Figure~\ref{fig:ppi-type-i-error} shows results from our synthetic factorial sweep under MCAR labeling: uncorrected
statistics spike false positives under several bias mechanisms, while every PPI-corrected test holds near the nominal $\alpha=0.05$. Figure~\ref{fig:typeI-factorial} shows the same factorial design under MNAR labeling of which items receive human labels. Under MNAR, paired $t$, Wilcoxon, and RM-ANOVA stay near nominal on average, while Friedman and the independent-groups tests inflate, most under strong MNAR.  

\begin{figure*}[t]
  \centering
  \includegraphics[width=\linewidth]{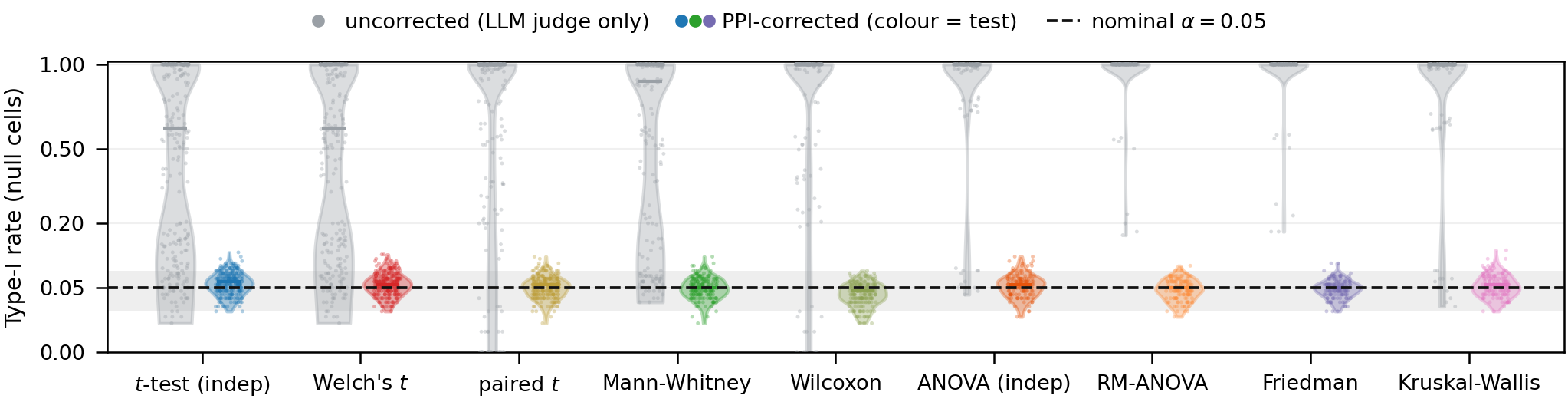}
  \caption{Type I error control of PPI-corrected tests on real LLM judge data (counterpart to Figure~\ref{fig:ppi-type-i-error}) under an MCAR labeling regime. Uncorrected rejection rates (gray) reach 1.0 for every test; PPI-corrected rates (colored) cluster tightly around nominal $\alpha=0.05$. 200 Monte Carlo replicates per test and corpus.}
  \label{fig:typeI-real-violin}
\end{figure*}

We repeated this same Type-I check on real judge data, specifically four of the five corpora listed in Table~\ref{tab:real-judge-data}: wmt\_da, privacy\_judge, arena, appstore (iclr\_metareview
is excluded here because it has only a single
judge model, and a Type-I null comparison needs at least two judges to compare). We pool across item count $N$ and judge pair/triple within each cell. Figure~\ref{fig:typeI-real-violin} shows the results, demonstrating that our PPI corrected tests' Type I error control remains calibrated in the face of real judge data, while without PPI, the false positive rate spikes (gray pillars). Table~\ref{tab:ppi:typeI:pertest} contains descriptive statistics for all three figures. 

\begin{table*}[t]
\centering
\scriptsize
\setlength{\tabcolsep}{2pt}
\begin{tabular}{lrrrrrrrrr!{\hspace{4pt}\vrule\hspace{4pt}}rr}
\toprule
 & \multicolumn{3}{c}{Synthetic, MCAR} & \multicolumn{3}{c}{Synthetic, MNAR} & \multicolumn{3}{c}{Real judge data} & \multicolumn{2}{c}{Power at $d{=}0.3$} \\
\cmidrule(lr){2-4}\cmidrule(lr){5-7}\cmidrule(lr){8-10}\cmidrule(lr){11-12}
Test & PPI & Unc. & Max & PPI & Unc. & Max & PPI & Unc. & Max & PPI & Subset \\
\midrule
$t$-test (indep) & .047 & .153 & .100 & \cellcolor{red!19}.063 & .153 & .210 & \cellcolor{red!18}.059 & .571 & .120 & .438 & .165 \\
Welch's $t$ & .048 & .153 & .110 & \cellcolor{red!19}.063 & .153 & .210 & \cellcolor{red!17}.058 & .571 & .115 & .363 & .148 \\
paired $t$ & .053 & .154 & .120 & \cellcolor{red!16}.054 & .154 & .220 & \cellcolor{red!16}.053 & .788 & .110 & .477 & .232 \\
Mann-Whitney $U$ & .049 & .154 & .120 & .051 & .154 & .255 & .052 & .637 & .110 & .383 & .165 \\
Wilcoxon & .044 & .156 & .125 & .047 & .157 & .180 & .043 & .855 & .080 & .328 & .145 \\
ANOVA (indep) & .053 & .154 & .110 & \cellcolor{red!17}.057 & .154 & .245 & \cellcolor{red!17}.056 & .896 & .110 & .927 & .355 \\
RM-ANOVA & .048 & .157 & .105 & .049 & .156 & .200 & .050 & .965 & .090 & .820 & .385 \\
Friedman & .051 & .172 & .120 & \cellcolor{red!19}.061 & .172 & .590 & .050 & .965 & .095 & .512 & .360 \\
Kruskal-Wallis & .052 & .156 & .120 & \cellcolor{red!16}.054 & .155 & .285 & \cellcolor{red!17}.056 & .887 & .125 & .890 & .418 \\
\bottomrule
\end{tabular}
\caption{Type I error of the nine PPI-corrected tests at nominal $\alpha{=}0.05$ (200 MC reps per cell), the
numbers behind Figures~\ref{fig:ppi-type-i-error} (MCAR) and~\ref{fig:typeI-factorial} (MNAR).
\textbf{PPI} and \textbf{Unc.} are the mean rejection rate across null cells with and without
correction. \textbf{Max} is the worst single null cell, the analogue of MinCov in the CI tables.
Synthetic columns pool 2{,}046 MCAR and 4{,}092 MNAR null cells per test over continuous and
Likert data; real columns pool 264 cells per test over the four multi-judge corpora of
Table~\ref{tab:real-judge-data}. Power is at Cohen's $d{=}0.3$ against a human-only subset of the same labeling budget. Per-factor breakdowns are in
Supplementary~Table~S1.}
\label{tab:ppi:typeI:pertest}
\end{table*}

\begin{table*}[t]
\centering
\scriptsize
\renewcommand{\arraystretch}{1.15}
\begin{fitwide}
\begin{tabular}{|l|p{2.4cm}|c|r|>{\raggedright\arraybackslash}p{4.3cm}|r|r|c|}
\hline
\textbf{Corpus} & \textbf{Eval type} & \textbf{Scale} & \textbf{$N$ items} & \textbf{Judge models used} & \textbf{Human label} & \textbf{Judge score} & \textbf{Pooled $r$} \\
& & & & & (mean $\pm$ SD) & (mean $\pm$ SD) & \\
\hline
Chatbot Arena & Binary & $\{0,1\}$ & 1{,}293 & Claude Haiku 4.5, Gemma-4 26B-A4B, GPT-OSS 20B, Inkling & $0.514 \pm 0.500$ & $0.521 \pm 0.500$ & 0.24 \\
WMT DA & Continuous & 0--100 & 1{,}000 & Claude Haiku 4.5, Gemma-4 26B-A4B, Inkling & $73.97 \pm 23.56$ & $72.48 \pm 24.99$ & 0.22 \\
App Store reviews & Likert & 1--5 & 300 & Claude Haiku 4.5, Gemma-4 26B-A4B, Inkling & $3.28 \pm 1.85$ & $2.84 \pm 1.79$ & 0.82 \\
Privacy Judge & Continuous (avg. Likert) & [1.0, 5.0] & 250 & Claude 3.5 Haiku, GPT-4o, Gemini 2.0 Flash, Gemma-3 4B, Llama-3.2 1B (5 of 13 collected) & $2.40 \pm 0.86$ & $2.57 \pm 1.24$ & 0.73 \\
ICLR Metareview & Binary & $\{0,1\}$ & 4{,}000 & ReviewerToo Metareviewer (single judge) & $0.283 \pm 0.451$ & $0.503 \pm 0.500$ & 0.11 \\
WMT DA (paired) & Continuous (paired) & 0--100 & 1{,}000 & Claude Haiku 4.5, Gemma-4 26B-A4B, Inkling & $66.87 \pm 33.40$ & $71.00 \pm 25.81$ & 0.25 \\
\hline
\end{tabular}
\end{fitwide}
\caption{The real LLM judge corpora (human\_label, judge\_score) used to validate PPI correction against
real judge bias. ``Pooled $r$'' is the Pearson correlation across every (item, judge) row,
pooled over judges. Two corpora differ a bit: Privacy Judge's human\_label is an aggregate of 50--68 survey participants per item; and ICLR Metareview's judge\_score comes from one automated  pipeline (55.5\% raw accuracy against real conference decisions over 18,322 evaluated papers, from which we drew a fixed-seed 4,000-paper
sample).}
\label{tab:real-judge-data}
\end{table*}

\subsection{Type I error by judge-human alignment, binary and continuous data}
\label{sec:alignment-binary-continuous}

We repeat the same alignment-bucketed Type-I check for binary (Cohen's
$\kappa$) and continuous (Pearson $r$) judge scores under MCAR labeling,
pairing each alignment bucket with a noise-only control so that an elevated
false-positive rate can be attributed to bias rather than to the bucketing
itself, and averaging over the five two-group tests (Student's, Welch's, and paired $t$,
Mann--Whitney $U$, Wilcoxon) for continuous and Likert data, and over Welch's and paired $t$ for binary data. Figure~\ref{fig:alignment-panels} in the main
text reports the result: PPI correction stays well-calibrated across the
alignment spectrum for all three data types, removing bias that common IRR
metrics do not spot. Note that IRR near $0$ means no discernible human--judge
correlation, so noise dominates any effect there is to find.

\begin{figure*}[t]
  \centering
  \includegraphics[width=\linewidth]{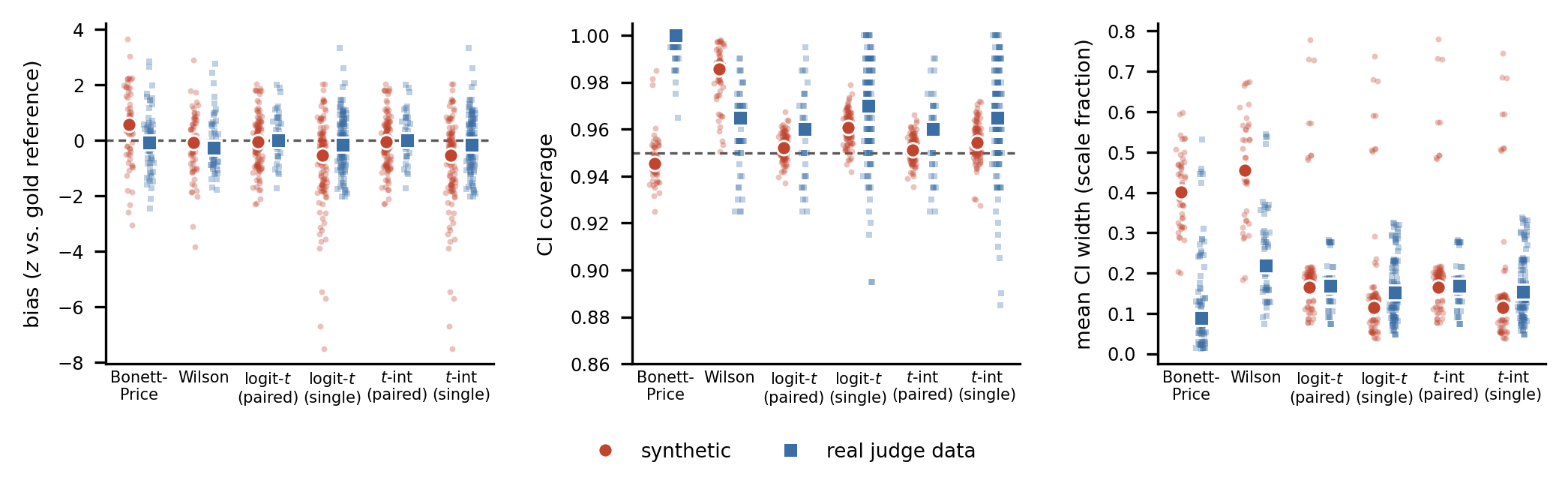}
  \caption{PPI correction of the four CI methods \pkg{} returns for
  \texttt{method="auto"}, on the synthetic suite (red circles) and real judge data (blue squares). Small marks are individual scenarios, large marks their median; the dashed lines mark zero bias and the nominal 95\% target.  Bonett-Price's real-data column runs conservative (99.7\%) because our real data simulation gives both arms the same human labels, leaving its rectifier
  no disagreements to learn from. 2{,}000 replicates per synthetic scenario, 200 per real-data scenario.} %
  \label{fig:ci-methods-comparison}
\end{figure*}

\subsection{PPI correction of main-path CI methods} \label{app:ppi:ci-methods}

\pkg{} also PPI-corrects marginal CIs and pairwise CIs that its \texttt{compare()} method returns by default, following our recommendations (Fig.~\ref{fig:ci-decision-tree}): Wilson and Bonett-Price and for binary marginal/pairwise CIs, a logit-$t$ interval for bounded $[0,1]$ data (we do not correct NIG; hence for the PPI path, we return logit-$t$, its runner-up), and (we include here) a $t$-interval in the case of unbounded numeric data. However, as these are more straightforward to apply PPI to, we did not focus on them in our main text. Figure~\ref{fig:ci-methods-comparison} shows these PPI corrections for CIs remain well-calibrated across synthetic and real judge data, with coverage near or above nominal.

\subsection{Do PPI omnibus tests preserve the ability to detect real differences?}
\label{sec:ppi-power}

Figure~\ref{fig:five-way-comparison-ppi-power} showed that our two-group PPI tests gain power over human-only subset under bias. Here, Figure~\ref{fig:five-way-comparison-omnibus} shows similar power gains hold for our four omnibus tests.

\section{Extending PPI correction to rank-based nonparametric tests}
\label{sec:appendix-rank-based}

\begin{figure*}[t]
  \centering
  \includegraphics[width=\linewidth]{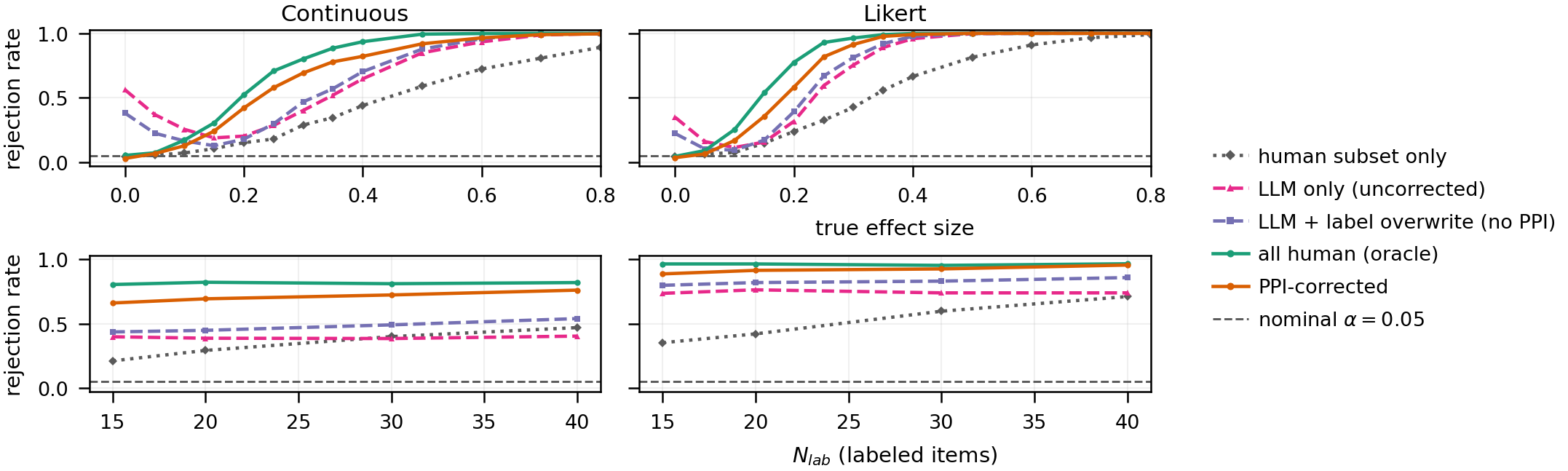}
  \caption{Omnibus-test (3+ condition) analogue of
  Figure~\ref{fig:five-way-comparison-ppi-power}, pooling
  ANOVA-independent, ANOVA-repeated, Friedman, and Kruskal-Wallis
  rejection rates across the same five-estimator comparison (all-human
  oracle, PPI-corrected, LLM-plus-label-overwrite,
  raw uncorrected LLM-only judgment, and a human-only subset matched in
  size to the labeled budget) for continuous and Likert data
  types. Top row: rejection rate as a function of true effect size at a fixed 20\% labeled-item budget. Bottom row: rejection rate as a function of labeled-item count $N_\mathrm{lab}$ at effect size Cohen's $d{=}0.3$. PPI-corrected power tracks close behind the oracle and clearly
  ahead of both the human-only subset and the uncorrected LLM baselines in every panel. 200 Monte Carlo replicates per condition.}
  \label{fig:five-way-comparison-omnibus}
\end{figure*}

Prediction-powered inference was formulated for estimators that are averages of per-item terms, where the correction decomposes linearly~\cite{angelopoulos2023ppiplusplus, angelopoulos2023prediction}. Rank statistics deviate, as an item's rank depends on every other item, so the item-wise decomposition cannot be applied to the literal statistic. However, technically PPI only needs an asymptotically linear estimator of a well-defined estimand. For rank-based tests we therefore identified a population estimand that is defined on any subsample and represents the same hypothesis, corrected that estimand, and referred the corrected statistic to a reference distribution whose calibration we verify empirically (Appendix~\ref{app:ppi}; Figs.~\ref{fig:typeI-factorial}, \ref{fig:typeI-real-violin} and~\ref{fig:five-way-comparison-omnibus}). The four tests are different variations on that template, with Wilcoxon using the H\'ajek projection of a one-sample $U$-statistic.\footnote{The projection of a $U$-statistic onto sums of per-item terms, which is linear up to $o_p(n^{-1/2})$~\cite[Ch.~12]{vandervaart1998asymptotic}.}

In the sections below, we describe each test. Items are assumed i.i.d., the labeled subset is a uniform random sample of the items (MCAR), and every variance is first-order. Mann-Whitney, Kruskal-Wallis, and Wilcoxon evaluate the variance of the corrected statistic under a sharper null than the weak null $\theta{=}0$ they test (exchangeability or sign symmetry, stated in each subsection), a score-type construction with the variance a function of the data under $H_0$ rather than of the estimate. This does not by itself guarantee validity under every distribution satisfying the weaker null; that finite-sample behavior is what the simulations of Appendix~\ref{app:ppi} measure. A separate check (equal medians, human-side spread ratios up to four, 1{,}000 replicates per cell; classical inflation to roughly twice nominal) found the PPI tests inheriting the Behrens--Fisher miscalibration of the classical Mann-Whitney and Kruskal-Wallis tests, as expected, without compounding it, with Wilcoxon and Friedman at nominal under a design that kept paired differences symmetric. In every test the power-tuning weight $\lambda$ is estimated from the data, clipped to $[0,1]$, and shrunk as in \S\ref{app:ppi:adaptive}; each subsection's variance formula shows whether $\lambda$'s own sampling variance enters. Throughout, a superscript $H$ or $J$ names the source (human labels or judge scores) and a subscript $\mathrm{lab}$ or $\mathrm{unlab}$ the sample; $\psi_i$ is item $i$'s first-order influence, and $N_\mathrm{lab}$ the total labeled count.

\subsection{PPI-corrected Friedman test}
\label{app:rank:friedman}

Friedman is the repeated-measures rank counterpart to one-way ANOVA: each subject's scores are ranked across the $k$ conditions. Because ranking uses only the subject's own row, the per-condition mean rank survives restriction to the labeled subset and no reformulation is needed. Writing $\bar r^{H}_{\mathrm{lab}}$, $\bar r^{J}_{\mathrm{lab}}$, $\bar r^{J}_{\mathrm{unlab}}$ for the $k$-vectors of per-condition mean ranks of the labeled human data and of the labeled and disjoint-unlabeled judge data, the corrected condition means are
\[
\hat m \;=\; \bar r^{H}_{\mathrm{lab}} + \lambda\,\big(\bar r^{J}_{\mathrm{unlab}} - \bar r^{J}_{\mathrm{lab}}\big),
\]
with one $\lambda$ shared across conditions, since the $k$ conditions share one judge. Let $P = I - \tfrac1k\mathbf{1}\mathbf{1}^{\top}$ project onto the contrast space, and let $V^{H}_{\mathrm{lab}}$, $V^{J}_{\mathrm{lab}}$, $V^{J}_{\mathrm{unlab}}$ and $C$ be the estimated covariance matrices of the three mean-rank vectors (each a row-sample covariance divided by its number of rows) and the cross-covariance of the two labeled ones. Then $\lambda = \operatorname{tr}(PCP)/\operatorname{tr}\big(P(V^{J}_{\mathrm{unlab}} + V^{J}_{\mathrm{lab}})P\big)$ minimizes the trace of the contrast-projected covariance of $\hat m$, the vector form of the PPI++ weight.

The corrected means are no longer literal sample mean ranks, so no exact finite-sample null distribution is available for them. The statistic keeps the Iman--Davenport~\cite{iman1980approximations} $F$-on-ranks shape but replaces its residual with a delta-method covariance of $\hat m$:
\[
F \;=\; \frac{\sum_j (\hat m_j - \bar{\hat m})^2}{\operatorname{tr}(P\hat V P)},
\]
\[
\begin{aligned}
\hat V \;=\;& V^{H}_{\mathrm{lab}} + \lambda^2\big(V^{J}_{\mathrm{unlab}} + V^{J}_{\mathrm{lab}}\big)\\
&- \lambda\,(C + C^{\top}) + r r^{\top}\,\widehat{\operatorname{Var}}(\lambda),
\end{aligned}
\]
with $r = \bar r^{J}_{\mathrm{unlab}} - \bar r^{J}_{\mathrm{lab}}$. We refer $F$ to $F\big(k-1,\,(N_\mathrm{lab}-1)(k-1)\big)$, the labeled count setting the denominator degrees of freedom because that is where $\hat V$'s noise lives, and rely on the simulations of Appendix~\ref{app:ppi} for its calibration.

\subsection{PPI-corrected Mann-Whitney $U$ test}
\label{app:rank:mwu}

For two groups $A$ and $B$ the estimand is $\theta = P(A{>}B) + \tfrac{1}{2}P(A{=}B) - \tfrac{1}{2}$, a two-sample $U$-statistic that equals zero whenever the groups share a distribution, ties included, and, up to the shift, is the classical $U/(n_A n_B)$. It is defined on any subsample, so the corrected estimator takes the standard power-tuned form
\[
\hat\theta \;=\; \hat\theta^{H}_{\mathrm{lab}} + \lambda\big(\hat\theta^{J}_{\mathrm{unlab}} - \hat\theta^{J}_{\mathrm{lab}}\big),
\]
with $\lambda = \widehat{\operatorname{Cov}}\big(\hat\theta^{H}_{\mathrm{lab}}, \hat\theta^{J}_{\mathrm{lab}}\big) / \widehat{\operatorname{Var}}\big(\hat\theta^{J}_{\mathrm{unlab}} - \hat\theta^{J}_{\mathrm{lab}}\big)$, the PPI++ weight with its covariance and variance taken over PPI bootstrap replicates~\cite{zrnic2024note}. Because $\lambda$ multiplies only the judge terms, the estimator is unbiased at any fixed $\lambda$.

The test uses a null-structured H\'ajek-projection covariance. Write $h_i$ and $j_i$ for item $i$'s human label and judge score, and $\Phi_H$, $\Phi_J$ for the empirical mid-CDFs (ties at half weight) of the \emph{other} group's human labels and judge scores. The first-order influence of an $A$ item on $\theta$ is $F_B$ at that item~\cite[Thm.~12.3]{vandervaart1998asymptotic}, which under $H_0$ the other group's mid-CDF estimates, so every item contributes one number,
\[
\psi_i \;=\; \Phi_H(h_i) - \lambda\,\Phi_J(j_i) \ \text{ if } i \text{ is labeled}, \qquad \psi_i \;=\; \lambda\,\Phi_J(j_i) \ \text{ if not},
\]
signed by group ($+$ for $A$, $-$ for $B$); neither the sign nor a centering constant enters a variance. Since the labeled and unlabeled items enter $\hat\theta$ through separate averages,
\[
\widehat{\operatorname{Var}}(\hat\theta) \;=\; \sum_{g \in \{A,B\}} \left[\frac{\operatorname{Var}_{\mathrm{lab},g}(\psi)}{n_{\mathrm{lab},g}} + \frac{\operatorname{Var}_{\mathrm{unlab},g}(\psi)}{n_{\mathrm{unlab},g}}\right],
\]
the sample variances taken within each group's labeled and unlabeled items. Differencing the human and judge influence per item, rather than assembling $V^{H}_{\mathrm{lab}} + \lambda^2(V^{J}_{\mathrm{unlab}} + V^{J}_{\mathrm{lab}}) - 2\lambda C$ from separately estimated variances and a cross-covariance, avoids a near-cancellation at $\lambda \approx 1$. The statistic $\hat\theta/\widehat{\mathrm{SE}}$ is referred to a $t$ distribution on $N_\mathrm{lab}-1$ degrees of freedom, with $N_\mathrm{lab}$ the total labeled count over both groups. 

The reported effect size is the rank-biserial correlation $2\hat\theta$, corrected by the same estimator. Unlike the classical statistic, a normalized rank count that is bounded by construction, the PPI-corrected $\hat\theta$ is a sum and is not confined to $[-\tfrac12,\tfrac12]$; therefore, upon display to users in \pkg{}, we clip the reported correlation to $[-1,1]$.

\subsection{PPI-corrected Kruskal-Wallis test}
\label{app:rank:kw}

Kruskal-Wallis compares the $k$ group mean ranks of a pooled ranking, and pooled ranks do not survive subsampling. Each group's mean rank is, however, an exact affine function of its pairwise dominance probabilities against the other groups. With $\theta_{ab} = P(\text{group } a > \text{group } b) + \tfrac12 P(\text{group } a = \text{group } b)$ and $\delta_{ab} = \theta_{ab} - \tfrac12$,
\[
\begin{gathered}
\bar R_j \;=\; \frac{N+1}{2} + s_j, \qquad s_j = \sum_{l \neq j} n_l\,\delta_{jl},\\
H \;=\; \frac{12}{N(N+1)}\sum_j n_j\, s_j^2
\end{gathered}
\]
up to the tie correction, since within-group midranks average to $(n_j+1)/2$ and each other group $l$ contributes $n_l\theta_{jl}$. So $H$ is a fixed quadratic form in the vector $\delta$ of all $\binom{k}{2}$ pairwise Mann-Whitney estimands, and we correct that vector. Each $\hat\delta_{ab}$ takes the power-tuned form of \S\ref{app:rank:mwu}, with one $\lambda$ shared across pairs, since the pairs share one judge: $\lambda = \operatorname{tr}(C)/\operatorname{tr}\big(V^{J}_{\mathrm{unlab}} + V^{J}_{\mathrm{lab}}\big)$, where $C$ is the cross-covariance of the labeled human and judge pair vectors and $V^{J}$ the covariances of the judge pair vectors, all taken over a joint bootstrap that resamples every group once per replicate (pairs sharing a group are correlated). The covariance used below has rank $k{-}1$, so the test covers only the part of $\delta$ that determines the group rank means; like classical Kruskal-Wallis, it has no power against cyclic dominance. At $k=2$ the estimand reduces to Mann-Whitney's.

Under exchangeability every group's items share one human and one judge distribution, so the reference mid-CDFs $\Phi_H$, $\Phi_J$ are now pooled across all groups. (Each group is only $1/k$ of the pool. At $k=2$ a pooled reference would be half fitted to the group it scores, which is why \S\ref{app:rank:mwu} uses the other group instead.) With that substitution, an item's first-order influence on every pair containing its group is the same scalar $\psi_i$ of \S\ref{app:rank:mwu}, signed by the group's position in the pair. The H\'ajek projection of $\hat\delta$ is therefore $M\bar\psi$, with $\bar\psi_g$ group $g$'s mean of $\psi$ and $M$ the oriented incidence matrix of the complete graph $K_k$. That matrix has rank $k-1$, so the first-order covariance of $\hat\delta$ has rank exactly $k-1$ whenever every group's influence variance is positive. We estimate it in that form,
\[
\hat\Sigma \;=\; \sum_g v_g\, c_g c_g^{\top}, \qquad v_g = \frac{\operatorname{Var}_{\mathrm{lab},g}\big(\psi^{H}{-}\lambda\,\psi^{J}\big)}{n_{\mathrm{lab},g}} + \lambda^2\,\frac{\operatorname{Var}_{\mathrm{unlab},g}\big(\psi^{J}\big)}{n_{\mathrm{unlab},g}},
\]
over the columns $c_g$ of $M$, with $v_g$ the variance of group $g$'s contribution, the sum of two independent averages as in \S\ref{app:rank:mwu}, so $\mathrm{df} = k-1$. 

The statistic $Q = \hat\delta^{\top}\hat\Sigma^{+}\hat\delta$ is asymptotically $\chi^2_{k-1}$, but that reference is liberal at small $N_\mathrm{lab}$, where a noisy $\hat\Sigma$ inflates $E[Q]$. As a small-sample calibration we refer $Q(\nu-\mathrm{df}+1)/(\nu\,\mathrm{df})$ to $F(\mathrm{df},\,\nu-\mathrm{df}+1)$, the Hotelling-$T^2$ scaling~\cite{hotelling1931generalization}, with $\nu = N_\mathrm{lab}$ the total labeled count (falling back to $\chi^2_{\mathrm{df}}$ when $\nu \le \mathrm{df}$); this is a calibration choice, not a claim that $\hat\Sigma$ is Wishart. The reported effect size, $4\cdot\mathrm{mean}(\hat\delta_{ab}^2)\in[0,1]$ and zero under $H_0$, is a scalar summary we define here and PPI-correct separately, with the fixed $\lambda{=}1$ estimator and a percentile-bootstrap interval.

\subsection{PPI-corrected Wilcoxon signed-rank test}
\label{app:rank:wilcoxon}

For paired differences $D = X - Y$, the estimand is the Walsh-average dominance probability
\[
\theta_W \;=\; P\!\left(\tfrac{D + D'}{2} > 0\right) + \tfrac12\,P\!\left(\tfrac{D + D'}{2} = 0\right) - \tfrac{1}{2},
\]
with $D, D'$ independent copies. Its estimator on observed differences $d_1,\dots,d_n$ is
\[
\hat\theta_W \;=\; \frac{2}{n(n+1)}\sum_{i \le j}\Big[\mathbf{1}\{d_i + d_j > 0\} + \tfrac12\,\mathbf{1}\{d_i + d_j = 0\}\Big] - \tfrac12,
\]
which includes the $i = j$ terms (a contribution of order $1/n$) and counts ties at half weight. This is the sign construction behind the Hodges-Lehmann estimator~\cite{hodges1963estimates}, and $H_0\!:\theta_W = 0$ says the Walsh pseudomedian is zero. It is implied by, but weaker than, the classical signed-rank null of differences symmetric about zero; under symmetry it is the usual Wilcoxon null. By Tukey's identity~\cite{hollander2013nonparametric}, with no zero differences and no ties among the $|d_i|$, the classical $W^{+}$ equals the number of positive Walsh averages, so $\hat\theta_W$ is then an exact affine function of $W^{+}$. The mid-probability form with self-pairs is the extension we use for zeros and ties, which retains information that a per-item sign construction discards under heavy ties (e.g.\ integer Likert scores).

The corrected estimator applies the same functional to the human and judge differences on the labeled subset and to the judge differences on the unlabeled items,
\[
\hat\theta \;=\; \hat\theta^{H}_{\mathrm{lab}} + \lambda\big(\hat\theta^{J}_{\mathrm{unlab}} - \hat\theta^{J}_{\mathrm{lab}}\big),
\qquad
\lambda \;=\; \frac{\widehat{\operatorname{Cov}}\big(\hat\theta^{H}_{\mathrm{lab}}, \hat\theta^{J}_{\mathrm{lab}}\big)}{V^{J}_{\mathrm{unlab}} + V^{J}_{\mathrm{lab}}}.
\]

The covariance and judge variances here are analytic, from the estimand's H\'ajek projection: with $\psi_i = \frac1n\sum_{j}\big[\mathbf{1}\{d_i + d_j > 0\} + \tfrac12\,\mathbf{1}\{d_i + d_j = 0\}\big]$ the empirical first-order component of item $i$ (the one-sample analogue of the $\psi_i$ of \S\ref{app:rank:mwu}), $\widehat{\operatorname{Var}}(\hat\theta_W) = 4\operatorname{Var}(\psi)/n$, and the covariance of the human and judge estimators on the labeled items is the corresponding covariance of their projections. Here $V^{J}_{\mathrm{lab}}$ and $V^{J}_{\mathrm{unlab}}$ are this same $4\operatorname{Var}(\psi^{J})/n$ evaluated on the labeled and unlabeled judge differences. Each is computed within its own sample (the Walsh kernel pairs items inside a single sample, so unlike \S\ref{app:rank:mwu} there is no shared reference distribution to fix).

The human-side variance is evaluated under $H_0$ rather than at the estimate. Under sign symmetry, flipping the signs of the observed labeled human differences at random, conditional on their magnitudes, leaves the null distribution unchanged, so the variance of $\hat\theta^{H}_{\mathrm{lab}}$ across such flips is its null variance: the classical randomization reference for a signed-rank statistic~\cite[Ch.~15]{lehmann2005testing}, valid in the presence of ties. We estimate it from 200 random sign flips. If every labeled difference is zero, or either labeled variance is numerically zero, the human term falls back to its H\'ajek variance, which is itself zero in that degenerate case, leaving the judge terms to carry the standard error. Writing $V^{H}_{\mathrm{null}}$ for it, $\rho$ for the estimated correlation between the human and judge projections on the labeled items, and $r = \hat\theta^{J}_{\mathrm{unlab}} - \hat\theta^{J}_{\mathrm{lab}}$,
\[
\widehat{\operatorname{Var}}(\hat\theta) \;=\; V^{H}_{\mathrm{null}} + \lambda^2\big(V^{J}_{\mathrm{unlab}} + V^{J}_{\mathrm{lab}}\big) - 2\lambda\,\rho\,\sqrt{V^{H}_{\mathrm{null}}\,V^{J}_{\mathrm{lab}}} + r^2\,\widehat{\operatorname{Var}}(\lambda),
\]
which keeps the estimated correlation so that the quadratic form stays non-negative. This is score rather than Wald inference: $\theta_W$ is proportion-like, with variance typically largest near $0$ and collapsing toward $\pm\tfrac12$, so a plug-in variance at $\hat\theta_W$ shrinks as $|\hat\theta_W|$ grows and a two-sided test built on it is anti-conservative. The statistic $\hat\theta/\widehat{\mathrm{SE}}$ is referred to a $t$ distribution on $N_\mathrm{lab}-1$ degrees of freedom. 

The reported effect size is the rank-biserial correlation $2\hat\theta$, clipped to $[-1,1]$ as in \S\ref{app:rank:mwu}.

\begin{figure*}[t]\centering
\includegraphics[width=\linewidth]{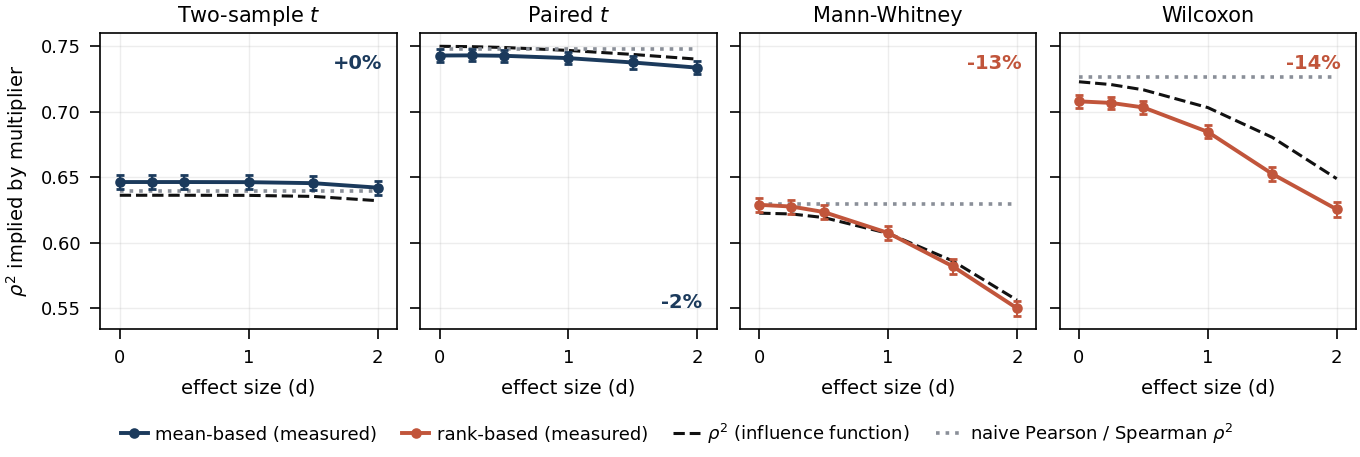}
\caption{The variance multiplier is a property of judge quality only for parametric tests, and degrades with effect size for rank-based ones. Each panel sweeps the true effect at fixed judge quality ($N=1000$, $N_\mathrm{lab}=100$, continuous, near-centred truth marginal), plotting the
$\rho^2$ implied by the measured multiplier (solid, $\pm1$ MC standard error over
20{,}000 replicates). Parametric tests stay flat, as the algebra requires~\cite{angelopoulos2023prediction}, while rank-based tests fall  ($-0.079\pm0.008$ Mann-Whitney, $-0.083\pm0.007$ Wilcoxon). Dashed lines show that \pkg{}' influence-function $\rho^2$, output from its judge\_alignment method, stays within $5\%$ of the measured value.} %
\label{fig:le-esinv}\end{figure*}

\clearpage
\onecolumn  %

\section{Label efficiency and robustness of the $\rho^2\ge0.4$ rule of thumb}
\label{app:label-eff-rt}

We sweep judge quality against labeling budget on a pool of
${N=}1000$ items, labeling $N_\mathrm{lab}\in\{15,\allowbreak 20,\allowbreak 30,\allowbreak 40,\allowbreak 60,\allowbreak 90,\allowbreak 130,\allowbreak 200\}$ of them, at six judge-quality tiers calibrated to target $\rho^2$, four effect sizes ($0.15$, $0.20$, $0.25$ and $0.35$ of the eval type's population SD, see Table~\ref{tab:pop-sd}), and two judge-error distributions (Gaussian, and a contaminated judge that is mostly right but occasionally badly wrong), with 500 replicates per cell. The \emph{label-efficiency multiplier} is the number of human labels a classical test would need to match PPI's power, divided by $N_\mathrm{lab}$.

\begin{table}[t]
\footnotesize
\begin{tabular}{llp{0.42\linewidth}@{}}
\toprule
\textbf{Planned test} & \textbf{Use} & \textbf{Computed between LLM judge \& humans on\ldots} \\
\midrule
\multicolumn{3}{@{}l}{\textit{Estimation and two-condition tests}} \\
\addlinespace[2pt]
Mean, proportion, CI              & $\rho_P^2$ & raw scores, per measure \\
Independent-samples $t$-test      & $\rho_P^2$ & raw scores, within each condition \\
Paired $t$-test                   & $\rho_P^2$ & \emph{difference} scores ($D_i = Y_{1,i} - Y_{2,i}$) \\
Mann--Whitney $U$                 & $\rho_S^2$ & each item's mid-rank placement against the other condition \\
Wilcoxon signed-rank              & $\rho_S^2$ & \emph{difference} scores, via their H\'ajek projection \\
\addlinespace[3pt]
\multicolumn{3}{@{}l}{\textit{Omnibus tests ($k \ge 3$ conditions): Use the \texttt{judge\_alignment} function.}} \\
\addlinespace[2pt]
One-way ANOVA        & $\rho_P^2$ & scores centered within each condition, pooled \\
RM ANOVA             & $\rho_P^2$ & residuals after removing participant \emph{and} condition means \\
Kruskal--Wallis      & $\rho_S^2$ & scores centered within condition, pooled, then ranked \\
Friedman             & $\rho^2$ & within-participant rankings of the $k$ conditions, condition-centered \\
\bottomrule
\end{tabular}
\caption{Which judge-human correlation $\rho^2$ to use in the $N_{\text{eff}}$ formula, and what to compute it on. \pkg{}' \texttt{judge\_alignment} computes the correct value for a given test from labeled data. For Mann--Whitney and Wilcoxon it correlates influence-function values, which reduce to Spearman's $\rho_S$ at a null effect. (Note: For within-subjects/repeated-measures designs, randomly sample rows (subjects) with all $k$ conditions filled, and have human raters label all $k$ per selected subject.)}
\label{tab:which-rho}
\end{table}

\begin{figure*}[t]\centering
\includegraphics[width=0.7\linewidth]{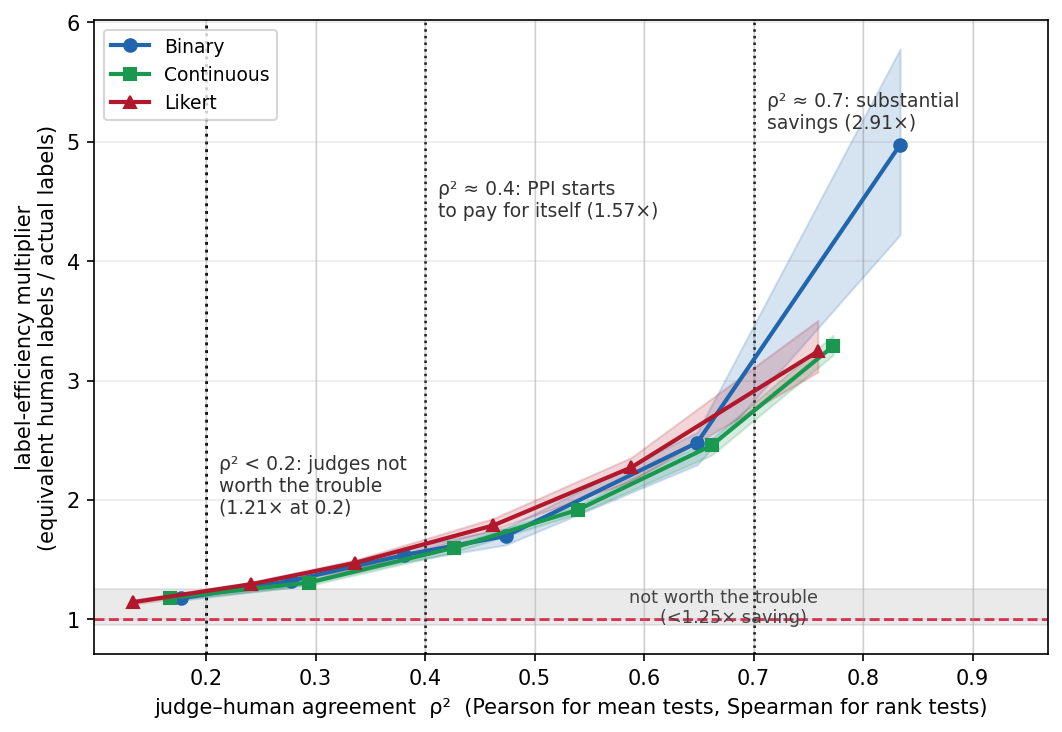}
\caption{Label-efficiency multiplier against judge-human agreement $\rho^2$, one line per data type, with bootstrap 95\% CIs on the median. The shaded band marks savings under $1.25\times$: statistically above $1$, but perhaps not worth the cost. By $\rho^2\approx0.4$ all three data types have cleared it cleanly. Results are averaged over parametric and nonparametric tests using their respective $\rho$, showing the threshold is invariant to test type,
data type, effect size, and setup. A per-family breakdown is in Figure~\ref{fig:le-lookup}. 500 replicates per cell (same run as Figure~\ref{fig:label_efficiency}).}
\label{fig:le-threshold}

\vspace{1.0em}

\includegraphics[width=\linewidth]{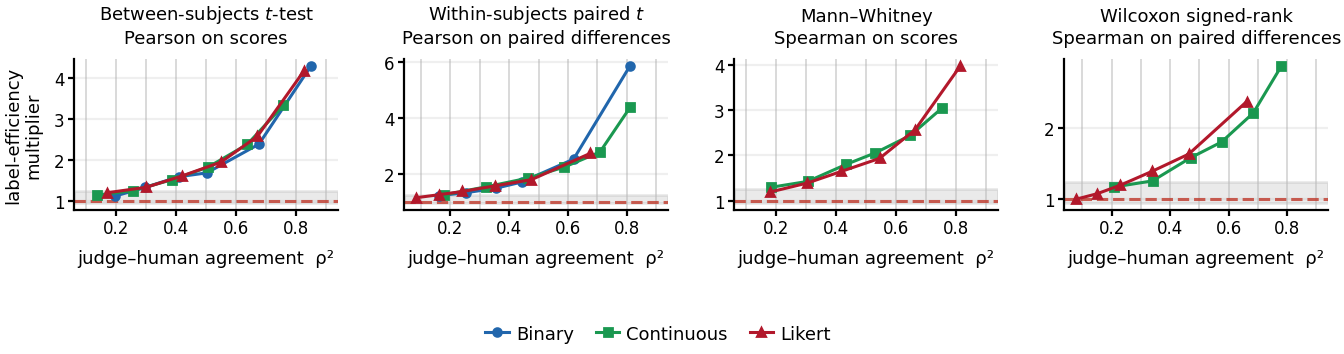}
\caption{Which correlation to measure, by experimental design: Pearson for
mean-based tests, Spearman for rank-based, and the same correlations on paired
differences for paired designs. Rank tests do not show binary because they are not appropriate there. Same run as Figure~\ref{fig:le-threshold}.} 
\label{fig:le-lookup}\end{figure*}

\textit{Rules of thumb.} Figure~\ref{fig:le-threshold} plots the
multiplier against judge-human agreement. Below $\rho^2\approx0.2$ the saving
is under $1.25\times$ for every data type, teetering just above 1. At $\rho^2\approx0.4$ every data type clears $1.25\times$, which is why we recommend it as the threshold to aim for. Figure~\ref{fig:label_efficiency} breaks this out by labeling budget. Figure~\ref{fig:le-esinv} shows the multiplier is invariant to effect size for parametric tests, approximately so for the rank-based ones. A rank test's effective $\rho^2$ declines as the
true effect grows, because rank-based tests' remaining variance concentrates in rare rank reversals. The decline costs under $2\%$ at Cohen's $d\le0.5$. It should not necessarily be interpreted as a loss of power, however. %

$\rho$ is the Pearson correlation between the judge's and the human's influence
values, which follows from the general PPI variance reduction, $1-\operatorname{Corr}^2$ between the influence functions of the labeled and predicted statistics~\cite{angelopoulos2023ppiplusplus}. For a mean that quantity is the score itself, so $\rho$ is a plain Pearson's $r$. For a rank-based test it is the item's position within the sample, an empirical-CDF value, which at a null effect gives Spearman's $\rho_S$. For a
paired design it is the per-item \emph{difference} rather than the raw score.

\section{End-to-end validation of \texttt{compare()}} \label{app:e2e}

We feed \texttt{compare()} the same synthetic data generation suite (\S\ref{sec:datagen:synth}), sweeping 23 distribution shapes across binary,
continuous, and Likert data; $N \in \{15, \ldots, 1000\}$ items per arm;
$k \in \{2, 3, 5, 10\}$ arms for FWER correction; cross-arm reliability $\mathrm{ICC} \in \{0.05, 0.20, 0.60\}$ (on the non-PPI path; PPI cells use $0.20$); and, on the PPI path, human-label
budgets of $10$, $20$, and $40\%$ of items. Judge bias is applied only on the PPI path; the non-PPI arm
analyzes the evaluation results directly, treated them as trusted. Beyond run settings (score range, data type, bootstrap count, and seed), we pass only \texttt{factors=}, \texttt{metric=}, and (where applicable) \texttt{alignment=}: every method choice is left to \texttt{method="auto"} and the routing described in \S\ref{sec:fwer}, except the FWER correction below. For FWER correction of PPI hypothesis tests, we set \texttt{correction="shaffer"}. We do this for consistency of what we validated here, so that \pkg{} routes to the PPI-Wilcoxon and PPI-paired-$t$ tests for numeric and binary data, respectively, rather than Romano-Wolf, as R-W would replace the pairwise $p$-values. A PPI-corrected Romano-Wolf exists in \pkg{}, but we spare the reader the added complexity. 

Table~\ref{tab:e2e:bydatatype} provides descriptive statistics behind Figure~\ref{fig:e2e}.

\begin{table}[H]
\centering
\scriptsize
\setlength{\tabcolsep}{4pt}
\begin{tabular}{lrrrrrrr}
\toprule
 & \multicolumn{3}{c}{CI coverage (target $.95$)} & & \multicolumn{3}{c}{PPI power gain over labels-only (points)} \\
\cmidrule(lr){2-4}\cmidrule(lr){6-8}
Data type & Per-arm & Pairwise & Family & Type-I & Peak & at $N_\mathrm{lab}$ & at $N_\mathrm{lab}{=}400$ \\
\midrule
Binary & 0.957 & 0.952 & 0.969 & 0.031 & +18.9 & 100 & +0.5 \\
Continuous & 0.952 & 0.949 & 0.962 & 0.040 & +11.7 & 40 & +0.0 \\
Likert & 0.952 & 0.950 & 0.958 & 0.040 & +13.1 & 40 & +4.1 \\
\midrule
\textit{All} & 0.954 & 0.950 & 0.963 & 0.037 & +14.0 & 40 & +1.6 \\
\bottomrule
\end{tabular}
\caption{End-to-end results of \texttt{compare()} by data type (100 MC reps per cell), a descriptive summary of
Figure~\ref{fig:e2e}.} %
\label{tab:e2e:bydatatype}
\end{table}

\clearpage
\onecolumn  %
\section{Full-size plots of IRR metrics under differential judge bias, per data type}
\label{app:irr-peak}

\begin{figure}[H]
  \centering
  \includegraphics[width=\linewidth,height=0.80\textheight,keepaspectratio]{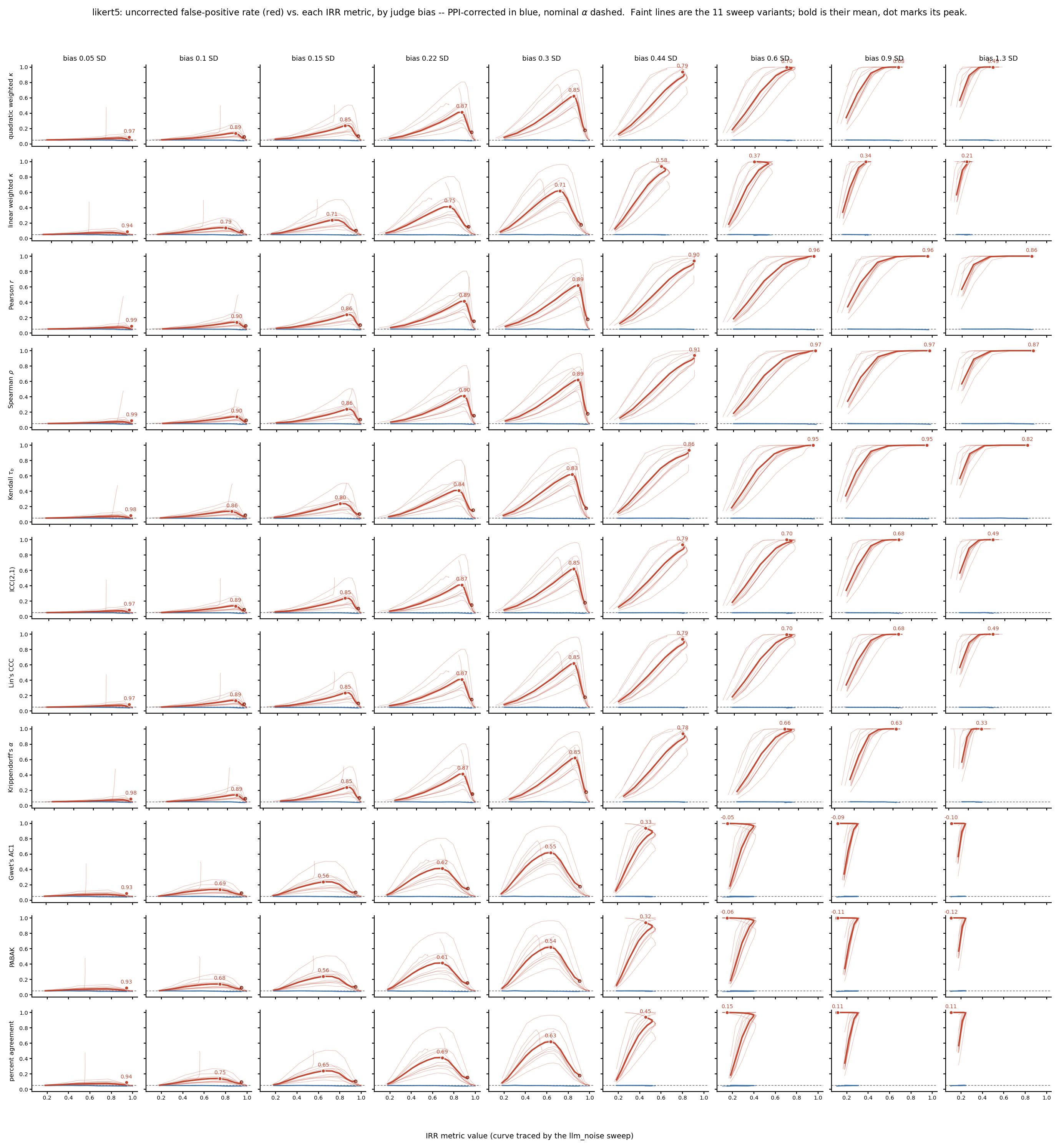}
  \caption{5-point Likert: uncorrected false-positive rate (red) against
  each IRR metric (rows), by judge bias (columns, in population SDs). Faint
  lines are the eleven judge variants; bold is their mean, with the dot
  marking its peak. PPI-corrected rates in blue hold at nominal ($\alpha$
  dashed) in every cell.}
  \label{fig:irrpeak:likert5}
\end{figure}

\begin{figure}[H]
  \centering
  \includegraphics[width=\linewidth,height=0.86\textheight,keepaspectratio]{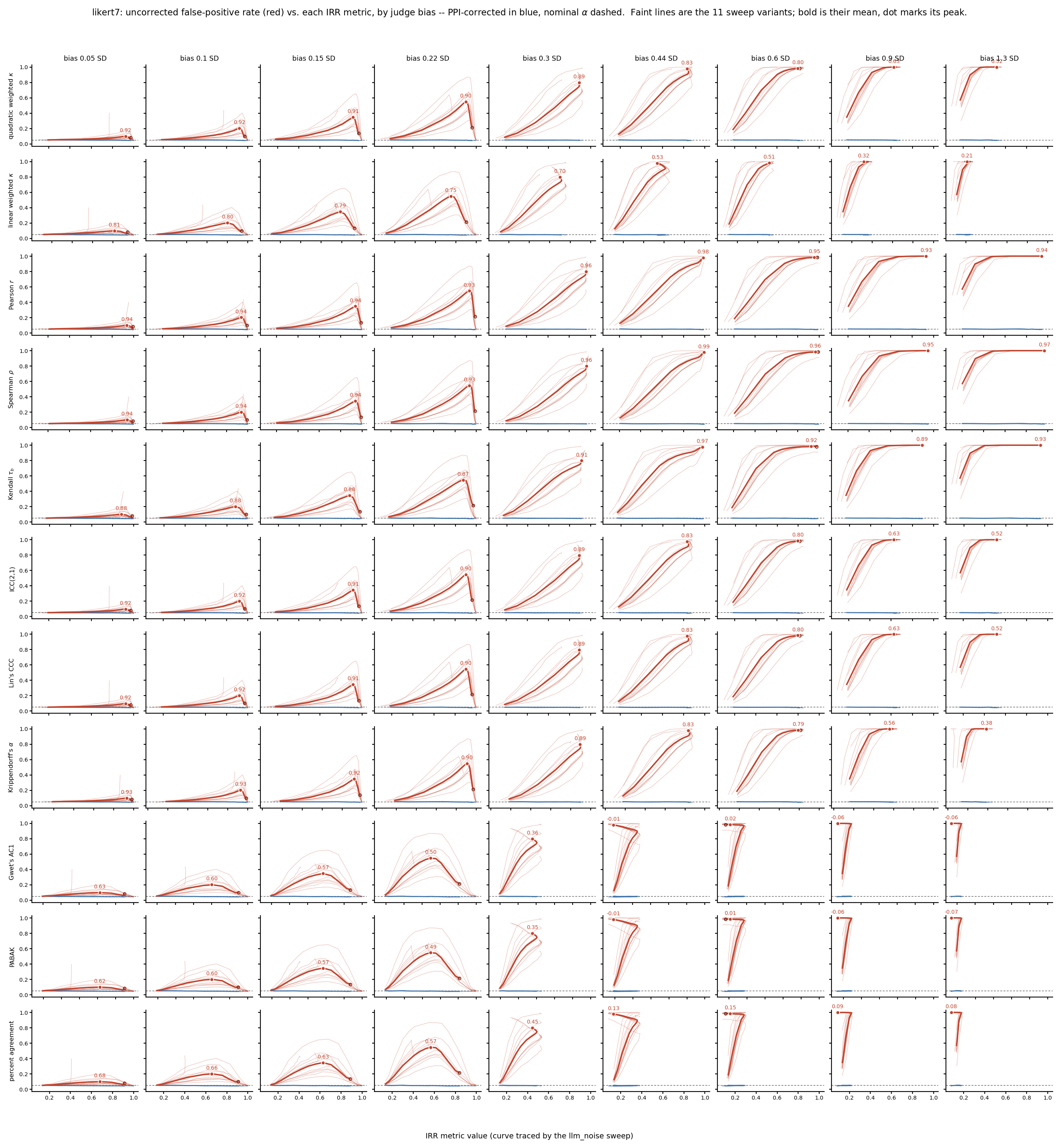}
  \caption{7-point Likert, same layout as Fig.~\ref{fig:irrpeak:likert5}.
  The peaks sit higher (0.92--0.95) because half a scale point, the
  rounding threshold, is a smaller fraction of the scale's SD.}
  \label{fig:irrpeak:likert7}
\end{figure}

\begin{figure}[H]
  \centering
  \includegraphics[width=\linewidth,height=0.86\textheight,keepaspectratio]{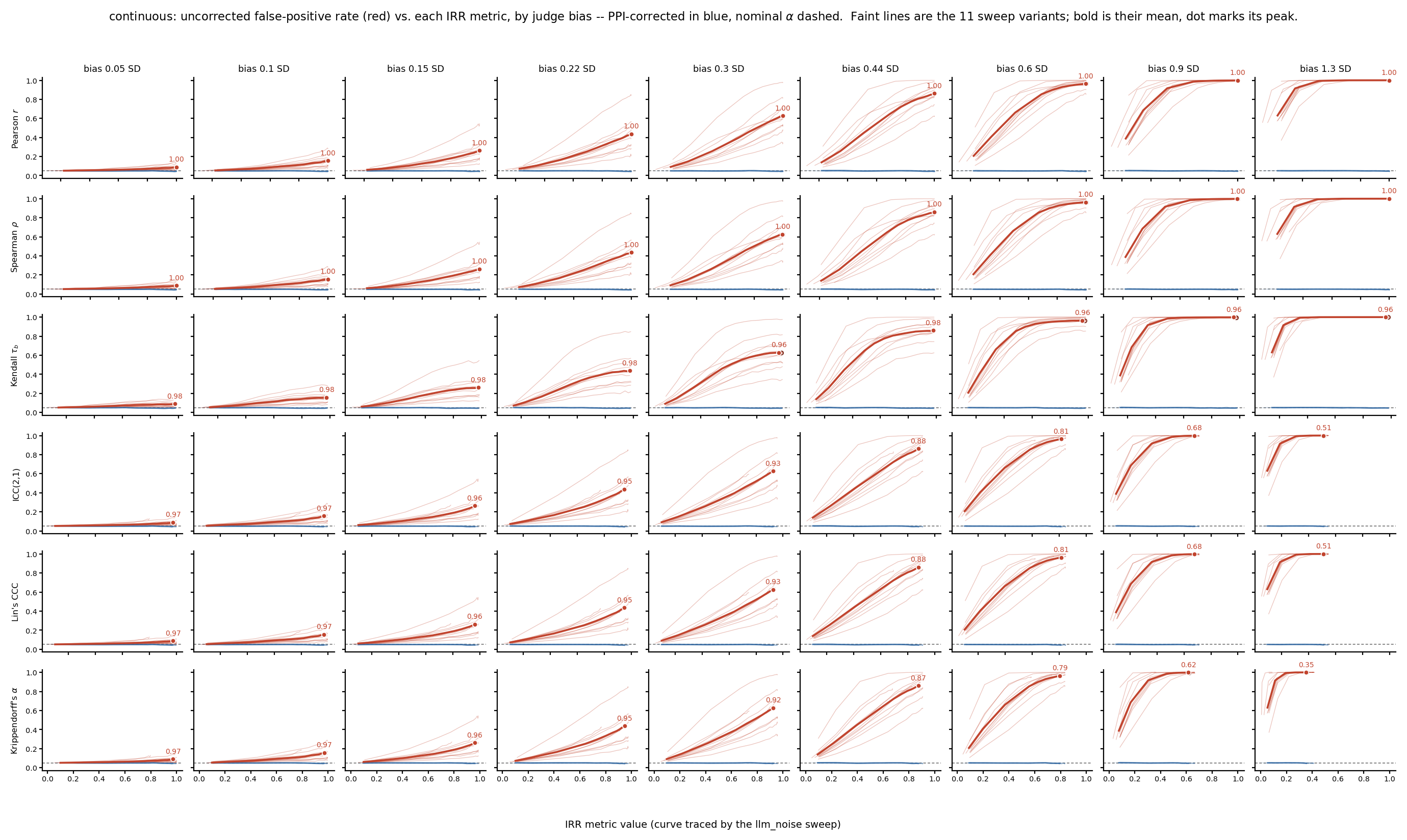}
  \caption{Continuous data IRR metrics. False positive risk rises monotonically across IRR metrics.}
  \label{fig:irrpeak:continuous}
\end{figure}

\begin{figure}[H]
  \centering
  \includegraphics[width=\linewidth,height=0.86\textheight,keepaspectratio]{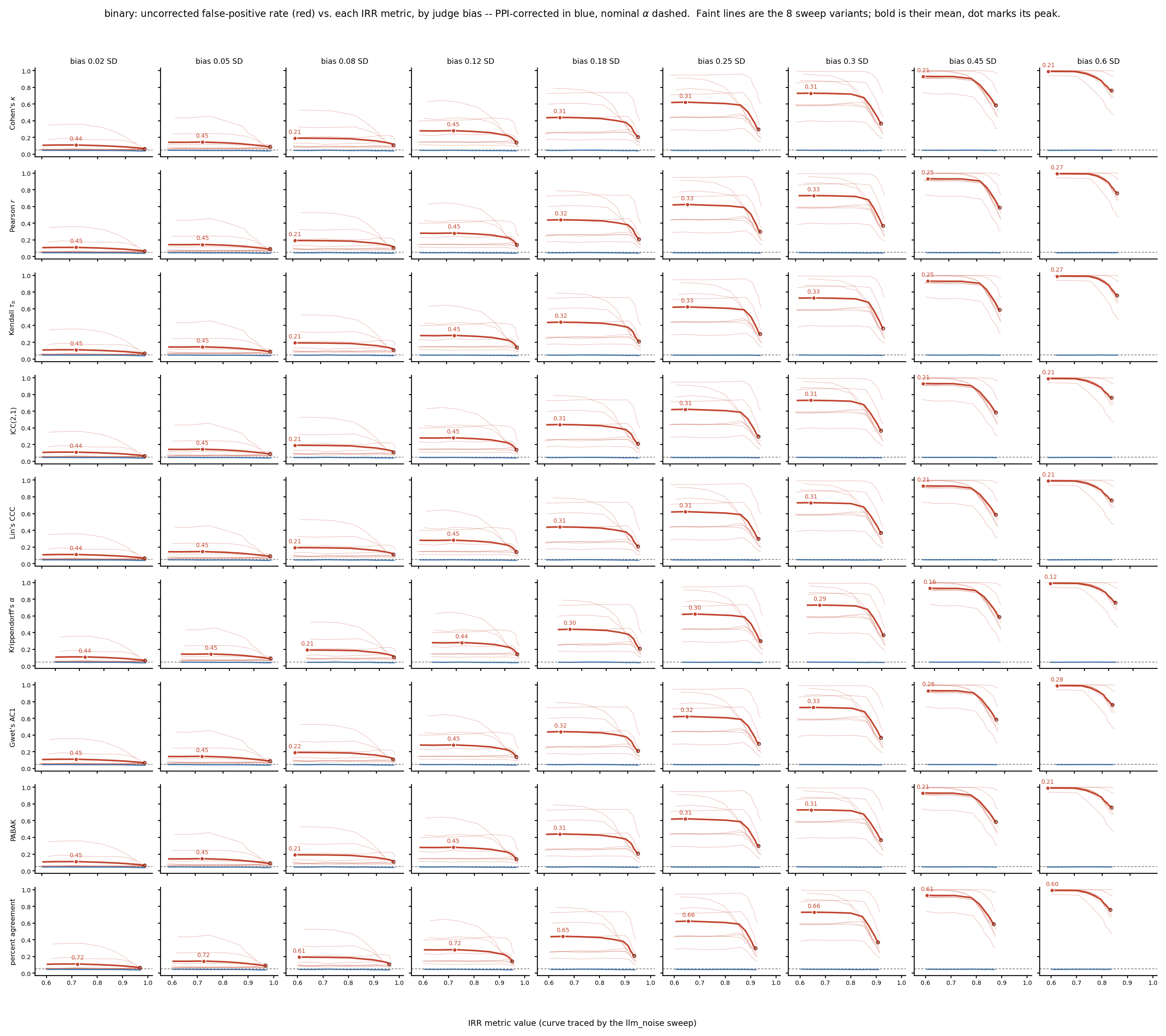}
  \caption{Binary data IRR metrics as judge bias grows. False positive risk falls as agreement rises.}
  \label{fig:irrpeak:binary}
\end{figure}

Figures~\ref{fig:irrpeak:likert5}--\ref{fig:irrpeak:binary} sweep systematic judge bias against judge noise for 5- and 7-point Likert, continuous, and binary data, plotting the uncorrected false-positive rate against every common IRR metric for that data type across up to eleven judge variants (1{,}000 replicates per cell). PPI-corrected two-sample tests held at an average Type I error of 0.047 across all 5{,}420 cells.

\end{document}